%% file: Main.tex
\documentclass{article}
\usepackage{cite}
\usepackage{amsmath,amssymb,amsfonts}
\usepackage{amsthm}
\usepackage{mathrsfs}
\usepackage{algorithmic}
\usepackage{float}
\usepackage{graphicx}
\usepackage{subfigure}
\usepackage[justification=centering]{caption}
\usepackage{textcomp}
\usepackage{xcolor}
\usepackage{colortbl,booktabs}
\usepackage{times}  
\usepackage{helvet}  
\usepackage{courier}  
\usepackage{soul}
\usepackage{bm}
\usepackage{bbm}
\newtheorem{theorem}{\textbf{Theorem}}
\newtheorem{remark}{\textbf{Remark}}
\newtheorem{theorem*}{Theorem}

\newtheorem{Definition}{Definition}

\newtheorem{definition}{Problem}
\newtheorem{Definition*}{Assumption}

\DeclareMathOperator*{\argmin}{arg\,min}

\usepackage{microtype}
\usepackage{graphicx}
\usepackage{subfigure}
\usepackage{booktabs} 
\usepackage{hyperref}
\newcommand{\httpsurl}[1]{\href{https://#1}{\nolinkurl{#1}}}

\usepackage[accepted]{icml2021}

\begin{document}

\twocolumn[
\icmltitle{Learning Bounds for Open-Set Learning}



\icmlsetsymbol{equal}{*}
\begin{icmlauthorlist}
\icmlauthor{Zhen Fang}{equal,to}
\icmlauthor{Jie Lu}{to}
\icmlauthor{Anjin Liu}{equal,to}
\icmlauthor{Feng Liu}{to}
\icmlauthor{Guangquan Zhang}{to}
\end{icmlauthorlist}

\icmlaffiliation{to}{AAII, University of Technology Sydney}

\icmlcorrespondingauthor{Zhen Fang}{zhen.fang@student.uts.edu.au}
\icmlcorrespondingauthor{Jie Lu}{jie.lu@uts.edu.au}
\icmlcorrespondingauthor{Anjin Liu}{anjin.liu@uts.edu.au}

\icmlkeywords{Machine Learning, ICML}

\vskip 0.3in
]



\printAffiliationsAndNotice{\icmlEqualContribution}
\begin{abstract}

Traditional supervised learning aims to train a classifier in the closed-set world, where training and test samples share \emph{the same} label space. 
In this paper, we target a more challenging and realistic setting: \textit{open-set learning} (OSL), where there exist test samples from the classes that are unseen during training. 
Although researchers have designed many methods from the algorithmic perspectives, there are few methods that provide generalization guarantees on their ability to achieve consistent performance on different training samples drawn from the same distribution.
Motivated by the \emph{transfer learning} and \emph{probably approximate correct} (PAC) theory, we make a bold attempt to study OSL by proving its generalization error$-$given  training samples with size $n$, the estimation error  will get close to order $O_p(1/\sqrt{n})$.
This is the first study to provide a generalization bound for OSL, which we do by theoretically investigating the risk of the target classifier on unknown classes.
According to our theory, a novel algorithm, called  \textit{auxiliary open-set risk} (AOSR) is proposed to address the OSL problem. 
Experiments verify the efficacy of AOSR. The code is available at
\httpsurl{github.com/Anjin-Liu/Openset_Learning_AOSR}.
\end{abstract}

\section{Introduction}
Supervised learning has achieved dramatic successes in many applications such as object detection \cite{DBLP:journals/corr/SimonyanZ14a}, speech recognition \cite{DBLP:conf/icml/GravesJ14} and natural language processing \cite{DBLP:conf/icml/CollobertW08}. These successes
are partly rooted in the closed-set assumption 
that training and test samples share a \textit{same label space}. Under this assumption, the standard supervised learning
is also regarded as \textit{closed-set learning} (CSL) \cite{DBLP:journals/corr/abs-1811-08581,yang2020convolutional}. 

However, the closed-set assumption is not realistic during the testing phase (i.e., there are no labels in the samples) since it is not known whether the classes of test samples are from the label space of training samples. Test samples may come from some classes (\textit{unknown classes}) that are not necessarily seen during training.
These unknown classes can emerge unexpectedly and drastically weaken the performance of existing closed-set algorithms \cite{DBLP:journals/ml/CardosoGF17,DBLP:conf/nips/DhamijaGB18,DBLP:conf/cvpr/PereraMJMWOP20}.

To solve supervised learning without closed-set assumption, \citeauthor{DBLP:journals/pami/ScheirerRSB13} (\citeyear{DBLP:journals/pami/ScheirerRSB13}) proposed a new problem setting, \textit{open-set learning} (OSL), in which the test samples can come from any classes, even unknown classes. An open-set classifier should classify samples from known classes into  correct known classes while recognizing samples from unknown classes into unknown classes.

Remarkable advances have been achieved in open-set learning. 
The key challenge of OSL algorithms is to recognize the unknown classes accurately. To address this challenge, different strategies have been proposed such as open-space risk \cite{DBLP:journals/pami/ScheirerRSB13} and extreme value theory \cite{DBLP:conf/eccv/JainSB14,DBLP:journals/pami/RuddJSB18}. Further, to adapt deep networks support to OSL, \citeauthor{DBLP:conf/cvpr/BendaleB16}  (\citeyear{DBLP:conf/cvpr/BendaleB16}), \citeauthor{DBLP:conf/bmvc/GeDG17} (\citeyear{DBLP:conf/bmvc/GeDG17}) proposed OpenMax and G-OpenMax, respectively.

While many OSL algorithms can be roughly interpreted as minimizing the open-space risk or using the extreme value theory, several disconnections still form non-negligible gaps between the theories and algorithms \cite{DBLP:journals/corr/abs-1811-08581,DBLP:conf/aaai/BoultCDGHS19}.
Very little theoretical groundwork has been undertaken to reveal the generalization ability of OSL from the perspective of learning theory.

This work aims to bridge the gap between the theory and algorithm for OSL from the perspective of learning theory. In particular, our theory answers an important question: under some assumptions,  given training samples with size $n$, then \textit{there exists an OSL algorithm such that the estimation error is close to $O_p(1/\sqrt{n})$}. This result reveals OSL problem can achieve an order of estimation error that is the same as CSL \cite{shalev2014understanding}.

Since the test samples contain unknown classes, the distribution of test samples is intrinsically different from that of training samples. Based on this fact, we aim to establish the OSL theory from \emph{transfer learning} \cite{luo2020adversarial,DBLP:journals/tai/NiuLWS20,DBLP:conf/cvpr/DongCSZX20,DBLP:journals/kbs/LuBHZXZ15,DBLP:conf/iccv/DongCSH19,DBLP:conf/eccv/DongCSLX20,DBLP:journals/tkde/PanY10,DBLP:journals/tcsv/DongCSYXD21,liu2019butterfly,wang2020prototype}, which learns knowledge for a given domain from a different, but relative domain. 
Using the transfer learning theory, we focus on constructing a suitable auxiliary domain, which contains the information of unknown classes. The construction of auxiliary domain  depends on \textit{covariate shift} \cite{DBLP:conf/icml/SanturkarSM18}. Transferring information from the auxiliary domain, we construct the generalization bound for OSL by using the transfer learning bound developed by \citeauthor{DBLP:conf/nips/Ben-DavidBCP06} (\citeyear{DBLP:conf/nips/Ben-DavidBCP06}), \citeauthor{DBLP:conf/colt/MansourMR09} (\citeyear{DBLP:conf/colt/MansourMR09}), \citeauthor{Fang2020OpenSD} (\citeyear{Fang2020OpenSD}), \citeauthor{DBLP:journals/corr/abs-2006-13022} (\citeyear{DBLP:journals/corr/abs-2006-13022,DBLP:journals/corr/abs-2101-01104}), \citeauthor{DBLP:conf/icml/LuoWHB20} (\citeyear{DBLP:conf/icml/LuoWHB20}).


Guided by our theory, we then devise an algorithm for OSL to bring the proposed OSL theory into reality. The novel algorithm \textit{auxiliary open-set risk} (AOSR) is a neural network-based algorithm. AOSR mainly utilizes the instance-weighting strategy to align training samples and auxiliary samples generated by an auxiliary domain. Then, minimizing the \textit{auxiliary risk} developed by our theory, AOSR can learn how to recognize unknown classes. 

The contributions of this paper are summarized as follows.

$\bullet$ We provide the theoretical analysis for open-set learning based on transfer learning and PAC theory. This is the \emph{first} work to investigate the  generalization error bound for open-set learning.

$\bullet$ Our theory answers an important question: under some assumptions,  there exists an OSL algorithm such that the order of the estimation error is close to $O_p(1/\sqrt{n})$, if given training samples with size $n$.

$\bullet$ 
We conduct experiments on toy and benchmark datasets. Experiments support our theoretical results and show that our theoretical guided algorithm AOSR can achieve competitive performance compared with several popular baselines.

\section{Related Works}
\textbf{Open-Set Learning Theory.} One of the pioneering theoretical works in this field was conducted by \citeauthor{DBLP:journals/pami/ScheirerRSB13} (\citeyear{DBLP:journals/pami/ScheirerRSB13,DBLP:journals/pami/ScheirerJB14}). They proposed the open-space risk, which means that when a sample is far from the training samples, there is an increased risk that the sample is from unknown classes. By minimizing the open-space risk, samples from unknown classes can be recognized. \citeauthor{DBLP:conf/eccv/JainSB14} (\citeyear{DBLP:conf/eccv/JainSB14}), \citeauthor{DBLP:journals/pami/RuddJSB18} (\citeyear{DBLP:journals/pami/RuddJSB18}) consider the extreme value theory to solve the OSL problem. Extreme value theory is a branch of statistics analyzing the distribution of samples of abnormally high or low values. \citeauthor{DBLP:journals/corr/abs-1808-00529} (\citeyear{DBLP:journals/corr/abs-1808-00529}) first proposed the PAC guarantees for open-set detection. Unfortunately, the test samples are required to be used in the training phase. \citeauthor{Fang2020OpenSD} (\citeyear{Fang2020OpenSD}) considered the open-set domain adaptation (OSDA) problem  \cite{DBLP:conf/icml/LuoWHB20,DBLP:journals/pami/BustoIG20} and proposed the first estimation for the generalization error of OSDA by constructing a special term, open-set difference. However, similar to \citeauthor{DBLP:journals/corr/abs-1808-00529} (\citeyear{DBLP:journals/corr/abs-1808-00529}), test samples are needed during the training phase. 

\textbf{Open-Set Algorithm.} We can roughly separate OSL algorithms into two different categories: shadow algorithms (e.g., \emph{support vector machine} (SVM)) and deep learning-based algorithms. In shadow algorithms, \citeauthor{DBLP:journals/pami/ScheirerRSB13} (\citeyear{DBLP:journals/pami/ScheirerRSB13,DBLP:journals/pami/ScheirerJB14}) proposed the OSL algorithms based on SVM. \citeauthor{DBLP:conf/eccv/JainSB14} (\citeyear{DBLP:conf/eccv/JainSB14}), \citeauthor{DBLP:journals/pami/RuddJSB18} (\citeyear{DBLP:journals/pami/RuddJSB18}) proposed OSL algorithms based on extreme value theory. Recently, deep-based algorithms have been developed dramatically. OpenMax as the first deep-based algorithm was proposed by \citeauthor{DBLP:conf/cvpr/BendaleB16} (\citeyear{DBLP:conf/cvpr/BendaleB16}), to replace SoftMax in deep networks. Later, \citeauthor{DBLP:conf/bmvc/GeDG17} (\citeyear{DBLP:conf/bmvc/GeDG17}) combined the generative adversarial networks (GAN) with OpenMax and proposed G-OpenMax. Counterfactual image generation proposed by \citeauthor{DBLP:conf/eccv/NealOFWL18} (\citeyear{DBLP:conf/eccv/NealOFWL18}) is the first OSL algorithm to uses the data augmentation technique by generating the unknown classes so that the decision boundaries between unknown and known classes can be figured out. \citeauthor{DBLP:conf/cvpr/OzaP19} (\citeyear{DBLP:conf/cvpr/OzaP19}) used class conditioned auto-encoders to solve OSL problem, and modeled reconstruction errors using the extreme value theory to find the threshold for identifying known/unknown classes.


\section{Theoretical Analysis of OSL}
In this section, we introduce the basic notations used in this paper and then provide theoretical analysis for open-set learning. All proofs can be found in Appendices B-E.

\subsection{Problem Setting and Concepts}
Here we introduce  the  definition of open-set learning (OSL).
\begin{Definition}[Domain]\label{Domain}
Given a feature (input) space $\mathcal{X}\subset \mathbb{R}^d$ and a label (output) space $\mathcal{Y}$, a domain is  a joint distribution $P_{X,Y}$, where random variables $X\in \mathcal{X}, Y\in \mathcal{X}$. 
\end{Definition}
\vspace{-0.15cm}
\textit{Known classes} are a subset of $\mathcal{Y}$. We define the label space of known classes as $\mathcal{Y}_k$. Then, the \textit{unknown classes} are from the space $\mathcal{Y} /\ \mathcal{Y}_k$. The open-set learning problem is defined as follows.
\begin{definition}[Open-Set Learning]\label{def:problem_setting}
Given independent and identically distributed (i.i.d.) samples  $S=\{(\mathbf{x}^i,\mathbf{y}^i)\}_{i=1}^n$ drawn from $P_{X,Y|Y\in \mathcal{Y}_k}$. The aim of open-set learning  is  to  train  a  classifier $f$ using  $S$ such  that $f$ can  classify
1)  the  sample from  known  classes into correct known classes; 
2) the sample from unknown classes into unknown classes.
\end{definition}

Note that it is not necessary to classify unknown samples into correct unknown classes. For the sake of simplicity, we set all unknown samples are allocated to one big unknown class. Hence, without loss of generality, we assume that
$\mathcal{Y}_k=\{{\mathbf{y}}_c\}_{c=1}^C,$ $\mathcal{Y}=\{{\mathbf{y}}_c\}_{c=1}^{C+1}$, where the label ${\mathbf{y}}_c\in \mathbb{R}^{{C+1}}$ is a one-hot vector, whose $c$-th coordinate is $1$ and other coordinates are $0$. Label ${\mathbf{y}}_{C+1}$ represents unknown classes.

Given a loss function $\ell:\mathbb{R}^{C+1}\times \mathbb{R}^{C+1}\rightarrow \mathbb{R}_{\geq 0}$  and any scoring (hypothesis) function ${\bm h}$ from $\{{\bm h}:\mathcal{X}\rightarrow \mathbb{R}^{C+1}\}$, the \textit{partial risks for known classes and unknown classes} are
\begin{equation}\label{partial risks}
\begin{split}
   & R_{P,k}({\bm h}):= \int_{\mathcal{X}\times \mathcal{Y}_k}\ell({\bm h}(\mathbf{x}),\mathbf{y}) {\rm d}P_{X,Y|Y\in \mathcal{Y}_k}(\mathbf{x},\mathbf{y}),
  \\ & R_{P,u}({\bm h}):= \int_{\mathcal{X}}\ell({\bm h}(\mathbf{x}),\mathbf{y}_{C+1}) {\rm d}P_{X|Y= \mathbf{y}_{C+1}}(\mathbf{x}).
\end{split}
\end{equation}

Then, the \textit{$\alpha$-risk} for $P_{X,Y}$ is 
\begin{equation}\label{risk}
  R_{P}^{\alpha}({\bm h}):= (1-\alpha)R_{P,k}({\bm h})+\alpha R_{P,u}({\bm h}),  
\end{equation}
where $\alpha$ is the weight estimating the importance of unknown classes. When $\alpha=P(Y=\mathbf{y}_{C+1})$, it is easy to check that
\begin{equation*}
    R_{P}^{\alpha}({\bm h})= \mathbb{E}_{(\mathbf{x},\mathbf{y})\sim P_{X,Y}} \ell({\bm h}(\mathbf{x}),\mathbf{y}).
\end{equation*}
Similarly, given a different joint distribution $Q_{X,Y}$, we can define $R_{Q,k}({\bm h}),R_{Q,u}({\bm h})$ and $R_{Q}^{\alpha}({\bm h})$.

Based on $\alpha$-risk, we define almost agnostic probably approximate correct (PAC) for OSL.
\begin{Definition}[Almost Agnostic  PAC Learnability]\label{def:APAC}
A hypothesis class $\mathcal{H}\subset \{{\bm h}:\mathcal{X}\rightarrow \mathbb{R}^{C+1}\}$ is almost agnostic PAC learnable for open-set learning, if given any $\epsilon_0>0$, there exists an OSL algorithm $A_{\epsilon_0}$ such that for given any joint distribution $P_{X,Y}$, there exists $m_{\mathcal{H}}:(0,1)^2 \rightarrow \mathbb{N}$ with the following property: for any $0<\epsilon,\delta<1$, when running the algorithm $A_{\epsilon_0}$ on $n>m_{\mathcal{H}}(\epsilon,\delta)$ i.i.d. samples drawn from $P_{X,Y|Y\in \mathcal{Y}_k}$, the algorithm $A_{\epsilon_0}$ returns a hypothesis $\widehat{\bm h}$ such that, with probability
of at least $1- \delta>0$,
\begin{equation*}
    R_{P}^{\alpha}(\widehat{\bm h})\leq \min_{{\bm h}\in\mathcal{H}} R_{P}^{\alpha}({\bm h})+\epsilon+\epsilon_0.
\end{equation*}
\end{Definition}
\vspace{-0.3cm}
Theorems $\ref{T5}$ and $\ref{T6}$ imply there exists almost agnostic PAC learnable $\mathcal{H}$ for open-set learning under mild assumptions.

\subsection{Transfer Between Domains}

Since there are no samples regarding the unknown classes, we cannot directly analyze the partial risk for unknown classes only using samples $S$ from known classes. To analyze the partial risk for unknown classes, we introduce an auxiliary domain $Q_{X,Y}$, which is used to transfer the information from unknown classes.
\begin{Definition}[Auxiliary Domain]
A domain $Q_{X,Y}$ defined over $\mathcal{X}\times \mathcal{Y}$ is called the auxiliary domain for $P_{X,Y}$, if $Q_{X|Y\in \mathcal{Y}_k}=P_{X|Y\in \mathcal{Y}_k}, Q_{Y|X}=P_{Y|X}$ and $P_X\ll Q_X$.
\end{Definition}

It is clear that $P_{X,Y}$ and $Q_{X,Y}$ are same if we restrict both of them in the support set of known classes. 

\begin{remark}
\label{rem:Q_uniform}
Since we do not have any information about samples from unknown classes in the training set, it is unknown whether $Q_{X|Y=\mathbf{y}_{C+1}}=P_{X|Y=\mathbf{y}_{C+1}}$.  In Section~\ref{sec:contructQ}, we will introduce how to construct $Q_{X,Y}$ such that $Q_{X|Y=\mathbf{y}_{C+1}}$ is a uniform distribution. Namely, any sample drawn from $Q_{X|Y=\mathbf{y}_{C+1}}$ has the same probability.
\end{remark}


Then, it is interesting to know the discrepancy between $R_P^{\alpha}({\bm h})$ and $R_Q^{\alpha}({\bm h})$ given the same hypothesis ${\bm h}$. Before doing this, the disparity discrepancy between distributions need to be introduced.
\begin{Definition}[Disparity Discrepancy \cite{DBLP:conf/icml/0002LLJ19}]\label{doubleloss}Given distributions $P_X, Q_X$ over space ${\mathcal{X}}$,  a hypothesis space $\mathcal{H}\subset \{{\bm h}:{\mathcal{X}}\rightarrow \mathbb{R}^{C+1}\}$ and any hypothesis function ${\bm h}\in \mathcal{H}$, then disparity discrepancy $d_{{\bm h}, \mathcal{H}}^{\ell}(P_X,Q_X)$ is
\begin{equation}
\label{eq:dbl_dd}
\begin{split}
 \sup_{{\bm h}' \in \mathcal{H}} \Big | \int_{\mathcal{X}} \ell({\bm h}(\mathbf{x}),{\bm h}'(\mathbf{x})){\rm d}(P_X-Q_X)(\mathbf{x}) \Big | .
 \end{split}
\end{equation}
\end{Definition}

Using the disparity discrepancy, we can show that
\begin{theorem}\label{T1}
Given a loss $\ell$ satisfying the triangle inequality, and  a hypothesis space $\mathcal{H}\subset \{{\bm h}:\mathcal{X}\rightarrow \mathbb{R}^{C+1} \}$, if $Q_{X,Y}$ is the auxiliary domain for $P_{X,Y}$, then for any ${\bm h}\in \mathcal{H}$, the difference $|R_P^{\alpha}({\bm h})-R_Q^{\alpha}({\bm h})|$ is bounded by
\begin{equation*}
   \alpha  d^{\ell}_{{\bm h},\mathcal{H}}(P_{X|Y=\mathbf{y}_{C+1}},Q_{X|Y=\mathbf{y}_{C+1}})+\alpha \Lambda,
\end{equation*}
where $\alpha=Q(Y=\mathbf{y}_{C+1})$, $d^{\ell}_{{\bm h},\mathcal{H}}$ is the disparity discrepancy defined in Definition \ref{doubleloss},
\begin{equation}\label{combined risk}
\Lambda:=\min_{{\bm h}'\in \mathcal{H}} \big ( R_{P,u}({\bm h}')+R_{Q,u}({\bm h}')\big )
\end{equation}
is the combined risk for the unknown classes, $R_P^{\alpha}({\bm h})$ is the $\alpha$-risk for $P_{X,Y}$ and $R_Q^{\alpha}({\bm h})$ is the $\alpha$-risk for $Q_{X,Y}$.
\end{theorem}

Theorem \ref{T1} implies there exists a gap between $R_P^{\alpha}({\bm h})$ and $R_Q^{\alpha}({\bm h})$. The gap is related to domain discrepancy for unknown classes between $P_{X,Y}$ and $Q_{X,Y}$. To further eliminate the gap between $R_P^{\alpha}({\bm h})$ and $R_Q^{\alpha}({\bm h})$, additional conditions about the hypothesis space $\mathcal{H}$ are indispensable.
\begin{figure*}[t]
    \centering
    \includegraphics[scale=0.5, trim=110 235 0 330, clip]{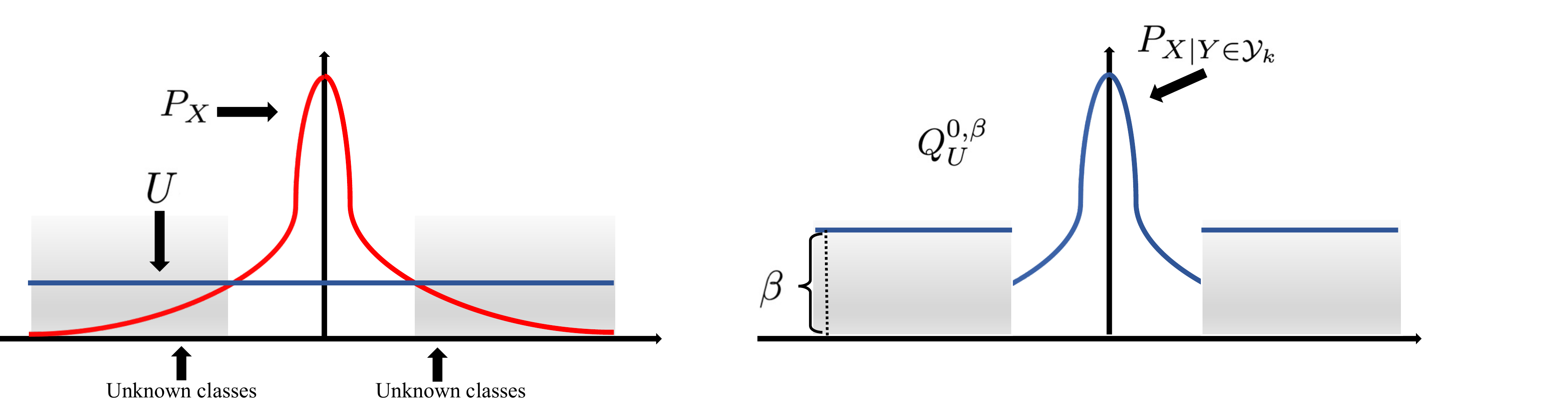}
    \caption{The left figure shows the auxiliary distribution $U$ and marginal distribution $P_X$ (red curve: distribution $P_X$; blue lines: distribution $U$ introduced in Definition \ref{def:IAD}; grey regions: the regions for unknown classes). The right figure shows the marginal distribution $Q_U^{0,\beta}$ of ideal auxiliary domain generated by $U$, $P_{X|Y\in \mathcal{Y}_k}$ and $L_{0,\beta}$ (blue lines and curve: $Q_U^{0,\beta}$ defined in Definition \ref{def:IAD}).}
    \label{fig:distribution}
    \vspace{-1em}
\end{figure*}
\begin{Definition*}[Realization for Unknown Classes]\label{realization}
A hypothesis $\mathcal{H}\subset \{{\bm h}:\mathcal{X}\rightarrow \mathcal{Y}\}$ is realization for unknown classes, if there exist a hypothesis function $\tilde{\bm h}\in \mathcal{H}$ and a distribution $\tilde{P}$ defined over $\mathcal{X}$ with ${\rm supp}~\tilde{P}=\mathcal{X}$ satisfying 
 for any ${\bm h}\in \mathcal{H}$, there exists ${\bm h}'\in \mathcal{H}$ such that ${\bm h}'(\mathbf{x})=\mathbf{y}_{C+1}$, if $\tilde{\bm h}(\mathbf{x})=\mathbf{y}_{C+1}$, otherwise,  ${\bm h}'(\mathbf{x})={\bm h}(\mathbf{x})$; and
\begin{equation*}   \int_{\mathcal{X}\times \mathcal{Y}} \ell(\phi\circ \tilde{\bm h}(\mathbf{x}), \phi(\mathbf{y})){\rm d} {P}_{Y|X}(\mathbf{y}|\mathbf{x}){\rm d} \tilde{P}(\mathbf{x})=0,
\end{equation*}
where $\phi$ is a function defined over $\mathcal{Y}$ and defined as follows
 $\phi(\mathbf{y})=\mathbf{y}_{C+1}$, if $\mathbf{y}=\mathbf{y}_{C+1}$;{ otherwise,} $\phi(\mathbf{y})=\mathbf{y}_1$.
\end{Definition*}
\begin{remark}
Assumption \ref{realization} implies that the hypothesis space $\mathcal{H}$ is complexity enough so that the unknown classes can be classified perfectly by many hypothesis functions. The assumption can be regarded as the open-set version of realization assumption \cite{shalev2014understanding,10.5555/2371238}. Realization assumption is a basic concept in learning theory.
\end{remark}
\begin{theorem}\label{T2}
Given a loss $\ell$ satisfying $\ell(\mathbf{y},\mathbf{y}')=0$ iff $\mathbf{y}=\mathbf{y}'$, and  a hypothesis space $\mathcal{H}\subset \{{\bm h}:\mathcal{X}\rightarrow \mathbb{R}^{C+1} \}$ satisfying Assumption \ref{realization}, if $Q_{X,Y}$ is the auxiliary domain for $P_{X,Y}$ and assume $P_X \ll Q_X \ll \tilde{P}$, where $\tilde{P}$ is the distribution introduced in Assumption \ref{realization}, then for any $0<\alpha<1$,
\begin{equation*}
\begin{split}
\min_{{\bm h}\in \mathcal{H}} R_Q^{\alpha}({\bm h})&={\min_{{\bm h}\in \mathcal{H}}}R_P^{\alpha}({\bm h}),
\\    \argmin_{{\bm h}\in \mathcal{H}} R_Q^{\alpha}({\bm h})&\subset {\argmin_{{\bm h}\in \mathcal{H}}}R_P^{\alpha}({\bm h}).
  \end{split}
\end{equation*}
\end{theorem}

\subsection{Construction of Ideal Auxiliary Domain}
\label{sec:contructQ}
As mentioned above, the auxiliary domain plays an important role to address the open-set learning problem from a transfer learning perspective. Thus, in this subsection, we first show how to construct an ideal auxiliary domain and then demonstrate how to estimate the ideal auxiliary domain via finite samples.
Given an auxiliary distribution $U$ such that $P_{X|Y\in \mathcal{Y}_k}\ll U$, we denote $r(\mathbf{x})$ as the density ratio between $P_{X|Y\in \mathcal{Y}_k}$ and $U$, i.e., for any $U$-measurable set $A$,
\begin{equation*}
    P_{X|Y\in \mathcal{Y}_k}(A)=\int_A r(\mathbf{x}) {\rm d} U(\mathbf{x}),
\end{equation*}
and denote ${Q}_U^{0,\beta}$ as the marginal distribution defined over $\mathcal{X}$, i.e.,  for any $U$-measurable set $A$,
\begin{equation}\label{Eq::ideal Dis}
    {Q}_U^{0,\beta}(A):= \gamma \int_{A} L_{0,\beta}(r(\mathbf{x})) {\rm d} U(\mathbf{x}),~~\textrm{here}
\end{equation}
\begin{equation}\label{Eq::gamma}
    \gamma = \frac{1}{1+\beta U(r=0)},
\end{equation}
\begin{equation*}
L_{0,\beta}(x)=\left\{
\begin{aligned}
~~~~~x+\beta,&~~~~~~~ x\leq 0, \\
~~~~~x,&~~~~~~~x > 0,\\
\end{aligned}
\right.
\end{equation*}
and $\beta>0$ is a parameter to tune the density of ${Q}_U^{0,\beta}$ for unknown classes.
Then we define the ideal auxiliary domain.
\begin{Definition}[Ideal Auxiliary Domain (IAD)]
\label{def:IAD}
Given the distribution $P_{X,Y}$ defined in Problem~\ref{def:problem_setting} and an auxiliary distribution $U$ defined over $\mathcal{X}$ such that $P_{X|Y\in \mathcal{Y}_k}\ll U$, then the ideal auxiliary domain regarding to $P_{X,Y}$ is
\begin{equation*}
   {Q}_U^{0,\beta}\cdot P_{Y|X},
\end{equation*}
where ${Q}_U^{0,\beta}$ is defined in Eq. (\ref{Eq::ideal Dis}).
\end{Definition}

In Definition \ref{def:IAD}, the probability value of distribution $Q_U^{0,\beta}$ in space $\mathcal{X} /\ {\rm supp}~r$ is a constant $\beta$. In detail, if $P_{X,Y}$ has no overlap between known and unknown classes, any sample from $Q_{X|Y=\mathbf{y}_{C+1}}$ shares same probability (see Figure \ref{fig:distribution}). In addition, an auxiliary distribution $U$ satisfying $P_{X|Y\in \mathcal{Y}_k}\ll U$ is needed. The samples drawn from $U$ can be generated by a gaussian distribution or uniform distribution with suitable support set. 

Given finite samples $T:=\{\tilde{\mathbf{x}}^j\}_{j=1}^m$ drawn (i.i.d.) from a given distribution $U$ as introduced in Definition \ref{def:IAD}. We introduce how to use $T$ and $S$ to construct an approximate form of ${Q}_U^{0,\beta}$ introduced in Definition \ref{def:IAD}.

To simple, we provide a mild assumption as follows.

\begin{Definition*}\label{Ass2}
Distributions $P_{X|Y\in \mathcal{Y}_k}$ and $U$ introduced in Definition \ref{def:IAD} are continuous distributions with density functions $p(\mathbf{x})$ and $q(\mathbf{x})$, respectively.
\end{Definition*}

\begin{remark}
The assumption that $P_{X|Y\in \mathcal{Y}_k}$ and $U$ are continuous can be replaced by a weaker assumption: $P_{X|Y\in \mathcal{Y}_k}$, $U\ll \mu$, where $\mu$ is a measure defined over $\mathcal{X}$. With the weaker assumption, all theorems still hold.
\end{remark}

Note that the density ratio $r=p/q$ required in ${Q}_U^{0,\beta}$ is unknown. To compute the density ratio $r$ using $S$ and $T$, the density ratio estimation methods are indispensable. Considering the property of statistical convergence,
we use \textit{kernelized variant of unconstrained least-squares importance fitting} (KuLSIF) \cite{DBLP:journals/ml/KanamoriSS12} to estimate the density ratio in the theoretical part: given RKHS space $\mathcal{H}_K$,
 \begin{equation}\label{KuLSIF}
     \min_{w\in \mathcal{H}_K} \frac{\sum_{\mathbf{x}\in T}w^2(\mathbf{x})}{m} -2 \frac{\sum_{(\mathbf{x},\mathbf{y})\in S}w(\mathbf{x})}{n}+\lambda \|w\|_k^2,
 \end{equation}
 where $\lambda$ is the regularization parameter.
 Then, we assume $\widehat{w}$ is the solution of Eq. (\ref{KuLSIF}).

After instance re-weighting, we regard the following measure
\begin{equation}\label{New Dis}
\widehat{Q}_T^{\tau,\beta}:= \frac{\gamma}{m}\sum_{\mathbf{x}\in T} L_{\tau,\beta}({\widehat{w}(\mathbf{x}})) \delta_{\mathbf{x}},~~~
\end{equation}
as the approximation of $Q_{U}^{0,\beta}$, where $\gamma$ is defined in Eq. (\ref{Eq::gamma}),
\begin{equation*}
L_{\tau,\beta}(x)=\left\{
\begin{aligned}
~~~~~x+\beta,&~~~~~~~ x\leq \tau; \\
~~~~~x,&~~~~~~~2\tau \leq x;\\
~~~~~\big(1-\frac{\beta}{\tau}\big)x+2\beta,&~~~~~~~\tau<x<2\tau,
\end{aligned}
\right.
\end{equation*}
and $\tau > 0$ is the threshold to select whether a sample $\mathbf{x}\in T$ is from unknown classes or known classes.



\subsection{Empirical Estimation for IAD Risk}

In this subsection, we first set the ideal auxiliary domain ${Q}_U^{0,\beta}\cdot P_{Y|X}$ as $Q_{X,Y}$, then we analyze the IAD risk $R^{\alpha}_Q({\bm h})$ from an approximate view, where $\alpha =1-1/\big(1+\beta U(r=0)\big)$. In detail, the IAD risk $R^{\alpha}_Q({\bm h})$ can be written as follows
\begin{equation}\label{IAD risk}
R_{Q}^{\alpha}({\bm h})= \int_{\mathcal{X}} \ell({\bm h}(\mathbf{x}),\mathbf{y}) {\rm d}P_{Y|X}(\mathbf{y}|\mathbf{x}) {\rm d}{Q}_U^{0,\beta}(\mathbf{x}). 
\end{equation}
Then, we use $\widehat{Q}_T^{\tau,\beta}$ (see Eq. (\ref{New Dis})) to construct auxiliary risk to approximate the IAD risk.
\begin{Definition}[Auxiliary Risk]
\label{def:Auxiliary Risk}
Given  samples $S$ with size $n$ drawn from $P_{X,Y|Y\in \mathcal{Y}_k}$ and $T$ with size $m$ drawn from $U$, i.i.d., then the auxiliary risk for a hypothesis function ${\bm h}$ is
\begin{equation}\label{eq::empirical risk}
 \widehat{R}_{S,T}^{\tau,\beta}({\bm h}):=\widehat{R}_{S}({\bm h})+  \Delta_{S,T}^{\tau,\beta}({\bm h}),
\end{equation}
where
\begin{equation*}
\begin{split}
&\widehat{R}_{S}({\bm h}):=\frac{1}{n}\sum_{(\mathbf{x},\mathbf{y})\in S} \ell({\bm h}({\mathbf{x}}), \mathbf{y}),
\\ &  \Delta_{S,T}^{\tau,\beta}({\bm h}):=\max\{ \widehat{R}^{\tau,\beta}_T({\bm h},\mathbf{y}_{C+1})- \widehat{R}_{S}({\bm h},\mathbf{y}_{C+1}),0\},
\\& \widehat{R}^{\tau,\beta}_T({\bm h},\mathbf{y}_{K+1}):= \frac{1}{\gamma}\int_{\mathcal{X}} \ell({\bm h}(\mathbf{x}), \mathbf{y}_{C+1}){\rm d} \widehat{Q}_T^{\tau,\beta}(\mathbf{x})
\\ &~~~~~~~~~~~~~~~~~~~~~~~~~~ = \frac{1}{m}\sum_{\mathbf{x}\in T} L_{\tau,\beta}({\widehat{w}(\mathbf{x}})) \ell({\bm h}({\mathbf{x}}), \mathbf{y}_{C+1}), \\& \widehat{R}_{S}({\bm h},\mathbf{y}_{C+1}):=\frac{1}{n}\sum_{(\mathbf{x},\mathbf{y})\in S} \ell({\bm h}({\mathbf{x}}), \mathbf{y}_{C+1}),
\end{split}
\end{equation*}
\vspace{-0.2cm}
here $\widehat{Q}_T^{\tau,\beta}$ is defined in Eq. (\ref{New Dis}) and $\gamma$ is defined in Eq. (\ref{Eq::gamma}).
\end{Definition}

Theorem \ref{T3} implies that $(1-\alpha) \widehat{R}_{S,T}^{\tau,\beta}({\bm h})$ can approximate $R^{\alpha}_Q({\bm h})$ uniformly.
 \begin{theorem}\label{T3}
 Assume assumption \ref{Ass2} holds, the feature space $\mathcal{X}$ is compact and the hypothesis space $\mathcal{H}\subset \{{\bm h}:\mathcal{X}\rightarrow \mathbb{R}^{C+1} \}$ has finite Natarajan dimension \cite{shalev2014understanding}. Let the RKHS $\mathcal{H}_K$ be the Hilbert space with gaussian kernel. Suppose that loss function is bounded by $c$, the density $r\in \mathcal{H}_K$ and set the regularization parameter $\lambda=\lambda_{n,m}$ in {\rm KuLSIF} (see Eq. (\ref{KuLSIF})) such that
 \begin{equation*}
   \lim_{n,m\rightarrow 0}  \lambda_{n,m} =0,~~~\lambda_{n,m}^{-1}=O({\min \{n,m\}}^{1-\delta}),
 \end{equation*}
 \vspace{-0.1cm}
 where $0<\delta<1$ is any constant, then for any $0\leq \alpha<1$,
\begin{equation*}
 \begin{split}
&\sup_{{\bm h}\in \mathcal{H}}|(1-\alpha) \widehat{R}_{S,T}^{\tau,\beta}({\bm h})-R^{\alpha}_Q({\bm h})|
\\ \leq &c\big(\max\{1,\frac{\beta}{\tau}\}+\beta\big) O_p(\lambda^{\frac{1}{2}}_{n,m})+\gamma c \beta U(0<r\leq 2\tau),
\end{split}
\end{equation*}
where $O_p$ denotes the probabilistic order, $\gamma=1-\alpha, \beta=\frac{\alpha}{\gamma U(r=0)}$, $\widehat{R}_{S,T}^{\tau,\beta}({\bm h})$ is defined in Eq. (\ref{eq::empirical risk}), and $R^{\alpha}_Q({\bm h})$ is the IAD risk defined in Eq. (\ref{IAD risk}).
 \end{theorem}

Note that $U(0<r\leq 2\tau)\rightarrow 0$, if $\tau\rightarrow 0$, and Theorem \ref{T3} has indicated that if we omit the term $(1-\alpha) c \beta U(0<p/q\leq 2\tau)$ and set $m\geq n$, the gap between $(1-\alpha) \widehat{R}_{S,T}^{\tau,\beta}({\bm h})$ and $R^{\alpha}_Q({\bm h})$ is close to $O_p(1/\sqrt{n})$ by choosing a small $\delta$.

\subsection{Main Theoretical Results}
In this subsection, we analyze the relationship between $R^{\alpha}_P({\bm h})$ and $\widehat{R}_{S,T}^{\tau,\beta}({\bm h})$ based on Theorems~\ref{T1}, \ref{T2} and \ref{T3}.

\begin{theorem}[Uniform Bound Based on Transfer Learning]\label{T4} 
Given  the  same  conditions  and  assumptions  in Theorems $\ref{T1}$ and $\ref{T3}$, then for any $0\leq \alpha<1$, ${\bm h}\in \mathcal{H}$,
\begin{equation*}
 \begin{split}
&|(1-\alpha)\widehat{R}_{S,T}^{\tau,\beta}({\bm h})-R^{\alpha}_P({\bm h})|
\\ \leq &c\big(\max\{1,\frac{\beta}{\tau}\}+\beta\big) O_p(\lambda^{\frac{1}{2}}_{n,m})+\gamma c \beta U(0<r\leq 2\tau)\\ + &  \alpha  d^{\ell}_{{\bm h},\mathcal{H}}(P_{X|Y=\mathbf{y}_{C+1}},Q_{X|Y=\mathbf{y}_{C+1}})+\alpha \Lambda,
\end{split}
\end{equation*}
where $\lambda_{n,m}$ is defined in Theorem $\ref{T3}$, $\gamma=1-\alpha,\beta=\frac{\alpha}{\gamma U(r=0)}$, $\widehat{R}_{S,T}^{\tau,\beta}({\bm h})$ is defined in Eq. (\ref{eq::empirical risk}), $d^{\ell}_{{\bm h},\mathcal{H}}$ is the disparity discrepancy defined in Definition \ref{doubleloss}, $\Lambda$ is the combined risk defined in Eq. (\ref{combined risk}) and $O_p$ is the probabilistic order (independent of $c,\beta,\tau$ and $\alpha$).
\end{theorem}

Theorem \ref{T4} indicates that the gap between $(1-\alpha)\widehat{R}_{S,T}^{\tau,\beta}({\bm h})$ and $R^{\alpha}_P({\bm h})$ is controlled by four special terms. The combined risk $\Lambda$ and domain discrepancy for unknown classes can be regarded as constants. The other two terms could be small enough, if $n,m \rightarrow +\infty$ and $\tau$ is a small value.

\begin{theorem}[Estimation Error for OSL]\label{T5}
Given  the  same  conditions  and  assumptions  in Theorems $\ref{T2}$ and $\ref{T3}$, for any $0\leq \alpha<1$, if we assume
$
   \widehat{\bm h}\in  \argmin_{{\bm h}\in \mathcal{H}} \widehat{R}_{S,T}^{\tau,\beta}({\bm h}),
$
then $|R_P^{\alpha}(\widehat{\bm h}) -\min_{{\bm h}\in \mathcal{H}}R^{\alpha}_P({\bm h})|$ has an upper bound
\begin{equation*}
 \begin{split}
&c\big(\max\{1,\frac{\beta}{\tau}\}+\beta\big) O_p(\lambda^{\frac{1}{2}}_{n,m})+4\gamma c \beta U(0<r\leq 2\tau),
\end{split}
\end{equation*}
where $\lambda_{n,m}$ is defined in Theorem $\ref{T3}$, $\gamma=1-\alpha, \beta=\frac{\alpha}{\gamma U(r=0)}$, $\widehat{R}_{S,T}^{\tau,\beta}({\bm h})$ is defined in Eq. (\ref{eq::empirical risk}) and $O_p$ is the probabilistic order (independent of $c,\beta,\tau$ and $\alpha$).
\end{theorem}

If we select a small $\tau$ to make $U(0<r\leq 2\tau)$ small enough and set $m\geq n$, then under some assumptions, the following optimization problem
\begin{equation}\label{optimial problem}
\min_{{\bm h}\in \mathcal{H}} \widehat{R}_{S,T}^{\tau,\beta}({\bm h})
\end{equation}
is  almost {\textit{classifier-consistent}} $\footnote{The learned classifier by the algorithm is infinite-samples consistent to $\argmin_{{\bm h}\in \mathcal{H}} R^{\alpha}_P({\bm h})$.}$ with estimation error close to $O_p(1/\sqrt{n})$. Additionally, the weight estimation in Theorem \ref{T5} is crucial. To weaken the effect of weight estimation in area ${\rm supp}~P_{X|Y\in \mathcal{Y}_k}$, we introduce a proxy for $\widehat{R}_{S,T}^{\tau,\beta}({\bm h})$.
\begin{Definition}[Proxy of Auxiliary Risk]
\label{def:Auxiliary Risk2}
Given  samples $S$ with size $n$ drawn from $P_{X,Y|Y\in \mathcal{Y}_k}$ and $T$ with size $m$ drawn from $U$, i.i.d., then the auxiliary risk for a hypothesis function ${\bm h}$ is
\begin{equation}\label{eq::proxy for empirical risk}
 \widetilde{R}_{S,T}^{\tau,\beta}({\bm h}):=\widehat{R}_{S}({\bm h})+ \frac{\alpha \gamma'}{1-\alpha} \widehat{R}_{S,T,u}^{\tau,\beta}({\bm h}),
\end{equation}
where $\gamma'= 1/ U(r=0)$, $\widehat{R}_{S}({\bm h})$ is defined in Definition \ref{def:Auxiliary Risk},
\begin{equation*}
\begin{split}
 \widehat{R}_{S,T,u}^{\tau,\beta}({\bm h}) :=\frac{1}{m}\sum_{\mathbf{x}\in T} L_{\tau,\beta}^{-}({\widehat{w}(\mathbf{x}})) \ell({\bm h({\mathbf{x}})}, \mathbf{y}_{C+1}),
\end{split}
\end{equation*}
here  \begin{equation*}
L_{\tau,\beta}^{-}(x)=\left\{
\begin{aligned}
x+\beta,&~~~~ x\leq \tau; \\
0,&~~~~2\tau \leq x;\\
-\frac{\tau+\beta}{\tau}x+2\tau+2\beta,&~~~~\tau<x<2\tau.
\end{aligned}
\right.
\end{equation*}
\end{Definition}
\vspace{-0.2cm}
Then, a result similar to Theorem \ref{T5} for auxiliary risk $\widetilde{R}_{S,T}^{\tau,\beta}$ is given as follows.
\begin{theorem}[Estimation Error for OSL]\label{T6}
Given  the  same  conditions  and  assumptions  in Theorem $\ref{T5}$, for any $0\leq \alpha<1$, if we assume $\widetilde{\bm h}\in  \argmin_{{\bm h}\in \mathcal{H}} \widetilde{R}_{S,T}^{\tau,\beta}({\bm h})$,
then $|R_P^{\alpha}(\widetilde{\bm h}) -\min_{{\bm h}\in \mathcal{H}}R^{\alpha}_P({\bm h})|$ has an upper bound
\begin{equation*}
 \begin{split}
&c\gamma'\big(1+\tau+\frac{\beta}{\tau}+\beta\big) O_p(\lambda^{\frac{1}{2}}_{n,m})+4c\gamma' \alpha \beta U(0<r\leq 2\tau),
\end{split}
\end{equation*}
where $\lambda_{n,m}$, $\beta$ are introduced in Theorem \ref{T5}, $\gamma'$ and $\widetilde{R}_{S,T}^{\tau,\beta}({\bm h})$ are defined in Definition \ref{def:Auxiliary Risk2}, and $O_p$ is the probabilistic order (independent of $c,\beta,\tau, \gamma'$ and $\alpha$).
\end{theorem}

\section{A Principle Guided OSL Algorithm}\label{section4}
Inspired by Theorem \ref{T6}, we focus on the following problem
 \begin{equation}\label{optimial problem1}
\min_{{\bm \Theta}}\big ( \widehat{R}_{S}({\bm h}_{\bm \Theta})  + \mu \widehat{R}_{S,T,u}^{\tau,\beta}({\bm h}_{\bm \Theta}) \big ),
\end{equation}
where $\mu$ is a positive parameter, $\widehat{R}_{S,T,u}^{\tau,\beta}({\bm h})$ is defined in Eq. \eqref{eq::proxy for empirical risk}, ${\bm h}_{\bm \Theta}$ is a hypothesis function based on a neural network, and ${\bm \Theta}$ is parameters of the neural network.
To optimize ${\bm h}_{\bm \Theta}$ to solve the minimization problem defined in Eq.~\eqref{optimial problem1}, we have the following five steps.

\input{algorihtm}

\input{sec5}

\section{Discussion}
\textbf{Relation with Generative Models.} Algorithms based on generative models are the mainstream for OSL. CGDL \cite{DBLP:journals/corr/abs-2003-08823}, C2AE \cite{DBLP:conf/cvpr/OzaP19} and Counterfactual \cite{DBLP:conf/eccv/NealOFWL18} are the representative works based on generative models. AOSR can be regarded as the weight-based generative model, but is 
very different from the mainstream generative model-based algorithms (feature map-based generative model \cite{DBLP:journals/corr/abs-2003-08823,DBLP:conf/cvpr/OzaP19,DBLP:conf/eccv/NealOFWL18}). Form the theoretical perspective, it is necessary to develop theory to guarantee the generalization ability of feature map-based generative models. Here we propose an interesting and important problem: \textit{how to develop generalization theory for  feature map-based generative models under open-set assumption ?}

\textbf{Relation with PU Learning.} Positive-unlabeled learning (PU learning) \cite{DBLP:conf/nips/NiuPSMS16} is a special binary classification task, which assumes only unlabeled samples and positive samples (i.e., samples with positive labels) are available. Our theory is deeply related to PU learning. If we regard the known samples $S$ and the auxiliary samples as the positive samples and the unlabeled samples, respectively. Then, our theory degenerates into the PU learning theory.

\textbf{Remaining Problems in OSL Theory.} We list several interesting and important problems for OSL theory as follows.
\textit{1. How to construct weaker assumption to replace assumption \ref{realization} for achieving similar results ? } 
\\
\textit{2. Without assumption \ref{realization}, what will happen  ? } 
\\
\textit{3. Is it possible for OSL to achieve agnostic PAC learnability and achieve  fast learning rate $O_p(1/{n}^{a})$, for $a>0.5$ ? } 
\\
\textit{4. Is it possible to construct OSL learning theory by  stability theory \cite{DBLP:journals/jmlr/BousquetE02} ? } 
\section{Conclusion and Future Work}
This paper mainly focuses on the learning theory for open-set learning. The generalization error bounds proved in our work provide the first almost-PAC-style guarantee on open-set learning.
Based on our theory, a principle guided algorithm AOSR is proposed. Experiments on real datasets indicate that AOSR achieves competitive performance when compared with  baselines. In future, we will focus on developing more powerful OSL algorithms based on our theory and dynamic weight 
technique \cite{DBLP:conf/nips/FangL0S20}. With the dynamic weight , we can update the weight for each epoch and make a better integration between instance-weighting and deep learning.
\vspace{-0.1 cm}
\section*{Acknowledgments}
\vspace{-0.1 cm}
The  work introduced in this paper   was  supported  by  the Australian Research Council (ARC) under FL190100149. 


\bibliographystyle{icml2021}


\bibliography{Main.bbl}
\end{document}


\onecolumn

\icmltitle{Appendix: Learning Bounds for Open-Set Learning}

$\bullet$ Appendix A recalls some important definitions and concepts. 

$\bullet$ Appendix B provides the proof for Theorem 1.

$\bullet$ Appendix C provides the proof for Theorem 2. 

$\bullet$ Appendix D provides the proof for Theorem 3.

$\bullet$ Appendix E provides the proofs for Theorems 4, 5 and 6.

$\bullet$ Appendix F provides  details on datasets and parameter analysis.
\newpage

\section{Appendix A: Notations and Concepts}\label{AppA}
In this section, we introduce the definition of open-set learning and then introduce important concepts used in this paper.

Let $\mathcal{X}\subset \mathbb{R}^d$ be a feature space and $\mathcal{Y}:=\{\mathbf{y}_c\}_{c=1}^{C+1}$ be the label space, where the label $\mathbf{y}_c$ is a one-hot vector whose $c$-th coordinate is $1$ and the other coordinate is $0$.
\begin{Definition}[Domain, Known and Unknown Classes.]\label{Domain}
Given random variable $X\in \mathcal{X}$ and $Y\in \mathcal{Y}$, a domain is a joint distribution $P_{X,Y}$. The classes from $\mathcal{Y}_k:=\{\mathbf{y}_c\}_{c=1}^{C}$ is called known class and $\mathbf{y}_{C+1}$ is called unknown classes.
\end{Definition}

The open set learning problem is defined as follows.
\begin{definition}[Open-Set Learning]
Given independent and identically distributed (i.i.d.) samples  $S=\{(\mathbf{x}^i,\mathbf{y}^i)\}_{i=1}^n$ drawn from $P_{X,Y|Y\in \mathcal{Y}_k}$. Aim of open-set learning  is  to  train  a  classifier using  $S$ such  that $f$ can  classify
1)  the  sample from  known  classes into correct known classes; 
2) the sample from unknown classes into unknown classes.
\end{definition}

\begin{table}[h]
\caption{{Main notations and their descriptions.}}
\begin{center}
\normalsize
\begin{tabular}{p{6.5cm}p{9.5cm}}
\hline
Notation & ~~~~~~~~~~Description  \\ \hline
$\mathcal{X}$, $\mathcal{Y}=\{\mathbf{y}_i\}_{i=1}^{C+1}$, $\mathcal{Y}_k=\{\mathbf{y}_i\}_{i=1}^{C}$&  feature space, label space, label space for known classes\\ $X$, $Y$
& random variables on the feature space $\mathcal{X}$ and $\mathcal{Y}$\\$P_{X,Y},~Q_{X,Y}$& joint distributions\\$P_{X},~Q_{X}$&marginal distributions\\$P_{X,Y|Y\in \mathcal{Y}_k},Q_{X,Y|Y\in \mathcal{Y}_k}$& conditional distributions when label belongs to known classes\\$P_{X|Y= \mathbf{y}_{C+1}},Q_{X|Y= \mathbf{y}_{C+1}}$& conditional distributions when label belongs to unknown classes\\  $R_P^{\alpha}$, $R_Q^{\alpha}$ & $\alpha$-risks corresponding to $P_{X,Y},Q_{X,Y}$\\$R_{P,k}$, $R_{Q,k}$ & partial risks for known classes corresponding to $P_{X,Y},Q_{X,Y}$\\$R_{P,u}$, $R_{Q,u}$ & partial risks for unknown classes corresponding to $P_{X,Y},Q_{X,Y}$\\${\bm h}$ & hypothesis function from $\mathcal{X} \rightarrow \mathbb{R}^{C+1}$ \\ $\mathcal{H}$ & hypothesis space, a subset of $\{{\bm h}:\mathcal{X} \rightarrow \mathbb{R}^{C+1} \}$ \\  $\mathcal{H}_K$ & RKHS with kernel $K$\\ $U$& auxiliary distribution defined over $\mathcal{X}$\\ $Q_U^{0,\beta}P_{Y|X}$& ideal auxiliary domain defined over $\mathcal{X}\times \mathcal{Y}$\\$\widehat{Q}^{\tau,\beta}_{U}$ & the approximation of $Q_U^{0,\beta}$\\$w$ & weights\\$S,T$& samples drawn from $P_{X,Y}$ and $Q_X$, respectively\\$n,m$& sizes of samples $S$ and $T$\\ $d_{{\bm h},\mathcal{H}}^{\ell}, \Lambda$& disparity discrepancy, combined risk\\ $\widehat{R}_{S,T}^{\tau,\beta}$,$\widetilde{R}_{S,T}^{\tau,\beta}$& auxiliary risk, proxy of auxiliary risk\\ 
\hline
\end{tabular}
\end{center}
\end{table}
\newpage
\section{Appendix B: Proof of Theorem 1}\label{AppB}
\begin{proof}[Proof of Theorem. 1]
\begin{equation*}
\begin{split}
   & |R_P^{\alpha}({\bm h})-R_Q^{\alpha}({\bm h})|=|(1-\alpha) R_{P,k}({\bm h})+\alpha R_{P,u}({\bm h})-(1-\alpha) R_{Q,k}({\bm h})-\alpha R_{Q,u}({\bm h})|
 \\ = &\big |(1-\alpha) \int_{\mathcal{X}\times \mathcal{Y}_k}\ell({\bm h}(\mathbf{x}),\mathbf{y}) {\rm d}P_{X,Y|Y\in \mathcal{Y}_k}(\mathbf{x},\mathbf{y})+\alpha R_{P,u}({\bm h})-(1-\alpha) \int_{\mathcal{X}\times \mathcal{Y}_k}\ell({\bm h}(\mathbf{x}),\mathbf{y}) {\rm d}Q_{X,Y|Y\in \mathcal{Y}_k}(\mathbf{x},\mathbf{y})-\alpha R_{Q,u}({\bm h})\big |\\ = &\alpha \big | R_{P,u}({\bm h})-R_{Q,u}({\bm h}) \big |~~~{\rm we~~have~~used~~} Q_{X,Y|Y\in \mathcal{Y}_k}=P_{X,Y|Y\in \mathcal{Y}_k}\\ = & \alpha \big |\int_{\mathcal{X}}\ell({\bm h}(\mathbf{x}),\mathbf{y}_{C+1}) {\rm d}P_{X|Y=\mathbf{y}_{C+1}}(\mathbf{x})-\int_{\mathcal{X}}\ell({\bm h}(\mathbf{x}),\mathbf{y}_{C+1}) {\rm d}Q_{X|Y=\mathbf{y}_{C+1}}(\mathbf{x})\big |\\ \leq &\alpha \big |\int_{\mathcal{X}}\ell({\bm h}(\mathbf{x}),{\bm h}'(\mathbf{x})) {\rm d}P_{X|Y=\mathbf{y}_{C+1}}(\mathbf{x})-\int_{\mathcal{X}}\ell({\bm h}(\mathbf{x}),{\bm h}'(\mathbf{x})) {\rm d}Q_{X|Y=\mathbf{y}_{C+1}}(\mathbf{x})\big |\\+&\alpha \int_{\mathcal{X}}\ell({\bm h}'(\mathbf{x}),\mathbf{y}_{C+1}) {\rm d}P_{X|Y=\mathbf{y}_{C+1}}(\mathbf{x})+\alpha \int_{\mathcal{X}}\ell({\bm h}'(\mathbf{x}),\mathbf{y}_{C+1}) {\rm d}Q_{X|Y=\mathbf{y}_{C+1}}(\mathbf{x})~~~{\rm the~~triangle ~~inequality~~is~~used}\\ \leq & \alpha d^{\ell}_{{\bm h},\mathcal{H}}(P_{X|Y=\mathbf{y}_{C+1}},Q_{X|Y=\mathbf{y}_{C+1}})+\alpha \int_{\mathcal{X}}\ell({\bm h}'(\mathbf{x}),\mathbf{y}_{C+1}) {\rm d}P_{X|Y=\mathbf{y}_{C+1}}(\mathbf{x})+\alpha \int_{\mathcal{X}}\ell({\bm h}'(\mathbf{x}),\mathbf{y}_{C+1}) {\rm d}Q_{X|Y=\mathbf{y}_{C+1}}(\mathbf{x}).
\end{split}
\end{equation*}
Hence,
\begin{equation*}
\begin{split}
&|R_P^{\alpha}({\bm h})-R_Q^{\alpha}({\bm h})|=\min_{{\bm h}'\in \mathcal{H}}|R_P^{\alpha}({\bm h})-R_Q^{\alpha}({\bm h})|{\rm~Note~ that~ we~ minimize~} {\bm h}',~{\rm but~ not~} {\bm h}\\ \leq & \min_{{\bm h}'\in \mathcal{H}}\big (\alpha d^{\ell}_{{\bm h},\mathcal{H}}(P_{X|Y=\mathbf{y}_{C+1}},Q_{X|Y=\mathbf{y}_{C+1}})+\alpha \int_{\mathcal{X}}\ell({\bm h}'(\mathbf{x}),\mathbf{y}_{C+1}) {\rm d}P_{X|Y=\mathbf{y}_{C+1}}(\mathbf{x})+\alpha \int_{\mathcal{X}}\ell({\bm h}'(\mathbf{x}),\mathbf{y}_{C+1}) {\rm d}Q_{X|Y=\mathbf{y}_{C+1}}(\mathbf{x}) \big )\\ \leq &\alpha d^{\ell}_{{\bm h},\mathcal{H}}(P_{X|Y=\mathbf{y}_{C+1}},Q_{X|Y=\mathbf{y}_{C+1}})+\alpha \Lambda.
\end{split}
\end{equation*}
\end{proof}

\newpage
\section{Appendix C: Proof of Theorem 2}\label{AppC}
\begin{proof}[Proof of Theorem 2]
\textbf{Step 1.} Note that
\begin{equation*}
     \int_{\mathcal{X}\times \mathcal{Y}} \ell(\phi\circ \tilde{\bm h}(\mathbf{x}), \phi(\mathbf{y})){\rm d} {P}_{Y|X}(\mathbf{x}){\rm d} \tilde{P}(\mathbf{x})=0,
\end{equation*}
hence, if we set $\tilde{P}_{X,Y}=\tilde{P}P_{Y|X}$, then
\begin{equation*}
\begin{split}
  &   \int_{\mathcal{X}\times \mathcal{Y}} \ell(\phi\circ \tilde{\bm h}(\mathbf{x}), \phi(\mathbf{y})){\rm d} \tilde{P}_{X,Y|Y\in \mathcal{Y}_k}(\mathbf{x},\mathbf{y})=0,~~~
      \int_{\mathcal{X}} \ell(\phi\circ \tilde{\bm h}(\mathbf{x}), \phi(\mathbf{y}_{C+1})){\rm d} \tilde{P}_{X|Y=\mathbf{y}_{C+1}}(\mathbf{x})=0.
     \end{split}
\end{equation*}

Note that $\ell(\mathbf{y},\mathbf{y}')=0$ iff $\mathbf{y}=\mathbf{y}'$, hence, $\tilde{\bm h}(\mathbf{x})= \mathbf{y}_{C+1}$, for $\mathbf{x}\in {\rm supp}~\tilde{P}_{X|Y=\mathbf{y}_{C+1}}$ a.e. $\tilde{P}$ and $\tilde{\bm h}(\mathbf{x})\neq \mathbf{y}_{C+1}$, for $\mathbf{x}\in {\rm supp}~\tilde{P}_{X|Y\in \mathcal{Y}_{k}}$ a.e. $\tilde{P}$.

\textbf{Step 2.} Because ${P}_X \ll {Q}_X \ll \tilde{P}$,
then,
\begin{equation*}
   {\rm supp}~\tilde{P}_{X|Y\in \mathcal{Y}_k} \supset {\rm supp}~{Q}_{X|Y\in \mathcal{Y}_k}\supset {\rm supp}~{P}_{X|Y\in \mathcal{Y}_k} 
\end{equation*}
and
\begin{equation*}
   {\rm supp}~\tilde{P}_{X|Y=\mathbf{y}_{C+1}} \supset {\rm supp}~{Q}_{X|Y=\mathbf{y}_{C+1}}\supset {\rm supp}~{P}_{X|Y=\mathbf{y}_{C+1}}.
\end{equation*}

\textbf{Step 3.} We need to check that $\min_{{\bm h}\in \mathcal{H}}R_P^{\alpha}({\bm h})=(1-\alpha)\min_{{\bm h}\in \mathcal{H}}R_{P,k}({\bm h})$. First, it is clear that $\min_{{\bm h}\in \mathcal{H}}R_P^{\alpha}({\bm h})\geq (1-\alpha)\min_{{\bm h}\in \mathcal{H}}R_{P,k}({\bm h})$. If there exists ${\bm h}_P\in \mathcal{H}$ such that $\min_{{\bm h}\in \mathcal{H}}R_P^{\alpha}({\bm h})> (1-\alpha)\min_{{\bm h}\in \mathcal{H}}R_{P,k}({\bm h}_P)$. 

Set
\begin{equation*}
   \tilde{\bm h}_P(\mathbf{x})=\mathbf{y}_{C+1},~~{\rm if~~} \tilde{\bm h}(\mathbf{x})=\mathbf{y}_{C+1};~~{\rm otherwise,~~}\tilde{\bm h}_P(\mathbf{x})={\bm h}_P(\mathbf{x}),
\end{equation*}
hence, using the results of Step 1 and Step 2, we know
$
    \{\mathbf{x}:\tilde{\bm h}(\mathbf{x})=\mathbf{y}_{C+1}\}\supset {\rm supp}~{P}_{X|Y=\mathbf{y}_{C+1}}.
$
Then, 
\begin{equation*}
\begin{split}
&(1-\alpha)\int_{\mathcal{X}\times \mathcal{Y}_k}\ell({\bm h}_P(\mathbf{x}),\mathbf{y}) {\rm d}P_{X,Y|Y\in \mathcal{Y}_k}(\mathbf{x},\mathbf{y}) \\=&(1-\alpha)
\int_{\{{\rm supp}~P_{X|Y\in \mathcal{Y}_k}\}\times \mathcal{Y}_k}\ell({\bm h}_P(\mathbf{x}),\mathbf{y}) {\rm d}P_{X,Y|Y\in \mathcal{Y}_k}(\mathbf{x},\mathbf{y})\\=&
(1-\alpha)\int_{\{{\rm supp}~P_{X|Y\in \mathcal{Y}_k}\}\times \mathcal{Y}_k}\ell(\tilde{\bm h}_P(\mathbf{x}),\mathbf{y}) {\rm d}P_{X,Y|Y\in \mathcal{Y}_k}(\mathbf{x},\mathbf{y})~{\rm~have~used~} \tilde{\bm h}(\mathbf{x})\neq \mathbf{y}_{C+1}, ~{\rm for~}\mathbf{x}\in {\rm supp}~\tilde{P}_{X|Y\in \mathcal{Y}_k}~{\rm a.e.~}\tilde{P}\\ = &(1-\alpha)\int_{\{{\rm supp}~P_{X|Y\in \mathcal{Y}_k}\}\times \mathcal{Y}_k}\ell(\tilde{\bm h}_P(\mathbf{x}),\mathbf{y}) {\rm d}P_{X,Y|Y\in \mathcal{Y}_k}(\mathbf{x},\mathbf{y})+0\\ =& (1-\alpha)\int_{\{{\rm supp}~P_{X|Y\in \mathcal{Y}_k}\}\times \mathcal{Y}_k}\ell(\tilde{\bm h}_P(\mathbf{x}),\mathbf{y}) {\rm d}P_{X,Y|Y\in \mathcal{Y}_k}(\mathbf{x},\mathbf{y})+\alpha \int_{{\rm supp}~P_{X|Y=\mathbf{y}_{C+1}}}\ell(\tilde{\bm h}(\mathbf{x}),\mathbf{y}_{C+1}) {\rm d}P_{X|Y= \mathbf{y}_{C+1}}(\mathbf{x})\\ = &(1-\alpha)\int_{\{{\rm supp}~P_{X|Y\in \mathcal{Y}_k}\}\times \mathcal{Y}_k}\ell(\tilde{\bm h}_P(\mathbf{x}),\mathbf{y}) {\rm d}P_{X,Y|Y\in \mathcal{Y}_k}(\mathbf{x},\mathbf{y})+\alpha \int_{{\rm supp}~P_{X|Y=\mathbf{y}_{C+1}}}\ell(\tilde{\bm h}_P(\mathbf{x}),\mathbf{y}_{C+1}) {\rm d}P_{X|Y= \mathbf{y}_{C+1}}(\mathbf{x})\\ =& R^{\alpha}_P(\tilde{\bm h}_P)\geq \min_{{\bm h}\in \mathcal{H}}R^{\alpha}_P({\bm h}),
\end{split}
\end{equation*}
hence,
$\min_{{\bm h}\in \mathcal{H}}R_P^{\alpha}({\bm h})=(1-\alpha)\min_{{\bm h}\in \mathcal{H}}R_{P,k}({\bm h})$. Similarly, we can prove that $\min_{{\bm h}\in \mathcal{H}}R_Q^{\alpha}({\bm h})=(1-\alpha)\min_{{\bm h}\in \mathcal{H}}R_{Q,k}({\bm h})$.
 Because $Q_{X|Y\in \mathcal{Y}_k}=P_{X|Y\in \mathcal{Y}_k}$, hence, $\min_{{\bm h}\in \mathcal{H}}R_{Q,k}({\bm h})=\min_{{\bm h}\in \mathcal{H}}R_{P,k}({\bm h})$. 
 Using the results of Step 3, we obtain that
 \begin{equation}\label{Eqq:equalPQ}
\min_{{\bm h}\in \mathcal{H}}R_{Q}({\bm h})=\min_{{\bm h}\in \mathcal{H}}R_{P}({\bm h}).
\end{equation}
\\
\textbf{Step 4.} Given any ${\bm h}^*\in \argmin_{{\bm h}\in \mathcal{H}}R_P^{\alpha}({\bm h})$, then we construct $\tilde{\bm h}^*$ such that
\begin{equation*}
    \tilde{\bm h}^*(\mathbf{x})=\mathbf{y}_{C+1},~~{\rm if~~} \tilde{\bm h}(\mathbf{x})=\mathbf{y}_{C+1};~~{\rm otherwise,~~}\tilde{\bm h}^*(\mathbf{x})={\bm h}^*(\mathbf{x}).
\end{equation*}

It is clear that $\tilde{\bm h}^* \in \mathcal{H}$ according to Assumption 1.

Then,
\begin{equation*}
\begin{split}
&R_P^{\alpha}({\bm h}^*)
\\ \geq &(1-\alpha)\int_{\mathcal{X}\times \mathcal{Y}_k}\ell({\bm h}^*(\mathbf{x}),\mathbf{y}) {\rm d}P_{X,Y|Y\in \mathcal{Y}_k}(\mathbf{x},\mathbf{y}) \\=&(1-\alpha)
\int_{\{{\rm supp}~P_{X|Y\in \mathcal{Y}_k}\}\times \mathcal{Y}_k}\ell({\bm h}^*(\mathbf{x}),\mathbf{y}) {\rm d}P_{X,Y|Y\in \mathcal{Y}_k}(\mathbf{x},\mathbf{y})\\=&
(1-\alpha)\int_{\{{\rm supp}~P_{X|Y\in \mathcal{Y}_k}\}\times \mathcal{Y}_k}\ell(\tilde{\bm h}^*(\mathbf{x}),\mathbf{y}) {\rm d}P_{X,Y|Y\in \mathcal{Y}_k}(\mathbf{x},\mathbf{y})~{\rm~have~used~} \tilde{\bm h}(\mathbf{x})\neq \mathbf{y}_{C+1}, ~{\rm for~}\mathbf{x}\in {\rm supp}~\tilde{P}_{X|Y\in \mathcal{Y}_k}~{\rm a.e.~}\tilde{P}\\ = &(1-\alpha)\int_{\{{\rm supp}~P_{X|Y\in \mathcal{Y}_k}\}\times \mathcal{Y}_k}\ell(\tilde{\bm h}^*(\mathbf{x}),\mathbf{y}) {\rm d}P_{X,Y|Y\in \mathcal{Y}_k}(\mathbf{x},\mathbf{y})+0\\ =& (1-\alpha)\int_{\{{\rm supp}~P_{X|Y\in \mathcal{Y}_k}\}\times \mathcal{Y}_k}\ell(\tilde{\bm h}^*(\mathbf{x}),\mathbf{y}) {\rm d}P_{X,Y|Y\in \mathcal{Y}_k}(\mathbf{x},\mathbf{y})+\alpha \int_{{\rm supp}~P_{X|Y=\mathbf{y}_{C+1}}}\ell(\tilde{\bm h}(\mathbf{x}),\mathbf{y}_{C+1}) {\rm d}P_{X|Y= \mathbf{y}_{C+1}}(\mathbf{x})\\ = &(1-\alpha)\int_{\{{\rm supp}~P_{X|Y\in \mathcal{Y}_k}\}\times \mathcal{Y}_k}\ell(\tilde{\bm h}^*(\mathbf{x}),\mathbf{y}) {\rm d}P_{X,Y|Y\in \mathcal{Y}_k}(\mathbf{x},\mathbf{y})+\alpha \int_{{\rm supp}~P_{X|Y=\mathbf{y}_{C+1}}}\ell(\tilde{\bm h}^*(\mathbf{x}),\mathbf{y}_{C+1}) {\rm d}P_{X|Y= \mathbf{y}_{C+1}}(\mathbf{x})\\ =& R^{\alpha}_P(\tilde{\bm h}^*).
\end{split}
\end{equation*}
Hence,  for any ${\bm h}^*\in \argmin_{{\bm h}\in \mathcal{H}}R_P^{\alpha}({\bm h})$,
\begin{equation*}
    \int_{\mathcal{X}}\ell({\bm h}^*(\mathbf{x}),\mathbf{y}_{C+1}) {\rm d}P_{X|Y=\mathbf{y}_{C+1}}(\mathbf{x})=\int_{{\rm supp}~P_{X|Y=\mathbf{y}_{C+1}}}\ell({\bm h}^*(\mathbf{x}),\mathbf{y}_{C+1}) {\rm d}P_{X|Y= \mathbf{y}_{C+1}}(\mathbf{x})=0.
\end{equation*}
Similarly, we can prove that for any ${\bm h}^*\in \argmin_{{\bm h}\in \mathcal{H}}R_Q^{\alpha}({\bm h})$,
\begin{equation*}
    \int_{\mathcal{X}}\ell({\bm h}^*(\mathbf{x}),\mathbf{y}_{C+1}) {\rm d}Q_{X|Y=\mathbf{y}_{C+1}}(\mathbf{x})=0.
\end{equation*}

\textbf{Step 5.} Given any ${h}_Q\in \argmin_{{\bm h}\in \mathcal{H}}R_Q^{\alpha}({\bm h})$, we can find that (using result of Step 3)
\begin{equation*}
    R_Q^{\alpha}(h_Q) = (1-\alpha) R_{Q,k}(h_Q)=(1-\alpha) R_{P,k}(h_Q),
\end{equation*}
and
\begin{equation*}
    \int_{\mathcal{X}}\ell({\bm h}_Q(\mathbf{x}),\mathbf{y}_{C+1}) {\rm d}Q_{X|Y=\mathbf{y}_{C+1}}(\mathbf{x})=0.
\end{equation*}

Because
$P_X \ll Q_X$, we know
\begin{equation*}
    P_{X|Y=\mathbf{y}_{C+1}}\ll  Q_{X|Y=\mathbf{y}_{C+1}},
\end{equation*}
which implies that 
\begin{equation*}
    \int_{\mathcal{X}}\ell({\bm h}_Q(\mathbf{x}),\mathbf{y}_{C+1}) {\rm d}P_{X|Y=\mathbf{y}_{C+1}}(\mathbf{x})=0.
\end{equation*}
Hence,
\begin{equation*}
    R_Q^{\alpha}({\bm h}_Q) = (1-\alpha) R_{Q,k}({\bm h}_Q)=(1-\alpha) R_{P,k}({\bm h}_Q)+\alpha * 0= R_{P}^{\alpha}({\bm h}_Q).
\end{equation*}

Using the result (see Eq. (\ref{Eqq:equalPQ})) of Step 3,
\begin{equation*}
    \min_{{\bm h}\in \mathcal{H}} R_Q^{\alpha}({\bm h})=\min_{{\bm h}\in \mathcal{H}} R_P^{\alpha}({\bm h}).
\end{equation*}

We obtain that
\begin{equation*}
    {\bm h}_Q \in \argmin_{{\bm h}\in \mathcal{H}} R_P^{\alpha}({\bm h}),
\end{equation*}
this implies
\begin{equation*}
\argmin_{{\bm h}\in \mathcal{H}} R_Q^{\alpha}({\bm h})    \subset \argmin_{{\bm h}\in \mathcal{H}} R_P^{\alpha}({\bm h}).
\end{equation*}
\end{proof}

\newpage
\section{Appendix D: Proof of Theorem 3}\label{AppD}
\begin{lemma}\label{lemma 1}
For any ${\bm h}\in \mathcal{H}$,
\begin{equation*}
R^{\alpha}_Q({\bm h})=(1-\alpha) R_{P,k}({\bm h})+ \max\{R_Q({\bm h},\mathbf{y}_{C+1})-(1-\alpha)R_{P,k}({\bm h},\mathbf{y}_{C+1}),0\},
\end{equation*}
where $\alpha = Q(Y=\mathbf{y}_{C+1})$,
\begin{equation*}
    R_Q({\bm h},\mathbf{y}_{C+1})=\int_{\mathcal{X}} \ell({\bm h}(\mathbf{x}),\mathbf{y}_{C+1}){\rm d}Q_X(\mathbf{x}),
\end{equation*}
and
\begin{equation*}
    R_{P,k}({\bm h},\mathbf{y}_{C+1})=\int_{\mathcal{X}} \ell({\bm h}(\mathbf{x}),\mathbf{y}_{C+1}){\rm d}P_{X|Y\in \mathcal{Y}_k}(\mathbf{x}),
\end{equation*}
\end{lemma}
\begin{proof}
\textbf{Step 1.} We claim that $R^{\alpha}_Q({\bm h})=(1-\alpha) R_{P,k}({\bm h})+\alpha R_{Q,u}({\bm h})$.

First, it is clear that
\begin{equation}\label{Eq::1}
    R^{\alpha}_Q({\bm h})=(1-\alpha) R_{Q,k}({\bm h})+\alpha R_{Q,u}({\bm h}).
\end{equation}

Because $Q_{X,Y|Y\in \mathcal{Y}_k}=P_{X,Y|Y\in \mathcal{Y}_k}$, hence,
\begin{equation}\label{Eq::2}
\begin{split}
    R_{Q,k}({\bm h})=&\int_{\mathcal{X}\times \mathcal{Y}_k} \ell({\bm h}(\mathbf{x}),\mathbf{y}){\rm d}Q_{X,Y|Y\in \mathcal{Y}_k}(\mathbf{x},\mathbf{y})\\=&\int_{\mathcal{X}\times \mathcal{Y}_k} \ell({\bm h}(\mathbf{x}),\mathbf{y}){\rm d}P_{X,Y|Y\in \mathcal{Y}_k}(\mathbf{x},\mathbf{y})\\=&R_{P,k}({\bm h}).
\end{split}
\end{equation}
Combining Eq. (\ref{Eq::1}) with Eq. (\ref{Eq::2}), we have that
\begin{equation*}
    R^{\alpha}_Q({\bm h})=(1-\alpha) R_{P,k}({\bm h})+\alpha R_{Q,u}({\bm h}).
\end{equation*}

\textbf{Step 2.} We claim that $\alpha R_{Q,u}({\bm h})=\max\{R_Q({\bm h},\mathbf{y}_{C+1})-(1-\alpha)R_{P,k}({\bm h},\mathbf{y}_{C+1}),0\}$.

First, it is clear that
\begin{equation}\label{Eq::3}
\begin{split}
  R_Q({\bm h},\mathbf{y}_{C+1})=& (1-\alpha)\int_{\mathcal{X}}\ell({\bm h}(\mathbf{x}), \mathbf{y}_{C+1}){\rm d}Q_{X|Y\in \mathcal{Y}_k}+\alpha \int_{\mathcal{X}}\ell({\bm h}(\mathbf{x}), \mathbf{y}_{C+1}){\rm d}Q_{X|Y=\mathbf{y}_{C+1}}\\ =&(1-\alpha)\int_{\mathcal{X}}\ell({\bm h}(\mathbf{x}), \mathbf{y}_{C+1}){\rm d}P_{X|Y\in \mathcal{Y}_k}+\alpha \int_{\mathcal{X}}\ell({\bm h}(\mathbf{x}), \mathbf{y}_{C+1}){\rm d}Q_{X|Y=\mathbf{y}_{C+1}}\\ =&(1-\alpha)R_{P,k}({\bm h},\mathbf{y}_{C+1})+\alpha \int_{\mathcal{X}}\ell({\bm h}(\mathbf{x}), \mathbf{y}_{C+1}){\rm d}Q_{X|Y=\mathbf{y}_{C+1}}\\ = &(1-\alpha)R_{P,k}({\bm h},\mathbf{y}_{C+1})+\alpha R_{Q,u}({\bm h}).
  \end{split}
\end{equation}  
 Hence,
 \begin{equation*}
     \alpha R_{Q,u}({\bm h})=R_Q({\bm h},\mathbf{y}_{C+1})-(1-\alpha)R_{P,k}({\bm h},\mathbf{y}_{C+1}).
 \end{equation*}
 
 Because $\alpha R_{Q,u}({\bm h})\geq 0$, we obtain that
 \begin{equation*}
     \alpha R_{Q,u}({\bm h})=\max\{R_Q({\bm h},\mathbf{y}_{C+1})-(1-\alpha)R_{P,k}({\bm h},\mathbf{y}_{C+1}),0\}.
 \end{equation*}

\textbf{Step 3.} Combining the results of Steps 1 and Steps 2, we have that
\begin{equation*}
    R^{\alpha}_Q({\bm h})=(1-\alpha) R_{P,k}({\bm h})+ \max\{R_Q({\bm h},\mathbf{y}_{C+1})-(1-\alpha)R_{P,k}({\bm h},\mathbf{y}_{C+1}),0\}.
\end{equation*}
\end{proof}

\begin{lemma}\cite{Kanamori2009ConditionNA,DBLP:journals/ml/KanamoriSS12}.\label{K2009}
Assume the feature space $\mathcal{X}$ is compact. Let the RKHS $\mathcal{H}_K$ be the Hilbert space with Gaussian kernel. Suppose that the real density $p/q\in \mathcal{H}_K$ and set the regularization parameter $\lambda=\lambda_{n,m}$ in {\rm KuLSIF} such that
 \begin{equation*}
   \lim_{n,m\rightarrow 0}  \lambda_{n,m} =0,~~~\lambda_{n,m}^{-1}={O}({\min \{n,m\}}^{1-\delta}),
 \end{equation*}
 where $0<\delta<1$ is any constant, then
 \begin{equation*}
    \sqrt{ \int_{\mathcal{X}}(\widehat{w}(\mathbf{x})-r(\mathbf{x}))^2 {\rm d}U(\mathbf{x})} = O_p(\lambda_{n,m}^{\frac{1}{2}}),
 \end{equation*}
 and
 \begin{equation*}
    \|\widehat{w}\|_{\mathcal{H}_K} = O_p(1),
 \end{equation*}
 where $\widehat{w}$ is the solution of {\rm KuLSIF}.
\end{lemma}
\begin{proof}
The result 
\begin{equation*}
    \sqrt{ \int_{\mathcal{X}}(\widehat{w}(\mathbf{x})-r(\mathbf{x}))^2 {\rm d}U(\mathbf{x})} = O_p(\lambda_{n,m}^{\frac{1}{2}}),
 \end{equation*}
 can be found in Theorem 1 of \cite{Kanamori2009ConditionNA} and Theorem 2 of \cite{DBLP:journals/ml/KanamoriSS12}.

 The result
 \begin{equation*}
    \|\widehat{w}\|_{\mathcal{H}_K} = O_p(1)
 \end{equation*}
 can be found in the proving process (pages $27$-$28$) of Theorem 1 of \cite{Kanamori2009ConditionNA} and the proving process (pages $354$-$365$) of Theorem 2 of \cite{DBLP:journals/ml/KanamoriSS12}.
\end{proof}

Then, we introduce the Rademacher Complexity.
\begin{Definition}[{Rademacher Complexity}]
Let $\mathcal{F}$ be a class of real-valued functions defined in a space $\mathcal{Z}$. Given a distribution $P$ over $\mathcal{Z}$ and sample $\tilde{S}=\{{\bm z_1},...,{\bm z_{\tilde{n}}}\}\in \mathcal{Z}$ drawn i.i.d. from $P$, then  the \textit{Empirical Rademacher Complexity} of $\mathcal{F}$ with respect to the sample $\tilde{S}$ is
\begin{equation}
     \widehat{\Re}_{\tilde{S}}(\mathcal{F}):=\mathbb{E}_{\sigma}[\sup_{f\in \mathcal{F}}\frac{1}{\tilde{n}}\sum_{i=1}^{\tilde{n}}\sigma_if({\bm z}_i)],
\end{equation}
where $\sigma = (\sigma_1,...,\sigma_{\tilde{n}})$ are Rademacher variables, with $\sigma_i$s independent uniform random variables taking values in $-1,+1$. 

Then the Rademacher complexity
\begin{equation}
     {\Re}_{\tilde{n},P}(\mathcal{F}):=\mathbb{E}_{\tilde{S}\sim P^{\tilde{n}}}\widehat{\Re}_{\tilde{S}}(\mathcal{F}).
\end{equation}
\end{Definition}

With the Rademacher complexity, we have 
\begin{lemma}\label{4.1}\textbf{(Theorem 26.5 in} \cite{shalev2014understanding}.)
Given a space $\mathcal{Z}$, a function $l:R \times \mathcal{Z} \rightarrow \mathbb{R}_+$ and a hypothesis set $\mathcal{H}\subset \{f:\mathcal{Z}\rightarrow R\}$, let
\begin{equation*}
    \mathcal{F}:=l\circ \mathcal{H}=\{l(f({\bm z}),{\bm z}): f\in \mathcal{H}\},
\end{equation*} where $l\leq B$.
Then for a distribution $P$ on space $\mathcal{Z}$, data $\tilde{S}=\{{\bm z_1},...,{\bm z_{\tilde{n}}}\}\sim P$ i.i.d, we have  with a probability of at least $1-\delta>0$, for all $f\in \mathcal{F}$:
\begin{equation}
    \widehat{R}(f)-R(f)\leq 2 \widehat{\Re}_{\tilde{S}}(\mathcal{F})+4B\sqrt{\frac{2\log(4/\delta)}{\tilde{n}}},
\end{equation}
where $R(f):=\int_{\mathcal{Z}}l(f({\bm z}),{\bm z}){\rm d} Q({\bm z})$ and $\widehat{R}(f):=\frac{1}{\tilde{n}}\sum_{i=1}^{\tilde{n}} l(f({\bm z}_i),{\bm z}_i)$.
\end{lemma}
Using the same technique as in Lemma \ref{4.1}, we have
with a probability of at least $1-2\delta>0$, for all $f\in \mathcal{F}$:
\begin{equation}\label{Eq::fEstimation1}
    |R(f)-\widehat{R}(f)|\leq 2 \widehat{\Re}_{\tilde{S}}(\mathcal{F})+4B\sqrt{\frac{2\log(4/\delta)}{\tilde{n}}}.
\end{equation}

\begin{Definition}[Shattering \cite{shalev2014understanding}]Given a feature space $\mathcal{X}$, we say that a set $U \subset \mathcal{X}$ is shattered by $\mathcal{H}$ if there exist two functions ${\bm h}_0,{\bm h}_1: U\rightarrow \mathcal{Y}$, such that
\\
$\bullet$ For every $\mathbf{x}\in U$, ${\bm h}_0(\mathbf{x})\neq {\bm h}_1(\mathbf{x}).$\\
$\bullet$ For every $V \subset U$, there exists a function ${\bm h}\in \mathcal{H}$ such that
$\forall \mathbf{x}\in V, {\bm h}(\mathbf{x})={\bm h}_0(\mathbf{x})$ and $\forall \mathbf{x}\in U\backslash V, {\bm h}(\mathbf{x})={\bm h}_1(\mathbf{x})$.
\end{Definition}
Hence, we can define the Natarajan dimension as follows.
\begin{Definition}[Natarajan Dimension \cite{shalev2014understanding}] The Natarajan dimension of $\mathcal{H}$, denoted Ndim$(\mathcal{H})$, is the maximal size of a shattered set $U \subset X$ .
\end{Definition}
It is not difficult to see that in the case that there are exactly two classes,
Ndim($\mathcal{H}$) = VCdim($\mathcal{H}$). Therefore, the Natarajan dimension generalizes the VC
dimension.

\begin{lemma}\label{lemma4}
Assume that $\mathcal{H}\subset \{{\bm h}:\mathcal{X}\rightarrow \mathcal{Y}\}$ has finite Natarajan dimension and the loss function $\ell$ has upper bound $c$, then for any $0<\delta<1$,
\begin{equation*}
    \sup_{{\bm h}\in \mathcal{H}} |R_{P,k}({\bm h})-\widehat{R}_{S}({\bm h})|= cO_p(1/n^{\frac{1-\delta}{2}}),~~~ \sup_{{\bm h}\in \mathcal{H}} |R_{P,k}({\bm h},\mathbf{y}_{C+1})-\widehat{R}_{S}({\bm h},\mathbf{y}_{C+1})|= cO_p(1/n^{\frac{1-\delta}{2}}),
\end{equation*}
where
\begin{equation*}
\widehat{R}_{S}({\bm h}):=\frac{1}{n}\sum_{(\mathbf{x},\mathbf{y})\in S} \ell({\bm h({\mathbf{x}})}, \mathbf{y}),~~~\widehat{R}_{S}({\bm h},\mathbf{y}_{C+1}):=\frac{1}{n}\sum_{(\mathbf{x},\mathbf{y})\in S} \ell({\bm h({\mathbf{x}})}, \mathbf{y}_{C+1}).
\end{equation*}
\end{lemma}
\begin{proof}
Assume that the Natarajan dimension is $d$ and the upper bound of $\ell$ is $B$.

 Let $\mathcal{F}=\{\ell({\bm h}(\mathbf{x}),\mathbf{y}):{\bm h}\in \mathcal{H}\}$. Then the Natarajan lemma (Lemma 29.4 of \cite{shalev2014understanding})  tells us that
\begin{equation*}
    |\{{\bm h}(\mathbf{x}^1),...,{\bm h}(\mathbf{x}^{n})|{\bm h}\in \mathcal{H} \}| \leq n^{d} (C+1)^{2d}.
\end{equation*}
Denote $A=\{ (\ell({\bm h}(\mathbf{x}^1),{\bm h}'(\mathbf{x}^1)),...,\ell({\bm h}(\mathbf{x}^{n}),{\bm h}'(\mathbf{x}^{n}))|{\bm h},{\bm h}'\in \mathcal{H}\}.  $ This clearly implies that
\begin{equation*}
    |A|\leq |\{{\bm h}(\mathbf{x}^1),...,{\bm h}(\mathbf{x}^{n})|{\bm h}\in \mathcal{H} \}|^2 \leq (n)^{2d} (C+1)^{4d}.
\end{equation*}
Combining above inequality with Lemma 26.8 of \cite{shalev2014understanding} and inequality (\ref{Eq::fEstimation1}), we obtain with a probability of at least $1-2\delta>0$, 
\begin{equation*}
\begin{split}
 \sup_{{\bm h}\in \mathcal{H}} |R_{P,k}({\bm h})-\widehat{R}_{S}({\bm h})|\leq      2\widehat{\Re}_{S}(\mathcal{F})+4c\sqrt{\frac{2\log \frac{4}{\delta}}{n}}\leq & 2c\sqrt{\frac{4d\log n + 8d \log(C+1)}{n}}+4c\sqrt{\frac{2\log \frac{4}{\delta}}{n}},
    \end{split}
\end{equation*}
hence,
\begin{equation*}
    \sup_{{\bm h}\in \mathcal{H}} |R_{P,k}({\bm h})-\widehat{R}_{S}({\bm h})|=cO_p(1/n^{\frac{1-\delta}{2}}).
\end{equation*}

Using the same technique, we can also prove that
$
    \sup_{{\bm h}\in \mathcal{H}} |R_{P,k}({\bm h},\mathbf{y}_{C+1})-\widehat{R}_{S}({\bm h},\mathbf{y}_{C+1})|= cO_p(1/n^{\frac{1-\delta}{2}}).
$
\end{proof}

\begin{lemma}\label{lemma5}
Assume the feature space $\mathcal{X}$ is compact and the loss function has an upper bound $c$. Let the RKHS $\mathcal{H}_K$ is the Hilbert space with Gaussian kernel. Suppose that the real density $p/q\in \mathcal{H}_K$ and set the regularization parameter $\lambda=\lambda_{n,m}$ in {\rm KuLSIF} such that
 \begin{equation*}
   \lim_{n,m\rightarrow 0}  \lambda_{n,m} =0,~~~\lambda_{n,m}^{-1}={O}({\min \{n,m\}}^{1-\delta}),
 \end{equation*}
 where $0<\delta<1$ is any constant, then
 \begin{equation*}
   \sup_{{\bm h}\in \mathcal{H}}|R_{Q}({\bm h},\mathbf{y}_{C+1})-\gamma \widehat{R}^{\tau,\beta}_T({\bm h},\mathbf{y}_{C+1})|\leq \gamma \beta c U(\{\mathbf{x}: ~0<r(\mathbf{x})\leq 2\tau\})+c\big(\max\{1,\frac{\beta}{\tau}\}+\beta\big) O_p(\lambda^{\frac{1}{2}}_{n,m}),
 \end{equation*}
 where 
 \begin{equation*}
 \begin{split}
  &R_Q({\bm h},\mathbf{y}_{C+1})=\int_{\mathcal{X}} \ell({\bm h}(\mathbf{x}),\mathbf{y}_{C+1}){\rm d}Q_X(\mathbf{x}),~~~ 
      \widehat{R}^{\tau,\beta}_T({\bm h},\mathbf{y}_{K+1}):=\frac{1}{m}\sum_{\mathbf{x}\in T} L_{\tau,\beta}({\widehat{w}(\mathbf{x}})) \ell({\bm h({\mathbf{x}})}, \mathbf{y}_{K+1}),
      \end{split}
 \end{equation*}
 here $Q_{X}:=Q_{U}^{0,\beta}$, and $\widehat{w}$ is the solution of {\rm KuLSIF}.
\end{lemma}
\begin{proof}
\textbf{Step 1.} We claim that  
\begin{equation*}
\sup_{{\bm h}\in \mathcal{H}}|R_{Q}({\bm h},\mathbf{y}_{C+1})-\gamma {R}^{\tau,\beta}_U({\bm h},\mathbf{y}_{C+1})|\leq \gamma \beta c U(\{\mathbf{x}: ~0<r(\mathbf{x})\leq 2\tau\}),
\end{equation*}
where 
\begin{equation*}
   {R}^{\tau,\beta}_U({\bm h},\mathbf{y}_{C+1})=\int_{\mathcal{X}} L_{\tau,\beta}(r(\mathbf{x}))\ell({\bm h}(\mathbf{x}),\mathbf{y}_{C+1}){\rm d}U(\mathbf{x}),
\end{equation*}
here $r(\mathbf{x})=p(\mathbf{x})/q(\mathbf{x})$.

First, we note that
\begin{equation}\label{Eq::difference}
\begin{split}
 &  \big | \int_{\mathcal{X}} L_{0,\beta}(r(\mathbf{x}))\ell({\bm h}(\mathbf{x}),\mathbf{y}_{C+1}){\rm d}U(\mathbf{x})-\int_{\mathcal{X}} L_{\tau,\beta}(r(\mathbf{x}))\ell({\bm h}(\mathbf{x}),\mathbf{y}_{C+1}){\rm d}U(\mathbf{x})|
\\ \leq &  \big | \int_{\mathcal{X}} L_{0,\beta}(r(\mathbf{x}))\ell({\bm h}(\mathbf{x}),\mathbf{y}_{C+1}){\rm d}U(\mathbf{x})-\int_{\mathcal{X}} L_{\tau,\beta}(r(\mathbf{x}))\ell({\bm h}(\mathbf{x}),\mathbf{y}_{C+1}){\rm d}U(\mathbf{x})|\\ \leq & 
c \int_{\mathcal{X}} \big |L_{0,\beta}(r(\mathbf{x}))- L_{\tau,\beta}(r(\mathbf{x}))\big |{\rm d}U(\mathbf{x})\\ \leq &c \int_{\{\mathbf{x}: ~0<r(\mathbf{x})\leq 2\tau\}} \beta {\rm d}U(\mathbf{x})=\beta c U(\{\mathbf{x}: ~0<r(\mathbf{x})\leq 2\tau\}).
   \end{split}
\end{equation}
Because $Q_{X,Y}=Q_U^{0,\beta}P_{Y|X}$, then according to the definition of $Q_U^{0,\beta}$, we know
\begin{equation*}
    R_{Q}({\bm h},\mathbf{y}_{C+1})=\gamma \int_{\mathcal{X}} L_{0,\beta}(r(\mathbf{x}))\ell({\bm h}(\mathbf{x}),\mathbf{y}_{C+1}){\rm d}U(\mathbf{x}),
\end{equation*}
which implies
\begin{equation*}
    \sup_{{\bm h}\in \mathcal{H}}|R_{Q}({\bm h},\mathbf{y}_{C+1})-\gamma {R}^{\tau,\beta}_U({\bm h},\mathbf{y}_{C+1})|\leq \gamma \beta c U(\{\mathbf{x}: ~0<r(\mathbf{x})\leq 2\tau\}).
\end{equation*}

\textbf{Step 2.} We claim that
\begin{equation*}
   \sup_{{\bm h}\in \mathcal{H}}| {R}^{\tau,\beta}_U({\bm h},\mathbf{y}_{C+1})-\int_{\mathcal{X}}L_{\tau,\beta}(\widehat{w}(\mathbf{x}))\ell({\bm h}(\mathbf{x}),\mathbf{y}_{C+1}){\rm d}U(\mathbf{x})| \leq \max\{c,\frac{c\beta}{\tau}\}O_p(\lambda^{\frac{1}{2}}_{n,m}).
\end{equation*}
First, the Lipschitz constant for $L_{\tau,\beta}$ is smaller than $\max\{1,\frac{\beta}{\tau}\}$.

Then,
\begin{equation*}
\begin{split}
&\sup_{{\bm h}\in \mathcal{H}}| {R}^{\tau,\beta}_U({\bm h},\mathbf{y}_{C+1})-\int_{\mathcal{X}}L_{\tau,\beta}(\widehat{w}(\mathbf{x}))\ell({\bm h}(\mathbf{x}),\mathbf{y}_{C+1}){\rm d}U(\mathbf{x})|\\=&\sup_{{\bm h}\in \mathcal{H}}| \int_{\mathcal{X}}L_{\tau,\beta}(r(\mathbf{x}))\ell({\bm h}(\mathbf{x}),\mathbf{y}_{C+1}){\rm d}U(\mathbf{x})-\int_{\mathcal{X}}L_{\tau,\beta}(\widehat{w}(\mathbf{x}))\ell({\bm h}(\mathbf{x}),\mathbf{y}_{C+1}){\rm d}U(\mathbf{x}) \\ \leq &
   \sup_{{\bm h}\in \mathcal{H}}\int_{\mathcal{X}}|L_{\tau,\beta}(r(\mathbf{x}))-L_{\tau,\beta}(\widehat{w}(\mathbf{x}))|\ell({\bm h}(\mathbf{x}),\mathbf{y}_{C+1}){\rm d}U(\mathbf{x})|\\ \leq & 
    \sup_{{\bm h}\in \mathcal{H}} \sqrt{\int_{\mathcal{X}} |L_{\tau,\beta}(r(\mathbf{x}))-L_{\tau,\beta}(\widehat{w}(\mathbf{x}))|^2 {\rm d}U(\mathbf{x})}\sqrt{\int_{\mathcal{X}} \ell^2({\bm h}(\mathbf{x}),\mathbf{y}_{C+1}) {\rm d}U(\mathbf{x})}~~{\rm H\ddot{o}lder~~ Inequality}\\ \leq & c\sup_{{\bm h}\in \mathcal{H}} \sqrt{\int_{\mathcal{X}} |L_{\tau,\beta}(r(\mathbf{x}))-L_{\tau,\beta}(\widehat{w}(\mathbf{x}))|^2 {\rm d}U(\mathbf{x})}\\ \leq & \max\{c,\frac{c\beta}{\tau}\}\sup_{{\bm h}\in \mathcal{H}}\sqrt{\int_{\mathcal{X}} |r(\mathbf{x})-\widehat{w}(\mathbf{x})|^2 {\rm d}U(\mathbf{x})}.
   \end{split}
\end{equation*}
Lastly, using Lemma \ref{K2009}, 
\begin{equation*}
    \sup_{{\bm h}\in \mathcal{H}}| {R}^{\tau,\beta}_U({\bm h},\mathbf{y}_{C+1})-\int_{\mathcal{X}}L_{\tau,\beta}(\widehat{w}(\mathbf{x}))\ell({\bm h}(\mathbf{x}),\mathbf{y}_{C+1}){\rm d}U(\mathbf{x})|\leq \max\{c,\frac{c\beta}{\tau}\}O_p(\lambda^{\frac{1}{2}}_{n,m}).
\end{equation*}

\textbf{Step 3.} We claim that 
\begin{equation*}
     \sup_{{\bm h}\in \mathcal{H}}\big | \frac{1}{m} \sum_{\mathbf{x}\in T} L_{\tau,\beta}(\widehat{w}(\mathbf{x}))\ell({\bm h}(\mathbf{x}),\mathbf{y}_{C+1})-\int_{\mathcal{X}}L_{\tau,\beta}(\widehat{w}(\mathbf{x}))\ell({\bm h}(\mathbf{x}),\mathbf{y}_{C+1}){\rm d}U(\mathbf{x})\big |\leq c\big(\max\{1,\frac{\beta}{\tau}\}+\beta\big) O_p(\lambda^{\frac{1}{2}}_{n,m}).
\end{equation*}

First, we set $\mathcal{F}_{B}:= \{L_{\tau,\beta}(w)\ell({\bm h},\mathbf{y}_{C+1}): w\in \mathcal{H}_K, \|w\|_K \leq B, {\bm h}\in \mathcal{H}\}$. We consider
\begin{equation*}
    \sup_{f \in \mathcal{F}_{B}}\big | \frac{1}{m} \sum_{\mathbf{x}\in T} f(\mathbf{x})-\int_{\mathcal{X}}f(\mathbf{x}){\rm d}U(\mathbf{x})\big |
\end{equation*}

Using Lemma \ref{4.1} and inequality \ref{Eq::fEstimation1}, it is easy to check that for $1-2\delta>0$, we have
\begin{equation}\label{Eq::fEstimation}
    \sup_{f\in \mathcal{F}_B}\big | \frac{1}{m} \sum_{\mathbf{x}\in T} f(\mathbf{x})-\int_{\mathcal{X}}f(\mathbf{x}){\rm d}U(\mathbf{x})\big |\leq 2 \widehat{\Re}_{T}(\mathcal{F}_B)+4(B+\beta)c\sqrt{\frac{2\log(4/\delta)}{m}},
\end{equation}
here we have used $|f|\leq (B+\beta)c$, for any $f\in \mathcal{F}_B$.

Then, we consider $\widehat{\Re}_{T}(\mathcal{F}_B)$.
\begin{equation*}
\begin{split}
   &m\widehat{\Re}_{T}(\mathcal{F}_B) \\=& \mathbb{E}_{\sigma}[\sup_{\|w\|_K\leq B, {\bm h}\in \mathcal{H}}\sum_{i=1}^m \sigma_i L_{\tau,\beta}(w_i)\ell({{\bm h}_i,\mathbf{y}_{C+1}}) ]\\ = &  \mathbb{E}_{\sigma}[\sup_{\|w\|_K\leq B, {\bm h}\in \mathcal{H}}\sigma_1 L_{\tau,\beta}(w_1)\ell({{\bm h}_1,\mathbf{y}_{C+1}})+\sum_{i=2}^m \sigma_i L_{\tau,\beta}(w_i)\ell({{\bm h}_i,\mathbf{y}_{C+1}}) ] \\ =&\frac{1}{2}\mathbb{E}_{\sigma_2,...,\sigma_m}[\sup_{\|w\|_K\leq B, {\bm h}\in \mathcal{H}} L_{\tau,\beta}(w_1)\ell({{\bm h}_1,\mathbf{y}_{C+1}})+\sum_{i=2}^m \sigma_i L_{\tau,\beta}(w_i)\ell({{\bm h}_i,\mathbf{y}_{C+1}})\\&~~~~~~~~~~~~~~~+ \sup_{\|w\|_K\leq B, {\bm h}\in \mathcal{H}}-L_{\tau,\beta}(w_1)\ell({{\bm h}_1,\mathbf{y}_{C+1}})+\sum_{i=2}^m \sigma_i L_{\tau,\beta}(w_i)\ell({{\bm h}_i,\mathbf{y}_{C+1}})]\\ = &\frac{1}{2}\mathbb{E}_{\sigma_2,...,\sigma_m}[\sup_{\|w\|_K\leq B, {\bm h}\in \mathcal{H}}\sup_{\|w'\|_K \leq B, {\bm h}'\in \mathcal{H}} L_{\tau,\beta}(w_1)\ell({{\bm h}_1,\mathbf{y}_{C+1}})+\sum_{i=2}^m \sigma_i L_{\tau,\beta}(w_i)\ell({{\bm h}_i,\mathbf{y}_{C+1}}) \\&~~~~~~~~~~~~~~~~~~~~~~~~~~~~~~~~~~~~~~~~~~~~~~~~~~~~~~~~~~~~~~-L_{\tau,\beta}(w_1')\ell({{\bm h}_1',\mathbf{y}_{C+1}})+\sum_{i=2}^m \sigma_i L_{\tau,\beta}(w_i')\ell({{\bm h}_i',\mathbf{y}_{C+1}})] \\ \leq &\frac{1}{2}\mathbb{E}_{\sigma_2,...,\sigma_m}[\sup_{\|w\|_K\leq B, {\bm h}\in \mathcal{H}}\sup_{\|w'\|_K \leq B, {\bm h}'\in \mathcal{H}} |L_{\tau,\beta}(w_1)-L_{\tau,\beta}(w_1')|\ell({{\bm h}_1',\mathbf{y}_{C+1}})+L_{\tau,\beta}(w_1)|\ell({{\bm h}_1',\mathbf{y}_{C+1}})-\ell({{\bm h}_1,\mathbf{y}_{C+1}})| \\&~~~~~~~~~~~~~~~~~~~~~~~~~~~~~~~~~~~~~~~~~~~~~~~~~~~~~~~~~~~~~~+\sum_{i=2}^m \sigma_i L_{\tau,\beta}(w_i)\ell({{\bm h}_i,\mathbf{y}_{C+1}})+\sum_{i=2}^m \sigma_i L_{\tau,\beta}(w_i')\ell({{\bm h}_i',\mathbf{y}_{C+1}})]\\ \leq &\frac{1}{2}\mathbb{E}_{\sigma_2,...,\sigma_m}[\sup_{\|w\|_K\leq B, {\bm h}\in \mathcal{H}}\sup_{\|w'\|_K \leq B, {\bm h}'\in \mathcal{H}} Lc|w_1-w_1'|+(B+\beta)|\ell({{\bm h}_1',\mathbf{y}_{C+1}})-\ell({{\bm h}_1,\mathbf{y}_{C+1}})| \\&~~~~~~~~~~~~~~~~~~~~~~~~~~~~~~~~~~~~~~~~~~~~~~~~~~~~~~~~~~~~~~+\sum_{i=2}^m \sigma_i L_{\tau,\beta}(w_i)\ell({{\bm h}_i,\mathbf{y}_{C+1}})+\sum_{i=2}^m \sigma_i L_{\tau,\beta}(w_i')\ell({{\bm h}_i',\mathbf{y}_{C+1}})] \\  =&\mathbb{E}_{\sigma}[\sup_{\|w\|_K\leq B, {\bm h}\in \mathcal{H}} Lc\sigma_1 w_1+(B+\beta)\sigma_1 \ell({{\bm h}_1,\mathbf{y}_{C+1}}) +\sum_{i=2}^m \sigma_i L_{\tau,\beta}(w_i)\ell({{\bm h}_i,\mathbf{y}_{C+1}}) ] \\ &{\rm~~Repeat ~~the~~ process~~} m-1 {\rm~~ times~~for~~} i=2,...,m. \\ \leq &\mathbb{E}_{\sigma}[\sup_{\|w\|_K\leq B, {\bm h}\in \mathcal{H}} \sum_{i=1}^m Lc\sigma_i w_i+\sum_{i=1}^m(B+\beta)\sigma_i \ell({{\bm h}_i,\mathbf{y}_{C+1}}) ]\\ \leq & mLc \widehat{\Re}_{T}(\mathcal{H}_{K,B}) +m(B+\beta) \widehat{\Re}_{T}(\mathcal{F}),
   \end{split}
\end{equation*}
where $w_i=w(\tilde{\mathbf{x}}_i), {\bm h}_i={\bm h}(\tilde{\mathbf{x}}_i)$, $L=\max\{1,\frac{\beta}{\tau}\}$, $\mathcal{H}_{K,B}=\{w: w\in \mathcal{H}_K,\|w\|_K\leq B\}$ and $\mathcal{F}=\{\ell({\bm h},\mathbf{y}_{C+1}):{\bm h}\in \mathcal{H}\}$.

According to Theorem 5.5 of \cite{10.5555/2371238}, we obtain that
\begin{equation*}
    \widehat{\Re}_{T}(\mathcal{H}_{K,B}) \leq B \sqrt{\frac{1}{m}}.
\end{equation*}
According to the proving process of Lemma \ref{lemma4}, we obtain that
\begin{equation*}
    \widehat{\Re}_{T}(\mathcal{F}) \leq c\sqrt{\frac{4d\log m + 8d \log(C+1)}{m}},
\end{equation*}
where $d$ is the Natarajan Dimension of $\mathcal{H}$.

Hence,
\begin{equation*}
    \widehat{\Re}_{T}(\mathcal{F}_B) \leq BLc\sqrt{\frac{1}{m}}+(B+\beta)c\sqrt{\frac{4d\log m + 8d \log(C+1)}{m}}.
\end{equation*}

This implies that
for $1-2\delta>0$, we have
\begin{equation}\label{Eq::fEstimation-2}
\begin{split}
    &\sup_{w\in \mathcal{H}_K,\|w\|_K\leq B}\sup_{{\bm h}\in \mathcal{H}}\big | \frac{1}{m} \sum_{\mathbf{x}\in T} L_{\tau,\beta}(w(\mathbf{x}))\ell({\bm h}(\mathbf{x}),\mathbf{y}_{C+1})-\int_{\mathcal{X}}L_{\tau,\beta}(w(\mathbf{x}))\ell({\bm h}(\mathbf{x}),\mathbf{y}_{C+1}){\rm d}U(\mathbf{x})\big |\\\leq &2 B\max\{1,\frac{\beta}{\tau}\}c\sqrt{\frac{1}{m}}+2(B+\beta)c\sqrt{\frac{4d\log m + 8d \log(C+1)}{m}}+4(B+\beta)c\sqrt{\frac{2\log(4/\delta)}{m}}.
    \end{split}
\end{equation}

Because $\|\widehat{w}\|_K=O_p(1)$, then combining inequality \ref{Eq::fEstimation-2}, we know that
\begin{equation*}
\begin{split}
   & \sup_{{\bm h}\in \mathcal{H}}\big | \frac{1}{m} \sum_{\mathbf{x}\in T} L_{\tau,\beta}(\widehat{w}(\mathbf{x}))\ell({\bm h}(\mathbf{x}),\mathbf{y}_{C+1})-\int_{\mathcal{X}}L_{\tau,\beta}(w(\mathbf{x}))\ell({\bm h}(\mathbf{x}),\mathbf{y}_{C+1}){\rm d}U(\mathbf{x})\big |\\\leq &2 O_p(1)\max\{1,\frac{\beta}{\tau}\}cO_p(\sqrt{\frac{1}{m}})+2(O_p(1)+\beta)cO_p(\sqrt{\frac{4d\log m + 8d \log(C+1)}{m}})+4(O_p(1)+\beta)cO_p(\sqrt{\frac{2\log(4/\delta)}{m}}).
    \end{split}
\end{equation*}
This implies that
\begin{equation*}
    \sup_{{\bm h}\in \mathcal{H}}\big | \frac{1}{m} \sum_{\mathbf{x}\in T} L_{\tau,\beta}(\widehat{w}(\mathbf{x}))\ell({\bm h}(\mathbf{x}),\mathbf{y}_{C+1})-\int_{\mathcal{X}}L_{\tau,\beta}(w(\mathbf{x}))\ell({\bm h}(\mathbf{x}),\mathbf{y}_{C+1}){\rm d}U(\mathbf{x})\big |=c\big(\max\{1,\frac{\beta}{\tau}\}+\beta\big) O_p(\lambda^{\frac{1}{2}}_{n,m}).
\end{equation*}

\textbf{Step 4.} Using the results of Steps 1, 2 and 3, we have
\begin{equation*}
\begin{split}
   &\sup_{{\bm h}\in \mathcal{H}}|R_{Q}({\bm h},\mathbf{y}_{C+1})-\gamma \widehat{R}^{\tau,\beta}_T({\bm h},\mathbf{y}_{C+1})|\\ \leq &
   \sup_{{\bm h}\in \mathcal{H}}|R_{Q}({\bm h},\mathbf{y}_{C+1})-\gamma {R}^{\tau,\beta}_U({\bm h},\mathbf{y}_{C+1})|+\sup_{{\bm h}\in \mathcal{H}}|\gamma {R}^{\tau,\beta}_U({\bm h},\mathbf{y}_{C+1})-\gamma \int_{\mathcal{X}}L_{\tau,\beta}(\widehat{w}(\mathbf{x}))\ell({\bm h}(\mathbf{x}),\mathbf{y}_{C+1}){\rm d}U(\mathbf{x})|\\ + &  \sup_{{\bm h}\in \mathcal{H}}|\gamma \int_{\mathcal{X}}L_{\tau,\beta}(\widehat{w}(\mathbf{x}))\ell({\bm h}(\mathbf{x}),\mathbf{y}_{C+1}){\rm d}U(\mathbf{x})-\gamma \widehat{R}^{\tau,\beta}_T({\bm h},\mathbf{y}_{C+1})| \\ \leq &
   \gamma \beta c U(\{\mathbf{x}: ~0<r(\mathbf{x})\leq 2\tau\})+\gamma \max\{c,\frac{c\beta}{\tau}\}O_p(\lambda^{\frac{1}{2}}_{n,m})+c\big(\max\{1,\frac{\beta}{\tau}\}+\beta\big) O_p(\lambda^{\frac{1}{2}}_{n,m}).
 \end{split}
 \end{equation*}
 
 Note that $\gamma<1$, we can write 
 \begin{equation*}
     \sup_{{\bm h}\in \mathcal{H}}|R_{Q}({\bm h},\mathbf{y}_{C+1})-\gamma \widehat{R}^{\tau,\beta}_T({\bm h},\mathbf{y}_{C+1})|\leq \gamma \beta c U(\{\mathbf{x}: ~0<r(\mathbf{x})\leq 2\tau\})+c\big(\max\{1,\frac{\beta}{\tau}\}+\beta\big) O_p(\lambda^{\frac{1}{2}}_{n,m}).
 \end{equation*}
\end{proof}
\newpage
\begin{proof}[Proof of Theorem 3]
We separate the proof into three steps.

\textbf{Step 1.} We claim that
\begin{equation*}
\begin{split}
&\sup_{{\bm h}\in \mathcal{H}}|(1-\alpha)\Delta_{S,T}^{\tau,\beta}({\bm h})-\max \{R_Q({\bm h},\mathbf{y}_{C+1})-(1-\alpha)R_{P,k}({\bm h},\mathbf{y}_{C+1}),0\}| \\ \leq & \gamma \beta c U(\{\mathbf{x}: ~0<r(\mathbf{x})\leq  2\tau\})+c\big(\max\{1,\frac{\beta}{\tau}\}+\beta\big) O_p(\lambda^{\frac{1}{2}}_{n,m}).
\end{split}
\end{equation*}
First, it is easy to check that
\begin{equation*}
\begin{split}
    &\sup_{{\bm h}\in \mathcal{H}}|(1-\alpha)\Delta_{S,T}^{\tau,\beta}({\bm h})-\max \{R_Q({\bm h},\mathbf{y}_{C+1})-(1-\alpha)R_{P,k}({\bm h},\mathbf{y}_{C+1}),0\}|
\\ \leq &\sup_{{\bm h}\in \mathcal{H}}|(1-\alpha)\widehat{R}^{\tau,\beta}_T({\bm h},\mathbf{y}_{K+1})- (1-\alpha)\widehat{R}_{S}({\bm h},\mathbf{y}_{K+1})-R_Q({\bm h},\mathbf{y}_{C+1})+(1-\alpha)R_{P,k}({\bm h},\mathbf{y}_{C+1})|\\ \leq &
\sup_{{\bm h}\in \mathcal{H}}|(1-\alpha)\widehat{R}^{\tau,\beta}_T({\bm h},\mathbf{y}_{K+1})-R_Q({\bm h},\mathbf{y}_{C+1})|+(1-\alpha)\sup_{{\bm h}\in \mathcal{H}}| \widehat{R}_{S}({\bm h},\mathbf{y}_{K+1})-R_{P,k}({\bm h},\mathbf{y}_{C+1})|
\\ \leq &\gamma \beta c U(\{\mathbf{x}: ~0<r(\mathbf{x})\leq 2\tau\})+c\big(\max\{1,\frac{\beta}{\tau}\}+\beta\big) O_p(\lambda^{\frac{1}{2}}_{n,m})~~{\rm~Use~Lemma~\ref{lemma5}} \\+&(1-\alpha)\sup_{{\bm h}\in \mathcal{H}}| \widehat{R}_{S}({\bm h},\mathbf{y}_{K+1})-R_{P,k}({\bm h},\mathbf{y}_{C+1})| \\ \leq &\gamma \beta c U(\{\mathbf{x}: ~0<r(\mathbf{x})\leq 2\tau\})+c\big(\max\{1,\frac{\beta}{\tau}\}+\beta\big) O_p(\lambda^{\frac{1}{2}}_{n,m}) \\+&(1-\alpha)cO_p(\lambda^{\frac{1}{2}}_{n,m})~~{\rm~Use~Lemma~\ref{lemma4}}.
\end{split}
\end{equation*}
Hence, we can write 
\begin{equation*}
\begin{split}
   &\sup_{{\bm h}\in \mathcal{H}}|(1-\alpha)\Delta_{S,T}^{\tau,\beta}({\bm h})-\max \{R_Q({\bm h},\mathbf{y}_{C+1})-(1-\alpha)R_{P,k}({\bm h},\mathbf{y}_{C+1}),0\}| \\ \leq & \gamma \beta c U(\{\mathbf{x}: ~0<r(\mathbf{x})\leq 2\tau\})+c\big(\max\{1,\frac{\beta}{\tau}\}+\beta\big) O_p(\lambda^{\frac{1}{2}}_{n,m}).
 \end{split}
\end{equation*}

\textbf{Step 2.}
\begin{equation*}
\begin{split}
&\sup_{{\bm h}\in \mathcal{H}} |(1-\alpha) \widehat{R}_S({\bm h})+(1-\alpha)\Delta_{S,T}^{\tau,\beta}({\bm h})-(1-\alpha) R_{P,k}({\bm h})-\max \{R_Q({\bm h},\mathbf{y}_{C+1})-(1-\alpha)R_{P,k}({\bm h},\mathbf{y}_{C+1}),0\}| \\ \leq &\sup_{{\bm h}\in \mathcal{H}} |(1-\alpha) \widehat{R}_S({\bm h})-(1-\alpha) R_{P,k}({\bm h})|+\sup_{{\bm h}\in \mathcal{H}}|(1-\alpha)\Delta_{S,T}^{\tau,\beta}({\bm h})-\max \{R_Q({\bm h},\mathbf{y}_{C+1})-(1-\alpha)R_{P,k}({\bm h},\mathbf{y}_{C+1}),0\}| \\ \leq & (1-\alpha)cO_p(\lambda^{\frac{1}{2}}_{n,m})~~{\rm~Use~Lemma~\ref{lemma4}}
\\ +&\gamma \beta c U(\{\mathbf{x}: ~0<r(\mathbf{x})\leq 2\tau\})+c\big(\max\{1,\frac{\beta}{\tau}\}+\beta\big) O_p(\lambda^{\frac{1}{2}}_{n,m})~~{\rm~Use~the~result~of~Step~1}.
\end{split}
\end{equation*}
Hence, we can write
\begin{equation*}
    \begin{split}
 &\sup_{{\bm h}\in \mathcal{H}} |(1-\alpha) \widehat{R}_S({\bm h})+(1-\alpha)\Delta_{S,T}^{\tau,\beta}({\bm h})-(1-\alpha) R_{P,k}({\bm h})-\max \{R_Q({\bm h},\mathbf{y}_{C+1})-(1-\alpha)R_{P,k}({\bm h},\mathbf{y}_{C+1}),0\}| \\ \leq & \gamma \beta c U(\{\mathbf{x}: ~0<r(\mathbf{x})\leq 2\tau\})+c\big(\max\{1,\frac{\beta}{\tau}\}+\beta\big) O_p(\lambda^{\frac{1}{2}}_{n,m}). 
    \end{split}
\end{equation*}

\textbf{Step 3.} Note that 
\begin{equation*}
    R^{\alpha}_Q({\bm h})=(1-\alpha) R_{P,k}({\bm h})+ \max\{R_Q({\bm h},\mathbf{y}_{C+1})-(1-\alpha)R_{P,k}({\bm h},\mathbf{y}_{C+1}),0\}~~~{\rm Use~Lemma~\ref{lemma 1}.}
\end{equation*}
Hence,
\begin{equation*}
    \begin{split}
 &\sup_{{\bm h}\in \mathcal{H}} |(1-\alpha) \widehat{R}_S({\bm h})+(1-\alpha)\Delta_{S,T}^{\tau,\beta}({\bm h})- R^{\alpha}_Q({\bm h})| \\ \leq & \gamma \beta c U(\{\mathbf{x}: ~0<r(\mathbf{x})\leq 2\tau\})+c\big(\max\{1,\frac{\beta}{\tau}\}+\beta\big) O_p(\lambda^{\frac{1}{2}}_{n,m}). 
    \end{split}
\end{equation*}
\end{proof}

\newpage
\section{Appendix E: Proofs of Theorem 4, Theorem 5 and Theorem 6}\label{AppE}
\subsection{Proof for Theorem}
\begin{proof}[Proof of Theorem 4.]
According to Theorem 1, we know that for any ${\bm h}\in \mathcal{H}$,
\begin{equation}\label{T4P1}
\begin{split}
&|R_P^{\alpha}({\bm h})-R_Q^{\alpha}({\bm h})| \leq \alpha d^{\ell}_{{\bm h},\mathcal{H}}(P_{X|Y=\mathbf{y}_{C+1}},Q_{X|Y=\mathbf{y}_{C+1}})+\alpha \Lambda.
\end{split}
\end{equation}
According to Theorem 3, we know that for any ${\bm h}\in \mathcal{H}$,
\begin{equation}\label{T4P2}
    \begin{split}
 &|(1-\alpha) \widehat{R}_S({\bm h})+(1-\alpha)\Delta_{S,T}^{\tau,\beta}({\bm h})- R^{\alpha}_Q({\bm h})| \\ \leq & \gamma \beta c U(\{\mathbf{x}: ~0<r(\mathbf{x})\leq 2\tau\})+c\big(\max\{1,\frac{\beta}{\tau}\}+\beta\big) O_p(\lambda^{\frac{1}{2}}_{n,m}). 
    \end{split}
\end{equation}

Combining inequalities (\ref{T4P1}) and (\ref{T4P2}), we know that for any ${\bm h}\in \mathcal{H}$,
\begin{equation*}
    \begin{split}
 &|(1-\alpha) \widehat{R}_S({\bm h})+(1-\alpha)\Delta_{S,T}^{\tau,\beta}({\bm h})- R_P^{\alpha}({\bm h})| \\ \leq & \gamma \beta c U(\{\mathbf{x}: ~0<r(\mathbf{x})\leq 2\tau\})+c\big(\max\{1,\frac{\beta}{\tau}\}+\beta\big) O_p(\lambda^{\frac{1}{2}}_{n,m})+\alpha d^{\ell}_{{\bm h},\mathcal{H}}(P_{X|Y=\mathbf{y}_{C+1}},Q_{X|Y=\mathbf{y}_{C+1}})+\alpha \Lambda. 
    \end{split}
\end{equation*}
\end{proof}

\subsection{Proof for Theorem 5}
\begin{proof}[Proof of Theorem 5.]
Assume that
\begin{equation*}
    \widehat{{\bm h}} \in \argmin_{{\bm h}\in \mathcal{H}} \widehat{R}_{S,T}^{\tau,\beta}({\bm h}),~~~{\bm h}_Q \in \argmin_{{\bm h}\in \mathcal{H}} R_{Q}^{\alpha}({\bm h}). 
\end{equation*}

\textbf{Step 1.} It is easy to check that 
\begin{equation*}
\begin{split}
    R_{Q}^{\alpha}( \widehat{{\bm h}})-R_{Q}^{\alpha}({\bm h}_Q)=&R_{Q}^{\alpha}( \widehat{{\bm h}})-(1-\alpha)\widehat{R}_{S,T}^{\tau,\beta}(\widehat{\bm h})+(1-\alpha)\widehat{R}_{S,T}^{\tau,\beta}(\widehat{\bm h})-R_{Q}^{\alpha}({\bm h}_Q)\\ \leq &R_{Q}^{\alpha}( \widehat{{\bm h}})-(1-\alpha)\widehat{R}_{S,T}^{\tau,\beta}(\widehat{\bm h})+(1-\alpha)\widehat{R}_{S,T}^{\tau,\beta}({\bm h}_Q)-R_{Q}^{\alpha}({\bm h}_Q) \\ \leq & 2\sup_{{\bm h}\in \mathcal{H}} |(1-\alpha)\widehat{R}_{S,T}^{\tau,\beta}({\bm h})-R_{Q}^{\alpha}({\bm h})|,
    \end{split}
\end{equation*}
and
\begin{equation*}
\begin{split}
    R_{Q}^{\alpha}( \widehat{{\bm h}})-R_{Q}^{\alpha}({\bm h}_Q)=&R_{Q}^{\alpha}( \widehat{{\bm h}})-(1-\alpha)\widehat{R}_{S,T}^{\tau,\beta}({\bm h}_Q)+(1-\alpha)\widehat{R}_{S,T}^{\tau,\beta}({\bm h}_Q)-R_{Q}^{\alpha}({\bm h}_Q)\\ \geq &R_{Q}^{\alpha}( \widehat{{\bm h}})-(1-\alpha)\widehat{R}_{S,T}^{\tau,\beta}(\widehat{\bm h})+(1-\alpha)\widehat{R}_{S,T}^{\tau,\beta}({\bm h}_Q)-R_{Q}^{\alpha}({\bm h}_Q) \\ \geq & -2\sup_{{\bm h}\in \mathcal{H}} |(1-\alpha)\widehat{R}_{S,T}^{\tau,\beta}({\bm h})-R_{Q}^{\alpha}({\bm h})|,
    \end{split}
\end{equation*}
which implies that
\begin{equation*}
    |R_{Q}^{\alpha}( \widehat{{\bm h}})-R_{Q}^{\alpha}({\bm h}_Q)| \leq 2\sup_{{\bm h}\in \mathcal{H}} |(1-\alpha)\widehat{R}_{S,T}^{\tau,\beta}({\bm h})-R_{Q}^{\alpha}({\bm h})|.
\end{equation*}

Using the result of  Theorem 3, we obtain that
\begin{equation}\label{Eq::Th5-1-0}
    |R_{Q}^{\alpha}( \widehat{{\bm h}})-R_{Q}^{\alpha}({\bm h}_Q)| \leq 2 c\big(\max\{1,\frac{\beta}{\tau}\}+\beta\big) O_p(\lambda^{\frac{1}{2}}_{n,m})+2\gamma c \beta U(0<p/q\leq 2\tau).
\end{equation}

Then, using the result of Step 3 in the proof of Theorem 2, we obtain that
\begin{equation}\label{Eq::Th5-2-0}
    |R_{Q}^{\alpha}( \widehat{{\bm h}})-R_{P}^{\alpha}({\bm h}_Q)| \leq 2 c\big(\max\{1,\frac{\beta}{\tau}\}+\beta\big) O_p(\lambda^{\frac{1}{2}}_{n,m})+2\gamma c \beta U(0<p/q\leq 2\tau).
\end{equation}

\textbf{Step 2.} 
\begin{equation*}
\begin{split}
  \widehat{R}_{S,T}^{\tau,\beta}(\widehat{\bm h})&=  (1-\alpha) \widehat{R}_S(\widehat{\bm h})+(1-\alpha)\Delta_{S,T}^{\tau,\beta}(\widehat{\bm h})\\ &\leq (1-\alpha) \widehat{R}_S({\bm h}_Q)+(1-\alpha)\Delta_{S,T}^{\tau,\beta}({\bm h}_Q)\\ &\leq  R_{Q}^{\alpha}({\bm h}_Q) + c\big(\max\{1,\frac{\beta}{\tau}\}+\beta\big) O_p(\lambda^{\frac{1}{2}}_{n,m})+\gamma c \beta U(0<p/q\leq 2\tau){\rm~Using~Theorem~3}\\ & = (1-\alpha) \min_{{\bm h}\in \mathcal{H}} R_{Q,k}({\bm h})+c\big(\max\{1,\frac{\beta}{\tau}\}+\beta\big) O_p(\lambda^{\frac{1}{2}}_{n,m})+\gamma c \beta U(0<p/q\leq 2\tau)\\ &~{\rm Using~the~result~of~Step~3~in~proof~of~Theorem~2:}\min_{{\bm h}\in \mathcal{H}}R_Q^{\alpha}({\bm h})=(1-\alpha)\min_{{\bm h}\in \mathcal{H}}R_{Q,k}({\bm h})\\ & \leq (1-\alpha)  R_{Q,k}(\widehat{\bm h})+c\big(\max\{1,\frac{\beta}{\tau}\}+\beta\big) O_p(\lambda^{\frac{1}{2}}_{n,m})+\gamma c \beta U(0<p/q\leq 2\tau).
  \end{split}
\end{equation*}
Hence,
\begin{equation*}
\begin{split}
    &(1-\alpha)\Delta_{S,T}^{\tau,\beta}(\widehat{\bm h})\\  \leq &(1-\alpha)  R_{Q,k}(\widehat{\bm h})-(1-\alpha) \widehat{R}_S(\widehat{\bm h})+c\big(\max\{1,\frac{\beta}{\tau}\}+\beta\big) O_p(\lambda^{\frac{1}{2}}_{n,m})+\gamma c \beta U(0<p/q\leq 2\tau)\\ = &(1-\alpha)  R_{P,k}(\widehat{\bm h})-(1-\alpha) \widehat{R}_S(\widehat{\bm h})+c\big(\max\{1,\frac{\beta}{\tau}\}+\beta\big) O_p(\lambda^{\frac{1}{2}}_{n,m})+\gamma c \beta U(0<p/q\leq 2\tau)\\ \leq & (1-\alpha)c O_p(\lambda^{\frac{1}{2}}_{n,m})+c\big(\max\{1,\frac{\beta}{\tau}\}+\beta\big) O_p(\lambda^{\frac{1}{2}}_{n,m})+\gamma c \beta U(0<p/q\leq 2\tau)~{\rm Using~the~result~of~Lemma~\ref{lemma4}}\\ \leq  &2c\big(\max\{1,\frac{\beta}{\tau}\}+\beta\big) O_p(\lambda^{\frac{1}{2}}_{n,m})+\gamma c \beta U(0<p/q\leq 2\tau).
  \end{split}
\end{equation*}

Then, combining above inequality with the result of Step 1 in the proof of  Theorem 3, we obtain that
\begin{equation*}
\begin{split}
&    \max \{R_Q(\widehat{\bm h},\mathbf{y}_{C+1})-(1-\alpha)R_{P,k}(\widehat{\bm h},\mathbf{y}_{C+1}),0\}\\ \leq &3c\big(\max\{1,\frac{\beta}{\tau}\}+\beta\big) O_p(\lambda^{\frac{1}{2}}_{n,m})+2\gamma c \beta U(0<p/q\leq 2\tau).
    \end{split}
\end{equation*}

Because $\max \{R_Q(\widehat{\bm h},\mathbf{y}_{C+1})-(1-\alpha)R_{P,k}(\widehat{\bm h},\mathbf{y}_{C+1}),0\}=\alpha R_{Q,u}(\widehat{\bm h})$, we obtain that
\begin{equation*}
\begin{split}
&  \alpha R_{Q,u}(\widehat{\bm h}) \leq 3c\big(\max\{1,\frac{\beta}{\tau}\}+\beta\big) O_p(\lambda^{\frac{1}{2}}_{n,m})+2\gamma c \beta U(0<p/q\leq 2\tau).
    \end{split}
\end{equation*}

\textbf{Step 3.}
\begin{equation*}
\begin{split}
    &|R_{P}^{\alpha}( \widehat{{\bm h}})-R_{P}^{\alpha}({\bm h}_Q)|\\ \leq & |R_{P}^{\alpha}( \widehat{{\bm h}})- R_{Q}^{\alpha}( \widehat{{\bm h}})|+|R_{Q}^{\alpha}( \widehat{{\bm h}})-R_{P}^{\alpha}({\bm h}_Q)|\\ =& \alpha|R_{P,u}( \widehat{{\bm h}})- R_{Q,u}^{\alpha}( \widehat{{\bm h}})|+|R_{Q}^{\alpha}( \widehat{{\bm h}})-R_{P}^{\alpha}({\bm h}_Q)| \\  \leq & \alpha R_{Q,u}(\widehat{\bm h})+|R_{Q}^{\alpha}( \widehat{{\bm h}})-R_{P}^{\alpha}({\bm h}_Q)|  \\ \leq & 5 c\big(\max\{1,\frac{\beta}{\tau}\}+\beta\big) O_p(\lambda^{\frac{1}{2}}_{n,m})+4\gamma c \beta U(0<p/q\leq 2\tau)~{\rm Using~the~ results~ of~ Step~ 1~ and~ Step~ 2.}
    \end{split}
\end{equation*}
Briefly, we can write (absorbing 
coefficient $5$ into $O_p$)
\begin{equation*}
\begin{split}
    &|R_{P}^{\alpha}( \widehat{{\bm h}})-R_{P}^{\alpha}({\bm h}_Q)| \leq   c\big(\max\{1,\frac{\beta}{\tau}\}+\beta\big) O_p(\lambda^{\frac{1}{2}}_{n,m})+4\gamma c \beta U(0<p/q\leq 2\tau).
    \end{split}
\end{equation*}

Combining above inequality with Theorem 2, we obtain that
\begin{equation*}
\begin{split}
    &|R_{P}^{\alpha}( \widehat{{\bm h}})-\min_{{\bm h\in \mathcal{H}}}R_{P}^{\alpha}({\bm h})| \leq   c\big(\max\{1,\frac{\beta}{\tau}\}+\beta\big) O_p(\lambda^{\frac{1}{2}}_{n,m})+4\gamma c \beta U(0<p/q\leq 2\tau).
    \end{split}
\end{equation*}
\end{proof}
\subsection{Proof for Theorem 6}
\begin{lemma}\label{lemma6}
Assume the feature space $\mathcal{X}$ is compact and the loss function has an upper bound $c$. Let the RKHS $\mathcal{H}_K$ is the Hilbert space with Gaussian kernel. Suppose that the real density $p/q\in \mathcal{H}_K$ and set the regularization parameter $\lambda=\lambda_{n,m}$ in {\rm KuLSIF} such that
 \begin{equation*}
   \lim_{n,m\rightarrow 0}  \lambda_{n,m} =0,~~~\lambda_{n,m}^{-1}={O}({\min \{n,m\}}^{1-\delta}),
 \end{equation*}
 where $0<\delta<1$ is any constant, then
 \begin{equation*}
   \sup_{{\bm h}\in \mathcal{H}}|R_{Q,u}({\bm h})-\gamma' \widehat{R}^{\tau,\beta}_{S,T,u}({\bm h})|\leq \gamma' \beta c U(\{\mathbf{x}: ~0<r(\mathbf{x})\leq 2\tau\})+c\big(\max\{1,\frac{\beta}{\tau}\}+\beta\big) O_p(\lambda^{\frac{1}{2}}_{n,m}),
 \end{equation*}
 where $\gamma'=1/\big( \beta U(\{\mathbf{x}: ~r(\mathbf{x})=0\}) \big )$, and
 \begin{equation*}
 \begin{split}
  &R_{Q,u}({\bm h})=\int_{\mathcal{X}} \ell({\bm h}(\mathbf{x}),\mathbf{y}_{C+1}){\rm d}Q_{X|Y=\mathbf{y}_{C+1}}(\mathbf{x}),~~~ 
      \widehat{R}^{\tau,\beta}_{S,T,u}({\bm h}):=\frac{1}{m}\sum_{\mathbf{x}\in T} L_{\tau,\beta}({\widehat{w}(\mathbf{x}})) \ell({\bm h({\mathbf{x}})}, \mathbf{y}_{C+1}),
      \end{split}
 \end{equation*}
 here $Q_{X,Y}:=Q_{U}^{0,\beta}P_{Y|X}$, $\widehat{w}$ is the solution of {\rm KuLSIF}, and
 \begin{equation}\label{Lip2}
L_{\tau,\beta}^{-}(x)=\left\{
\begin{aligned}
~~~~~x+\beta,&~~~~~~~ x\leq \tau; \\
~~~~~0,&~~~~~~~2\tau \leq x;\\
~~~~~-\frac{\tau+\beta}{\tau}x+2\tau+2\beta,&~~~~~~~\tau<x<2\tau.
\end{aligned}
\right.
\end{equation}
\end{lemma}
\begin{proof}
\textbf{Step 1.} We claim that  
\begin{equation*}
\sup_{{\bm h}\in \mathcal{H}}|R_{Q,u}({\bm h})-\gamma' {R}^{\tau,\beta}_{U,u}({\bm h})|\leq \gamma' \beta c U(\{\mathbf{x}: ~0<r(\mathbf{x})\leq 2\tau\}),
\end{equation*}
where 
\begin{equation*}
   {R}^{\tau,\beta}_{U,u}({\bm h})=\int_{\mathcal{X}} L_{\tau,\beta}^{-}(r(\mathbf{x}))\ell({\bm h}(\mathbf{x}),\mathbf{y}_{C+1}){\rm d}U(\mathbf{x}),
\end{equation*}
here $r(\mathbf{x})=p(\mathbf{x})/q(\mathbf{x})$.

First, we note that
\begin{equation}\label{Eq::difference1}
\begin{split}
 &  \big | \int_{\mathcal{X}} L_{0,\beta}^{-}(r(\mathbf{x}))\ell({\bm h}(\mathbf{x}),\mathbf{y}_{C+1}){\rm d}U(\mathbf{x})-\int_{\mathcal{X}} L_{\tau,\beta}^{-}(r(\mathbf{x}))\ell({\bm h}(\mathbf{x}),\mathbf{y}_{C+1}){\rm d}U(\mathbf{x})|
\\ \leq &  \big | \int_{\mathcal{X}} L_{0,\beta}^{-}(r(\mathbf{x}))\ell({\bm h}(\mathbf{x}),\mathbf{y}_{C+1}){\rm d}U(\mathbf{x})-\int_{\mathcal{X}} L_{\tau,\beta}^{-}(r(\mathbf{x}))\ell({\bm h}(\mathbf{x}),\mathbf{y}_{C+1}){\rm d}U(\mathbf{x})|\\ \leq & 
c \int_{\mathcal{X}} \big |L_{0,\beta}^{-}(r(\mathbf{x}))- L_{\tau,\beta}^{-}(r(\mathbf{x}))\big |{\rm d}U(\mathbf{x})\\ \leq &c \int_{\{\mathbf{x}: ~0<r(\mathbf{x})\leq 2\tau\}}( \tau+\beta) {\rm d}U(\mathbf{x})=(\tau+\beta) c U(\{\mathbf{x}: ~0<r(\mathbf{x})\leq 2\tau\}).
   \end{split}
\end{equation}
Because $Q_{X,Y}=Q_U^{0,\beta}P_{Y|X}$, then according to the definition of $Q_U^{0,\beta}$, we know
\begin{equation*}
    R_{Q,u}({\bm h})=\gamma' \int_{\mathcal{X}} L_{0,\beta}^{-}(r(\mathbf{x}))\ell({\bm h}(\mathbf{x}),\mathbf{y}_{C+1}){\rm d}U(\mathbf{x}),
\end{equation*}
which implies
\begin{equation*}
    \sup_{{\bm h}\in \mathcal{H}}|R_{Q,u}({\bm h})-\gamma' {R}^{\tau,\beta}_{U,u}({\bm h})|\leq \gamma' (\tau+\beta) c U(\{\mathbf{x}: ~0<r(\mathbf{x})\leq 2\tau\}).
\end{equation*}

\textbf{Step 2.} We claim that
\begin{equation*}
   \sup_{{\bm h}\in \mathcal{H}}| {R}^{\tau,\beta}_{U,u}({\bm h})-\int_{\mathcal{X}}L_{\tau,\beta}^{-}(\widehat{w}(\mathbf{x}))\ell({\bm h}(\mathbf{x}),\mathbf{y}_{C+1}){\rm d}U(\mathbf{x})| \leq (c+\frac{c\beta}{\tau})O_p(\lambda^{\frac{1}{2}}_{n,m}).
\end{equation*}
First, the Lipschitz constant for $L_{\tau,\beta}^{-}$ is smaller than $1+\frac{\beta}{\tau}$.

Then,
\begin{equation*}
\begin{split}
&\sup_{{\bm h}\in \mathcal{H}}| {R}^{\tau,\beta}_{U,u}({\bm h})-\int_{\mathcal{X}}L_{\tau,\beta}^{-}(\widehat{w}(\mathbf{x}))\ell({\bm h}(\mathbf{x}),\mathbf{y}_{C+1}){\rm d}U(\mathbf{x})|\\=&\sup_{{\bm h}\in \mathcal{H}}| \int_{\mathcal{X}}L_{\tau,\beta}^{-}(r(\mathbf{x}))\ell({\bm h}(\mathbf{x}),\mathbf{y}_{C+1}){\rm d}U(\mathbf{x})-\int_{\mathcal{X}}L_{\tau,\beta}^{-}(\widehat{w}(\mathbf{x}))\ell({\bm h}(\mathbf{x}),\mathbf{y}_{C+1}){\rm d}U(\mathbf{x}) |\\ \leq &
   \sup_{{\bm h}\in \mathcal{H}}\int_{\mathcal{X}}|L_{\tau,\beta}^{-}(r(\mathbf{x}))-L_{\tau,\beta}^{-}(\widehat{w}(\mathbf{x}))|\ell({\bm h}(\mathbf{x}),\mathbf{y}_{C+1}){\rm d}U(\mathbf{x})\\ \leq & 
    \sup_{{\bm h}\in \mathcal{H}} \sqrt{\int_{\mathcal{X}} |L_{\tau,\beta}^{-}(r(\mathbf{x}))-L_{\tau,\beta}^{-}(\widehat{w}(\mathbf{x}))|^2 {\rm d}U(\mathbf{x})}\sqrt{\int_{\mathcal{X}} \ell^2({\bm h}(\mathbf{x}),\mathbf{y}_{C+1}) {\rm d}U(\mathbf{x})}~~{\rm H\ddot{o}lder~~ Inequality}\\ \leq & c\sup_{{\bm h}\in \mathcal{H}} \sqrt{\int_{\mathcal{X}} |L_{\tau,\beta}^{-}(r(\mathbf{x}))-L_{\tau,\beta}^{-}(\widehat{w}(\mathbf{x}))|^2 {\rm d}U(\mathbf{x})}\\ \leq & (c+\frac{c\beta}{\tau})\sup_{{\bm h}\in \mathcal{H}}\sqrt{\int_{\mathcal{X}} |r(\mathbf{x})-\widehat{w}(\mathbf{x})|^2 {\rm d}U(\mathbf{x})}.
   \end{split}
\end{equation*}
Lastly, using Lemma \ref{K2009}, 
\begin{equation*}
    \sup_{{\bm h}\in \mathcal{H}}| {R}^{\tau,\beta}_{U,u}({\bm h})-\int_{\mathcal{X}}L_{\tau,\beta}^{-}(\widehat{w}(\mathbf{x}))\ell({\bm h}(\mathbf{x}),\mathbf{y}_{C+1}){\rm d}U(\mathbf{x})|\leq (c+\frac{c\beta}{\tau})O_p(\lambda^{\frac{1}{2}}_{n,m}).
\end{equation*}

\textbf{Step 3.} We claim that 
\begin{equation*}
     \sup_{{\bm h}\in \mathcal{H}}\big | \frac{1}{m} \sum_{\mathbf{x}\in T} L_{\tau,\beta}^-(\widehat{w}(\mathbf{x}))\ell({\bm h}(\mathbf{x}),\mathbf{y}_{C+1})-\int_{\mathcal{X}}L_{\tau,\beta}^-(\widehat{w}(\mathbf{x}))\ell({\bm h}(\mathbf{x}),\mathbf{y}_{C+1}){\rm d}U(\mathbf{x})\big |\leq c\big(1+\frac{\beta}{\tau}+\beta\big) O_p(\lambda^{\frac{1}{2}}_{n,m}).
\end{equation*}

First, we set $\mathcal{F}_{B}:= \{L_{\tau,\beta}^{-}(w)\ell({\bm h},\mathbf{y}_{C+1}): w\in \mathcal{H}_K, \|w\|_K \leq B, {\bm h}\in \mathcal{H}\}$. We consider
\begin{equation*}
    \sup_{f \in \mathcal{F}_{B}}\big | \frac{1}{m} \sum_{\mathbf{x}\in T} f(\mathbf{x})-\int_{\mathcal{X}}f(\mathbf{x}){\rm d}U(\mathbf{x})\big |
\end{equation*}

Using Lemma \ref{4.1} and inequality \ref{Eq::fEstimation1}, it is easy to check that for $1-2\delta>0$, we have
\begin{equation}\label{Eq::fEstimation-1}
    \sup_{f\in \mathcal{F}_B}\big | \frac{1}{m} \sum_{\mathbf{x}\in T} f(\mathbf{x})-\int_{\mathcal{X}}f(\mathbf{x}){\rm d}U(\mathbf{x})\big |\leq 2 \widehat{\Re}_{T}(\mathcal{F}_B)+4(\tau+\beta)c\sqrt{\frac{2\log(4/\delta)}{m}},
\end{equation}
here we have used $|f|\leq (\tau+\beta)c$, for any $f\in \mathcal{F}_B$.

Then, we consider $\widehat{\Re}_{T}(\mathcal{F}_B)$.
\begin{equation*}
\begin{split}
   &m\widehat{\Re}_{T}(\mathcal{F}_B) \\=& \mathbb{E}_{\sigma}[\sup_{\|w\|_K\leq B, {\bm h}\in \mathcal{H}}\sum_{i=1}^m \sigma_i L_{\tau,\beta}^{-}(w_i)\ell({{\bm h}_i,\mathbf{y}_{C+1}}) ]\\ = &  \mathbb{E}_{\sigma}[\sup_{\|w\|_K\leq B, {\bm h}\in \mathcal{H}}\sigma_1 L_{\tau,\beta}^{-}(w_1)\ell({{\bm h}_1,\mathbf{y}_{C+1}})+\sum_{i=2}^m \sigma_i L_{\tau,\beta}^{-}(w_i)\ell({{\bm h}_i,\mathbf{y}_{C+1}}) ] \\ =&\frac{1}{2}\mathbb{E}_{\sigma_2,...,\sigma_m}[\sup_{\|w\|_K\leq B, {\bm h}\in \mathcal{H}} L_{\tau,\beta}^{-}(w_1)\ell({{\bm h}_1,\mathbf{y}_{C+1}})+\sum_{i=2}^m \sigma_i L_{\tau,\beta}^{-}(w_i)\ell({{\bm h}_i,\mathbf{y}_{C+1}})\\&~~~~~~~~~~~~~~~+ \sup_{\|w\|_K\leq B, {\bm h}\in \mathcal{H}}-L_{\tau,\beta}^{-}(w_1)\ell({{\bm h}_1,\mathbf{y}_{C+1}})+\sum_{i=2}^m \sigma_i L_{\tau,\beta}^{-}(w_i)\ell({{\bm h}_i,\mathbf{y}_{C+1}})]\\ = &\frac{1}{2}\mathbb{E}_{\sigma_2,...,\sigma_m}[\sup_{\|w\|_K\leq B, {\bm h}\in \mathcal{H}}\sup_{\|w'\|_K \leq B, {\bm h}'\in \mathcal{H}} L_{\tau,\beta}^{-}(w_1)\ell({{\bm h}_1,\mathbf{y}_{C+1}})+\sum_{i=2}^m \sigma_i L_{\tau,\beta}^{-}(w_i)\ell({{\bm h}_i,\mathbf{y}_{C+1}}) \\&~~~~~~~~~~~~~~~~~~~~~~~~~~~~~~~~~~~~~~~~~~~~~~~~~~~~~~~~~~~~~~-L_{\tau,\beta}^{-}(w_1')\ell({{\bm h}_1',\mathbf{y}_{C+1}})+\sum_{i=2}^m \sigma_i L_{\tau,\beta}^{-}(w_i')\ell({{\bm h}_i',\mathbf{y}_{C+1}})] \\ \leq &\frac{1}{2}\mathbb{E}_{\sigma_2,...,\sigma_m}[\sup_{\|w\|_K\leq B, {\bm h}\in \mathcal{H}}\sup_{\|w'\|_K \leq B, {\bm h}'\in \mathcal{H}} |L_{\tau,\beta}^{-}(w_1)-L_{\tau,\beta}^{-}(w_1')|\ell({{\bm h}_1',\mathbf{y}_{C+1}})+L_{\tau,\beta}^{-}(w_1)|\ell({{\bm h}_1',\mathbf{y}_{C+1}})-\ell({{\bm h}_1,\mathbf{y}_{C+1}})| \\&~~~~~~~~~~~~~~~~~~~~~~~~~~~~~~~~~~~~~~~~~~~~~~~~~~~~~~~~~~~~~~+\sum_{i=2}^m \sigma_i L_{\tau,\beta}^{-}(w_i)\ell({{\bm h}_i,\mathbf{y}_{C+1}})+\sum_{i=2}^m \sigma_i L_{\tau,\beta}^{-}(w_i')\ell({{\bm h}_i',\mathbf{y}_{C+1}})]\\ \leq &\frac{1}{2}\mathbb{E}_{\sigma_2,...,\sigma_m}[\sup_{\|w\|_K\leq B, {\bm h}\in \mathcal{H}}\sup_{\|w'\|_K \leq B, {\bm h}'\in \mathcal{H}} Lc|w_1-w_1'|+(B+\beta)|\ell({{\bm h}_1',\mathbf{y}_{C+1}})-\ell({{\bm h}_1,\mathbf{y}_{C+1}})| \\&~~~~~~~~~~~~~~~~~~~~~~~~~~~~~~~~~~~~~~~~~~~~~~~~~~~~~~~~~~~~~~+\sum_{i=2}^m \sigma_i L_{\tau,\beta}^{-}(w_i)\ell({{\bm h}_i,\mathbf{y}_{C+1}})+\sum_{i=2}^m \sigma_i L_{\tau,\beta}^{-}(w_i')\ell({{\bm h}_i',\mathbf{y}_{C+1}})] \\  =&\mathbb{E}_{\sigma}[\sup_{\|w\|_K\leq B, {\bm h}\in \mathcal{H}} Lc\sigma_1 w_1+(B+\beta)\sigma_1 \ell({{\bm h}_1,\mathbf{y}_{C+1}}) +\sum_{i=2}^m \sigma_i L_{\tau,\beta}^{-}(w_i)\ell({{\bm h}_i,\mathbf{y}_{C+1}}) ] \\ &{\rm~~Repeat ~~the~~ process~~} m-1 {\rm~~ times~~for~~} i=2,...,m. \\ \leq &\mathbb{E}_{\sigma}[\sup_{\|w\|_K\leq B, {\bm h}\in \mathcal{H}} \sum_{i=1}^m Lc\sigma_i w_i+\sum_{i=1}^m(B+\beta)\sigma_i \ell({{\bm h}_i,\mathbf{y}_{C+1}}) ]\\ \leq & mLc \widehat{\Re}_{T}(\mathcal{H}_{K,B}) +m(B+\beta) \widehat{\Re}_{T}(\mathcal{F}),
   \end{split}
\end{equation*}
where $w_i=w(\tilde{\mathbf{x}}_i), {\bm h}_i={\bm h}(\tilde{\mathbf{x}}_i)$, $L=1+\frac{\beta}{\tau}$, $\mathcal{H}_{K,B}=\{w: w\in \mathcal{H}_K,\|w\|_K\leq B\}$ and $\mathcal{F}=\{\ell({\bm h},\mathbf{y}_{C+1}):{\bm h}\in \mathcal{H}\}$.

According to Theorem 5.5 of \citeauthor{10.5555/2371238} (\citeyear{10.5555/2371238}), we obtain that
\begin{equation*}
    \widehat{\Re}_{T}(\mathcal{H}_{K,B}) \leq B \sqrt{\frac{1}{m}}.
\end{equation*}
According to the proving process of Lemma \ref{lemma4}, we obtain that
\begin{equation*}
    \widehat{\Re}_{T}(\mathcal{F}) \leq c\sqrt{\frac{4d\log m + 8d \log(C+1)}{m}},
\end{equation*}
where $d$ is the Natarajan Dimension of $\mathcal{H}$.

Hence,
\begin{equation*}
    \widehat{\Re}_{T}(\mathcal{F}_B) \leq BLc\sqrt{\frac{1}{m}}+(B+\beta)c\sqrt{\frac{4d\log m + 8d \log(C+1)}{m}}.
\end{equation*}

This implies that
for $1-2\delta>0$, we have
\begin{equation}\label{Eq::fEstimation2}
\begin{split}
    &\sup_{w\in \mathcal{H}_K,\|w\|_K\leq B}\sup_{{\bm h}\in \mathcal{H}}\big | \frac{1}{m} \sum_{\mathbf{x}\in T} L_{\tau,\beta}^{-}(w(\mathbf{x}))\ell({\bm h}(\mathbf{x}),\mathbf{y}_{C+1})-\int_{\mathcal{X}}L_{\tau,\beta}^{-}(w(\mathbf{x}))\ell({\bm h}(\mathbf{x}),\mathbf{y}_{C+1}){\rm d}U(\mathbf{x})\big |\\\leq &2 B(1+\frac{\beta}{\tau})c\sqrt{\frac{1}{m}}+2(B+\beta)c\sqrt{\frac{4d\log m + 8d \log(C+1)}{m}}+4(\tau+\beta)c\sqrt{\frac{2\log(4/\delta)}{m}}.
    \end{split}
\end{equation}

Because $\|\widehat{w}\|_K=O_p(1)$, then combining inequality \ref{Eq::fEstimation2}, we know that
\begin{equation*}
\begin{split}
   & \sup_{{\bm h}\in \mathcal{H}}\big | \frac{1}{m} \sum_{\mathbf{x}\in T} L_{\tau,\beta}^{-}(\widehat{w}(\mathbf{x}))\ell({\bm h}(\mathbf{x}),\mathbf{y}_{C+1})-\int_{\mathcal{X}}L_{\tau,\beta}^{-}(w(\mathbf{x}))\ell({\bm h}(\mathbf{x}),\mathbf{y}_{C+1}){\rm d}U(\mathbf{x})\big |\\\leq &2 O_p(1)(1+\frac{\beta}{\tau})cO_p(\sqrt{\frac{1}{m}})+2(O_p(1)+\beta)cO_p(\sqrt{\frac{4d\log m + 8d \log(C+1)}{m}})+4(\tau+\beta)cO_p(\sqrt{\frac{2\log(4/\delta)}{m}}).
    \end{split}
\end{equation*}
This implies that
\begin{equation*}
    \sup_{{\bm h}\in \mathcal{H}}\big | \frac{1}{m} \sum_{\mathbf{x}\in T} L_{\tau,\beta}^{-}(\widehat{w}(\mathbf{x}))\ell({\bm h}(\mathbf{x}),\mathbf{y}_{C+1})-\int_{\mathcal{X}}L_{\tau,\beta}^{-}(w(\mathbf{x}))\ell({\bm h}(\mathbf{x}),\mathbf{y}_{C+1}){\rm d}U(\mathbf{x})\big |\leq c\big(1+\frac{\beta}{\tau}+\tau+\beta\big) O_p(\lambda^{\frac{1}{2}}_{n,m}).
\end{equation*}

\textbf{Step 4.} Using the results of Steps 1, 2 and 3, we have
\begin{equation*}
\begin{split}
   &\sup_{{\bm h}\in \mathcal{H}}|R_{Q,u}({\bm h})-\gamma' \widehat{R}^{\tau,\beta}_{S,T,u}({\bm h})|\\ \leq &
   \sup_{{\bm h}\in \mathcal{H}}|R_{Q,u}({\bm h})-\gamma' {R}^{\tau,\beta}_{U,u}({\bm h})|+\sup_{{\bm h}\in \mathcal{H}}|\gamma' {R}^{\tau,\beta}_{U,u}({\bm h})-\gamma' \int_{\mathcal{X}}L_{\tau,\beta}^{-}(\widehat{w}(\mathbf{x}))\ell({\bm h}(\mathbf{x})){\rm d}U(\mathbf{x})|\\ + &  \sup_{{\bm h}\in \mathcal{H}}|\gamma' \int_{\mathcal{X}}L_{\tau,\beta}^-(\widehat{w}(\mathbf{x}))\ell({\bm h}(\mathbf{x}),\mathbf{y}_{C+1}){\rm d}U(\mathbf{x})-\gamma' \widehat{R}^{\tau,\beta}_{S,T,u}({\bm h},\mathbf{y}_{C+1})| \\ \leq &
   \gamma' \beta c U(\{\mathbf{x}: ~0<r(\mathbf{x})\leq 2\tau\})+\gamma' (c+\frac{c\beta}{\tau})O_p(\lambda^{\frac{1}{2}}_{n,m})+ c\gamma'\big(1+\frac{\beta}{\tau}+\tau+\beta\big) O_p(\lambda^{\frac{1}{2}}_{n,m}).
 \end{split}
 \end{equation*}
 
 We can write 
 \begin{equation*}
     \sup_{{\bm h}\in \mathcal{H}}|R_{Q,u}({\bm h})-\gamma' \widehat{R}^{\tau,\beta}_{S,T,u}({\bm h})|\leq \gamma' \beta c U(\{\mathbf{x}: ~0<r(\mathbf{x})\leq 2\tau\})+c\gamma'\big(1+\frac{\beta}{\tau}+\tau+\beta\big) O_p(\lambda^{\frac{1}{2}}_{n,m}).
 \end{equation*}
\end{proof}
\newpage
\begin{proof}[Proof of Theorem 6]
Assume that
\begin{equation*}
    \widetilde{\bm h} \in \argmin_{{\bm h}\in \mathcal{H}} \widetilde{R}_{S,T}^{\tau,\beta}({\bm h}),~~~{\bm h}_Q \in \argmin_{{\bm h}\in \mathcal{H}} R_{Q}^{\alpha}({\bm h}). 
\end{equation*}

\textbf{Step 1.} It is easy to check that 
\begin{equation*}
\begin{split}
    R_{Q}^{\alpha}( \widetilde{\bm h})-R_{Q}^{\alpha}({\bm h}_Q)=&R_{Q}^{\alpha}( \widetilde{\bm h})-\widetilde{R}_{S,T}^{\tau,\beta}(\widetilde{\bm h})+\widetilde{R}_{S,T}^{\tau,\beta}(\widetilde{\bm h})-R_{Q}^{\alpha}({\bm h}_Q)\\ \leq &R_{Q}^{\alpha}( \widetilde{\bm h})-\widetilde{R}_{S,T}^{\tau,\beta}(\widetilde{\bm h})+\widetilde{R}_{S,T}^{\tau,\beta}({\bm h}_Q)-R_{Q}^{\alpha}({\bm h}_Q) \\ \leq & 2\sup_{{\bm h}\in \mathcal{H}} |\widetilde{R}_{S,T}^{\tau,\beta}({\bm h})-R_{Q}^{\alpha}({\bm h})|,
    \end{split}
\end{equation*}
and
\begin{equation*}
\begin{split}
    R_{Q}^{\alpha}( \widetilde{\bm h})-R_{Q}^{\alpha}({\bm h}_Q)=&R_{Q}^{\alpha}( \widetilde{\bm h})-\widetilde{R}_{S,T}^{\tau,\beta}({\bm h}_Q)+\widetilde{R}_{S,T}^{\tau,\beta}({\bm h}_Q)-R_{Q}^{\alpha}({\bm h}_Q)\\ \geq &R_{Q}^{\alpha}( \widetilde{\bm h})-\widetilde{R}_{S,T}^{\tau,\beta}(\widetilde{\bm h})+\widetilde{R}_{S,T}^{\tau,\beta}({\bm h}_Q)-R_{Q}^{\alpha}({\bm h}_Q) \\ \geq & -2\sup_{{\bm h}\in \mathcal{H}} |\widetilde{R}_{S,T}^{\tau,\beta}({\bm h})-R_{Q}^{\alpha}({\bm h})|,
    \end{split}
\end{equation*}
which implies that
\begin{equation*}
    |R_{Q}^{\alpha}( \widetilde{\bm h})-R_{Q}^{\alpha}({\bm h}_Q)| \leq 2\sup_{{\bm h}\in \mathcal{H}} |\widetilde{R}_{S,T}^{\tau,\beta}({\bm h})-R_{Q}^{\alpha}({\bm h})|.
\end{equation*}

Using the result of  Lemma \ref{lemma6} and Lemma \ref{lemma4}, we obtain that
\begin{equation}\label{Eq::Th5-1}
    |R_{Q}^{\alpha}( \widetilde{\bm h})-R_{Q}^{\alpha}({\bm h}_Q)| \leq 2 c\gamma'\big(1+\tau+\frac{\beta}{\tau}+\beta\big) O_p(\lambda^{\frac{1}{2}}_{n,m})+2\gamma' c\alpha \beta U(0<p/q\leq 2\tau).
\end{equation}

Then, using the result of Step 3 in the proof of Theorem 2, we obtain that
\begin{equation}\label{Eq::Th5-2}
    |R_{Q}^{\alpha}( \widetilde{\bm h})-R_{P}^{\alpha}({\bm h}_Q)| \leq 2 c\gamma'\big(1+\tau+\frac{\beta}{\tau}+\beta\big) O_p(\lambda^{\frac{1}{2}}_{n,m})+2\gamma' c\alpha \beta U(0<p/q\leq 2\tau).
\end{equation}

\textbf{Step 2.} 
\begin{equation*}
\begin{split}
  \widetilde{R}_{S,T}^{\tau,\beta}(\widetilde{\bm h})&=  (1-\alpha) \widehat{R}_S(\widetilde{\bm h})+\alpha \gamma' \widehat{R}_{S,T,u}^{\tau,\beta}(\widetilde{\bm h})\\ &\leq (1-\alpha) \widehat{R}_S({\bm h}_Q)+\alpha \gamma' \widehat{R}_{S,T,u}^{\tau,\beta}({\bm h}_Q)\\ &\leq  R_{Q}^{\alpha}({\bm h}_Q) + c\gamma'\big(1+\tau+\frac{\beta}{\tau}+\beta\big) O_p(\lambda^{\frac{1}{2}}_{n,m})+\gamma' c\alpha \beta U(0<p/q\leq 2\tau){\rm~Using~ Lemma~ \ref{lemma6}~and~ Lemma~ \ref{lemma4}}\\ & = (1-\alpha) \min_{{\bm h}\in \mathcal{H}} R_{Q,k}({\bm h}) + c\gamma'\big(1+\tau+\frac{\beta}{\tau}+\beta\big) O_p(\lambda^{\frac{1}{2}}_{n,m})+\gamma' c\alpha \beta U(0<p/q\leq 2\tau)\\ &~{\rm Using~the~result~of~Step~3~in~proof~of~Theorem~2:}\min_{{\bm h}\in \mathcal{H}}R_Q^{\alpha}({\bm h})=(1-\alpha)\min_{{\bm h}\in \mathcal{H}}R_{Q,k}({\bm h})\\ & \leq (1-\alpha)  R_{Q,k}(\widetilde{\bm h}) + c\gamma'\big(1+\tau+\frac{\beta}{\tau}+\beta\big) O_p(\lambda^{\frac{1}{2}}_{n,m})+\gamma' c\alpha \beta U(0<p/q\leq 2\tau).
  \end{split}
\end{equation*}
Hence,
\begin{equation*}
\begin{split}
    &\alpha \gamma' \widehat{R}_{S,T,u}^{\tau,\beta}(\widetilde{\bm h})\\  \leq &(1-\alpha)  R_{Q,k}(\widetilde{\bm h})-(1-\alpha) \widehat{R}_S(\widetilde{\bm h})+ c\gamma'\big(1+\tau+\frac{\beta}{\tau}+\beta\big) O_p(\lambda^{\frac{1}{2}}_{n,m})+\gamma' c\alpha \beta U(0<p/q\leq 2\tau)\\ = &(1-\alpha)  R_{P,k}(\widetilde{\bm h})-(1-\alpha) \widehat{R}_S(\widetilde{\bm h})+ c\gamma'\big(1+\tau+\frac{\beta}{\tau}+\beta\big) O_p(\lambda^{\frac{1}{2}}_{n,m})+\gamma' c\alpha \beta U(0<p/q\leq 2\tau)\\ \leq & (1-\alpha)c O_p(\lambda^{\frac{1}{2}}_{n,m})+ c\gamma'\big(1+\tau+\frac{\beta}{\tau}+\beta\big) O_p(\lambda^{\frac{1}{2}}_{n,m})+\gamma' c\alpha \beta U(0<p/q\leq 2\tau)~{\rm Using~the~result~of~Lemma~\ref{lemma4}}\\ \leq  & 2c\gamma'\big(1+\tau+\frac{\beta}{\tau}+\beta\big) O_p(\lambda^{\frac{1}{2}}_{n,m})+\gamma' c\alpha \beta U(0<p/q\leq 2\tau).
  \end{split}
\end{equation*}

Then, combining the above inequality with the result of Lemma \ref{lemma6}, we obtain that
\begin{equation*}
\begin{split}
&  \alpha R_{Q,u}(\widetilde{\bm h}) \leq 3c\gamma'\big(1+\tau+\frac{\beta}{\tau}+\beta\big) O_p(\lambda^{\frac{1}{2}}_{n,m})+2\gamma' c\alpha \beta U(0<p/q\leq 2\tau).
    \end{split}
\end{equation*}

\textbf{Step 3.}
\begin{equation*}
\begin{split}
    &|R_{P}^{\alpha}( \widetilde{\bm h})-R_{P}^{\alpha}({\bm h}_Q)|\\ \leq & |R_{P}^{\alpha}( \widetilde{\bm h})- R_{Q}^{\alpha}( \widetilde{\bm h})|+|R_{Q}^{\alpha}( \widetilde{\bm h})-R_{P}^{\alpha}({\bm h}_Q)|\\ =& \alpha|R_{P,u}( \widetilde{\bm h})- R_{Q,u}^{\alpha}( \widetilde{\bm h})|+|R_{Q}^{\alpha}( \widetilde{\bm h})-R_{P}^{\alpha}({\bm h}_Q)| \\  \leq & \alpha R_{Q,u}(\widetilde{\bm h})+|R_{Q}^{\alpha}( \widetilde{\bm h})-R_{P}^{\alpha}({\bm h}_Q)|  \\ \leq & 5c\gamma'\big(1+\tau+\frac{\beta}{\tau}+\beta\big) O_p(\lambda^{\frac{1}{2}}_{n,m})+4\gamma' c\alpha \beta U(0<p/q\leq 2\tau)~{\rm Using~the~ results~ of~ Step~ 1~ and~ Step~ 2.}
    \end{split}
\end{equation*}
Briefly, we can write (absorbing 
coefficient $5$ into $O_p$)
\begin{equation*}
\begin{split}
    &|R_{P}^{\alpha}( \widetilde{\bm h})-R_{P}^{\alpha}({\bm h}_Q)| \leq   c\gamma'\big(1+\tau+\frac{\beta}{\tau}+\beta\big) O_p(\lambda^{\frac{1}{2}}_{n,m})+4\gamma' c\alpha \beta U(0<p/q\leq 2\tau).
    \end{split}
\end{equation*}

Combining the above inequality with Theorem 2, we obtain that
\begin{equation*}
\begin{split}
    &|R_{P}^{\alpha}( \widetilde{{\bm h}})-\min_{{\bm h\in \mathcal{H}}}R_{P}^{\alpha}({\bm h})|\leq   c\gamma'\big(1+\tau+\frac{\beta}{\tau}+\beta\big) O_p(\lambda^{\frac{1}{2}}_{n,m})+4\gamma' c\alpha \beta U(0<p/q\leq 2\tau).
    \end{split}
\end{equation*}
\end{proof}
\newpage
\section{Appendix F: Details on Experiments}\label{App6}

\subsection{Datasets}

    $\bullet$ MNIST dataset \cite{lecun-mnisthandwrittendigit-2010}.  The MNIST\footnote{http://yann.lecun.com/exdb/mnist/} database of handwritten digits, has a training set of $60,000$ samples, and a testing set of $10,000$ samples. The digits have been size-normalized and centered in a fixed-size image. Following the set up in \citeauthor{DBLP:conf/cvpr/YoshihashiSKYIN19} (\citeyear{DBLP:conf/cvpr/YoshihashiSKYIN19}), we use MNIST \cite{lecun-mnisthandwrittendigit-2010} as the training samples and use Omniglot \cite{ager2008omniglot}, MNIST-Noise, and Noise these datasets as unknown classes. Omniglot contains alphabet characters. Noise is synthesized by sampling each pixel value from a uniform distribution on $[0, 1]$ (i.i.d). MNIST-Noise is synthesized by adding noise on MNIST testing samples. Each dataset has $10, 000$ testing samples.

\begin{figure*}[th]
    \centering 
    \subfigure{\includegraphics[width=0.1\textwidth]{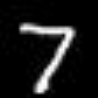}}
    \subfigure{\includegraphics[width=0.1\textwidth]{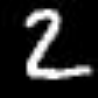}}
    \subfigure{\includegraphics[width=0.1\textwidth]{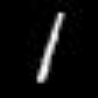}}
    \subfigure{\includegraphics[width=0.1\textwidth]{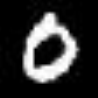}}
    \subfigure{\includegraphics[width=0.1\textwidth]{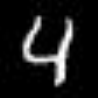}}
    \subfigure{\includegraphics[width=0.1\textwidth]{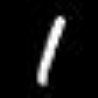}}
    \caption{MNIST.}
    \centering 
    \subfigure{\includegraphics[width=0.1\textwidth]{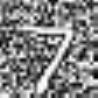}}
    \subfigure{\includegraphics[width=0.1\textwidth]{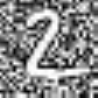}}
    \subfigure{\includegraphics[width=0.1\textwidth]{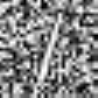}}
    \subfigure{\includegraphics[width=0.1\textwidth]{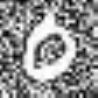}}
    \subfigure{\includegraphics[width=0.1\textwidth]{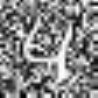}}
    \subfigure{\includegraphics[width=0.1\textwidth]{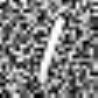}}
    \caption{MNIST-Noise.}
    \centering 
    \subfigure{\includegraphics[width=0.1\textwidth]{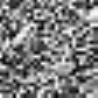}}
    \subfigure{\includegraphics[width=0.1\textwidth]{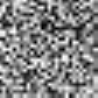}}
    \subfigure{\includegraphics[width=0.1\textwidth]{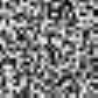}}
    \subfigure{\includegraphics[width=0.1\textwidth]{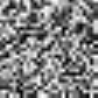}}
    \subfigure{\includegraphics[width=0.1\textwidth]{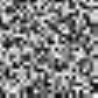}}
    \subfigure{\includegraphics[width=0.1\textwidth]{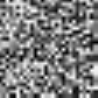}}
    \caption{Noise.}
     \centering 
    \subfigure{\includegraphics[width=0.1\textwidth]{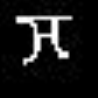}}
    \subfigure{\includegraphics[width=0.1\textwidth]{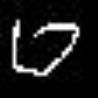}}
    \subfigure{\includegraphics[width=0.1\textwidth]{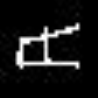}}
    \subfigure{\includegraphics[width=0.1\textwidth]{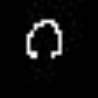}}
    \subfigure{\includegraphics[width=0.1\textwidth]{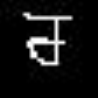}}
    \subfigure{\includegraphics[width=0.1\textwidth]{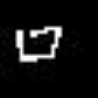}}
    \caption{Omniglot.}
\end{figure*}
\begin{table}[h]\label{MNIST}
\caption{{Introduction of MNIST Dataset in Open-set learning.}}
\begin{center}
\begin{tabular}{p{2.1cm}lp{1cm}lp{1.3cm}lp{2cm}lp{2cm}lp{1.5cm}}
\hline \hline
Dataset  &\#Sample&\#Class &Known/Unknown& Train/Test   \\ \hline
MNIST  &~~60,000&~10&Known Classes&Train\\
MNIST  &~~10,000&~10&Known Classes&Test\\
MNIST-Noise &~~10,000&~10&Unknown Classes&Test\\
Omniglot  &~~10,000&1,623&Unknown Classes&Test\\
Noise &~~10,000&~~1&Unknown Classes&Test\\
\hline
\hline
\end{tabular}
\end{center}
\end{table}
\newpage

    $\bullet$ CIFAR-$10$ dataset.  The CIFAR-$10$ dataset consists of $60,000$ $32\times32$ colour images in $10$ classes, with $6,000$ images per class. There are $50,000$ training images and $10,000$ testing images. 
 Following the set up in \citeauthor{DBLP:conf/cvpr/YoshihashiSKYIN19} (\citeyear{DBLP:conf/cvpr/YoshihashiSKYIN19}), we use the training samples from CIFAR-$10$ \cite{krizhevsky2010convolutional} as training samples in open-set learning problem. We collect unknown samples from datasets ImageNet and LSUN. Similar to \citeauthor{DBLP:conf/cvpr/YoshihashiSKYIN19} (\citeyear{DBLP:conf/cvpr/YoshihashiSKYIN19}), we resized or cropped them so that they would have the same sizes with known samples. Hence, we generated four datasets ImageNet-crop, ImageNet-resize, LSUN-crop and LSUN-resize as unknown classes.
\begin{figure*}[th]
    \centering 
    \subfigure{\includegraphics[width=0.1\textwidth]{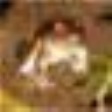}}
    \subfigure{\includegraphics[width=0.1\textwidth]{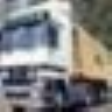}}
    \subfigure{\includegraphics[width=0.1\textwidth]{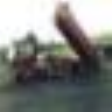}}
    \subfigure{\includegraphics[width=0.1\textwidth]{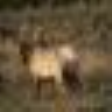}}
    \subfigure{\includegraphics[width=0.1\textwidth]{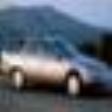}}
    \subfigure{\includegraphics[width=0.1\textwidth]{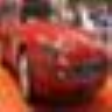}}
    \caption{CIFAR-10.}
    \centering 
    \subfigure{\includegraphics[width=0.1\textwidth]{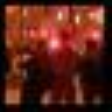}}
    \subfigure{\includegraphics[width=0.1\textwidth]{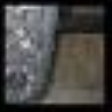}}
    \subfigure{\includegraphics[width=0.1\textwidth]{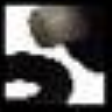}}
    \subfigure{\includegraphics[width=0.1\textwidth]{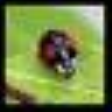}}
    \subfigure{\includegraphics[width=0.1\textwidth]{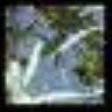}}
    \subfigure{\includegraphics[width=0.1\textwidth]{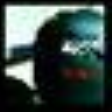}}
    \caption{ImageNet-crop.}
    \centering 
    \subfigure{\includegraphics[width=0.1\textwidth]{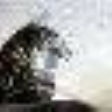}}
    \subfigure{\includegraphics[width=0.1\textwidth]{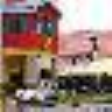}}
    \subfigure{\includegraphics[width=0.1\textwidth]{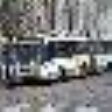}}
    \subfigure{\includegraphics[width=0.1\textwidth]{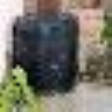}}
    \subfigure{\includegraphics[width=0.1\textwidth]{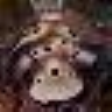}}
    \subfigure{\includegraphics[width=0.1\textwidth]{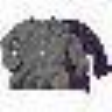}}
    \caption{ImageNet-resize.}
     \centering 
    \subfigure{\includegraphics[width=0.1\textwidth]{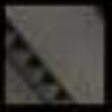}}
    \subfigure{\includegraphics[width=0.1\textwidth]{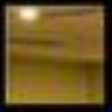}}
    \subfigure{\includegraphics[width=0.1\textwidth]{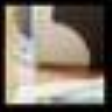}}
    \subfigure{\includegraphics[width=0.1\textwidth]{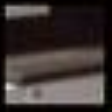}}
    \subfigure{\includegraphics[width=0.1\textwidth]{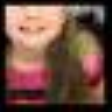}}
    \subfigure{\includegraphics[width=0.1\textwidth]{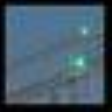}}
    \caption{LSUN-crop.}
     \centering 
    \subfigure{\includegraphics[width=0.1\textwidth]{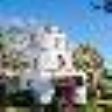}}
    \subfigure{\includegraphics[width=0.1\textwidth]{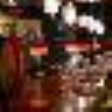}}
    \subfigure{\includegraphics[width=0.1\textwidth]{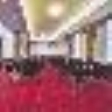}}
    \subfigure{\includegraphics[width=0.1\textwidth]{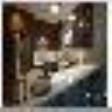}}
    \subfigure{\includegraphics[width=0.1\textwidth]{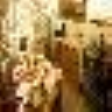}}
    \subfigure{\includegraphics[width=0.1\textwidth]{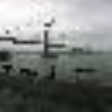}}
    \caption{LSUN-resize.}
\end{figure*}
\begin{table}[!]\label{MNIST100}
\caption{{Introduction of CIFAR-10 Dataset in Open-set learning.}}
\begin{center}
\begin{tabular}{p{2.4cm}lp{1cm}lp{1.3cm}lp{2cm}lp{2cm}lp{1.5cm}}
\hline \hline
Dataset  &\#Sample&\#Class &Known/Unknown& Train/Test   \\ \hline
CIFAR-10  &~~50,000&~10&Known Classes&Train\\
CIFAR-10  &~~10,000&~10&Known Classes&Test\\
ImageNet-crop &~~10,000&1,000&Unknown Classes&Test\\
ImageNet-resize  &~~10,000&1,000&Unknown Classes&Test\\
LSUN-crop &~~10,000&~10&Unknown Classes&Test\\
LSUN-resize &~~10,000&~10&Unknown Classes&Test\\
\hline
\hline
\end{tabular}
\end{center}
\end{table}

\subsection{Network Architecture and Experimental Setup}

All details can be found in \httpsurl{github.com/Anjin-Liu/Openset_Learning_AOSR}.

\subsection{Parameter Analysis and Influence of Model Capacity  }
\begin{figure*}[th]
    \centering 
    \subfigure[Parameter Analysis for $m$]{\includegraphics[width=0.45\textwidth]{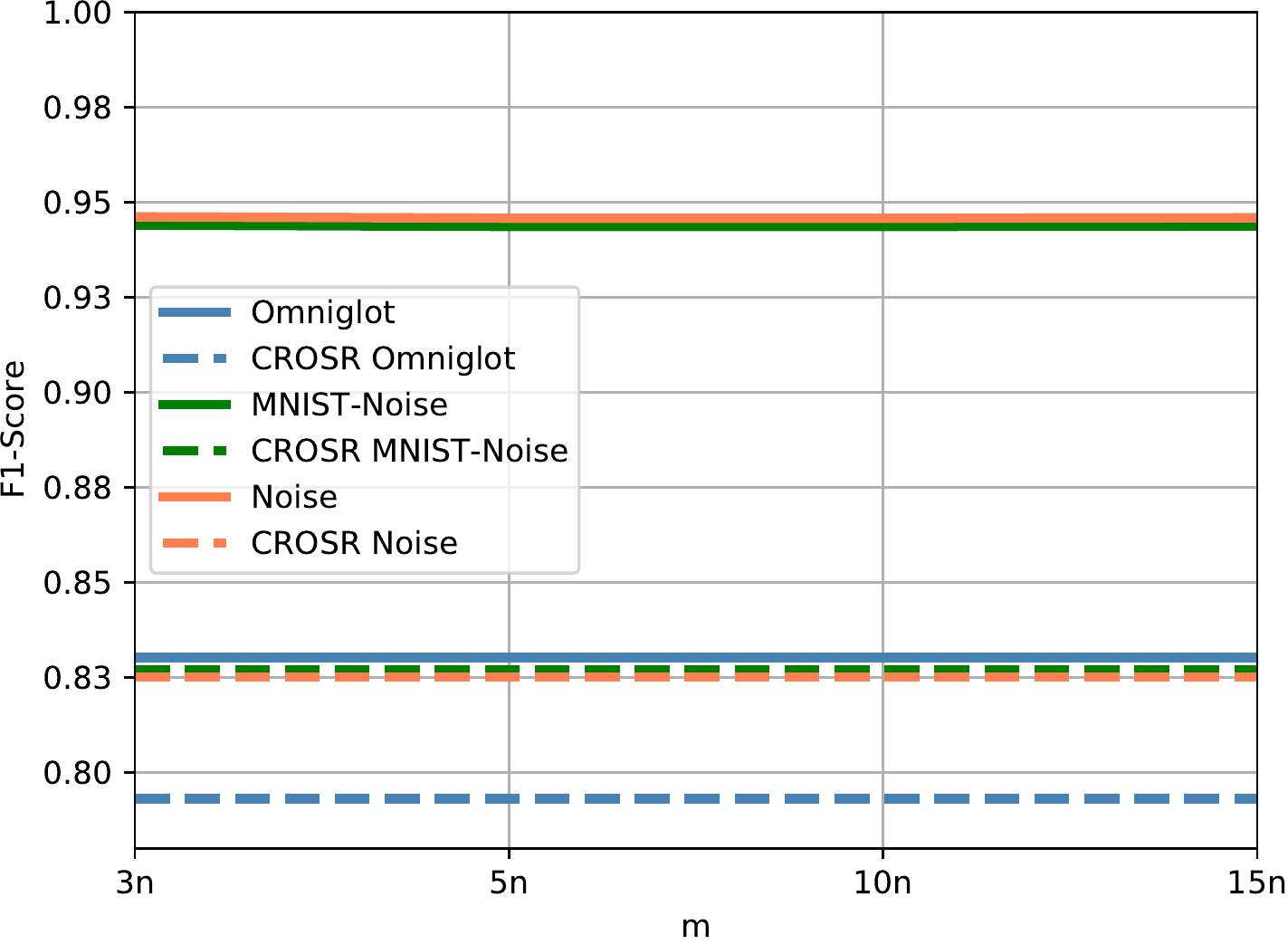}}
 ~~~   \subfigure[Influence of Model Capacity ]{\includegraphics[width=0.45\textwidth]{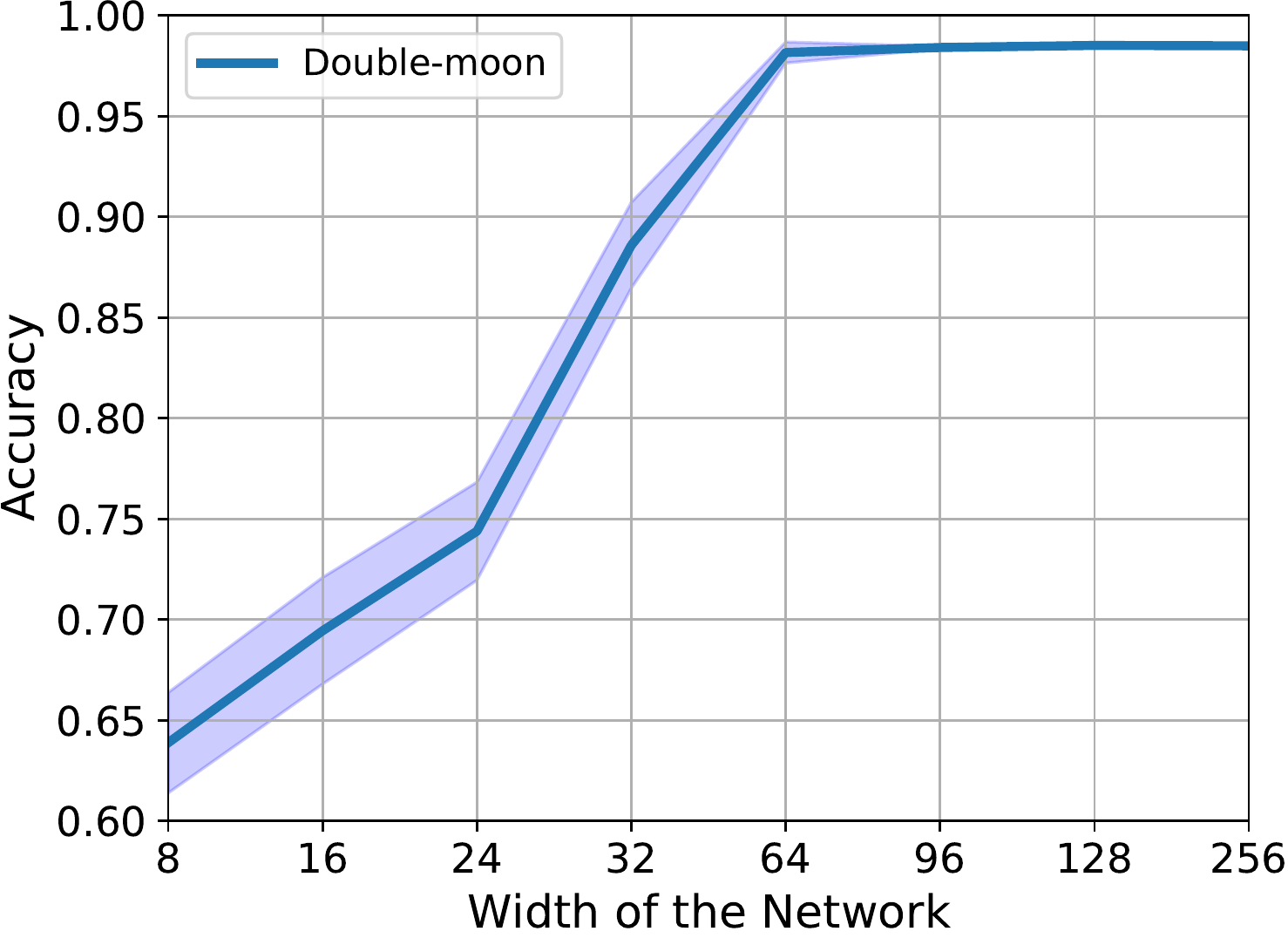}}
 \caption{Parameter Analysis and Influence of Model Capacity}\label{Ex}
\end{figure*}   

Experiment results on parameter $m$ are shown in Figure \ref{Ex} (a). $m$ is the size of generated samples $T$. We set $m=3n,5n,10n$ and $15n$. By changing $m$ in the range of $3n,5n,10n,15n$, AOSR achieves consistent performance. This result can be explained by our theory. Because when $m>n$, the increases of $m$ does not influence the error bound in Theorem 6. 

Experiment results on the width of the network are shown in Figure \ref{Ex} (b). We generate $2,000$ training samples and adjust the width for the  second
to the last layer from $8$ to $256$. For different width, we run $100$ times and report the mean accuracy and standard error. As increasing the network's width from $8$ to $256$, the accuracy of double-moon increases. When the width is larger than $64$, the performance achieves a stable performance. This means the model capacity has a profound impact on the performance of OSL. Generally, the larger the model capacity is, the better the model's performance is. This is because a larger hypothesis space $\mathcal{H}$ has a greater possibility to meet the conditions of Assumption 1 (realization assumption for unknown classes).

\bibliographystyle{icml2021}

\bibliography{Supp.bbl}

%% file: algorihtm.tex
\textbf{Step 1 (Feature Encoding).} Train the samples $S$ to get a closed-set classifier ${\bm h}_{{\bm \Theta}_0}$, and {designate the output of second to the last layer (without softmax) ${\bm l}$ of ${\bm h}_{{\bm \Theta}_0}$} as the encoded feature vector, i.e., $X_{\textrm{encoder}}={\bm l}(X)$. The new encoded feature space is denoted as $\mathcal{X}_{\textrm{encoder}}$.

\textbf{Step 2 (Initialize the Auxiliary Domain).} Randomly generate samples $T$ from space $\mathcal{X}_{\textrm{encoder}}$. By default, we generate $T$ by uniform distribution and set the size $m$ is $3n$. {We update the samples $S=\{({\bm l}(\mathbf{x}),\mathbf{y}):(\mathbf{x},\mathbf{y}) \in S\}$.}

\textbf{Step 3 (Construct the Auxiliary Domain).} Estimate the weights $\widehat{w}$ with samples $S$ and $T$ as the input. The higher the weight is, the more likely a generated sample belongs to the known classes. The parameters selection details are shown as follows. 

Weight estimation algorithm: In the theoretical part, KuLSIF is selected to estimate weights. Kernel mean matching (KMM) \cite{DBLP:journals/jmlr/GrettonBRSS12} is also an alternative solution \cite{DBLP:conf/alt/CortesMRR08}. However, in practice, KuLSIF and KMM have time complexity $O((m+n)^2)$ \cite{DBLP:journals/ml/KanamoriSS12}, which is not suitable for large datasets. The kernel bandwidth selection also impacts the overall performance \cite{liu2020learning}. Thus, we recommend using the outlier sample score (with range $[0, 1]$) given by isolation forest (iForest) \cite{liu2008isolation} as the sample weights, which has time complexity ${O}((n+m) \log (n+m))$. Close to $1$ means known classes while close to $0$ means unknown classes. 
    
The $\tau$ is a threshold to split the generated samples $T$ into known and unknown samples. Considering we are using iForest, based on the predicted sample score $[s_1,...,s_m]$ (descending order), we set $\tau=s_{[t*m]}$, where $t\in (0,1)$ is the proportion that the generated samples selected as unknown samples. We set $t=10\%$ as default.

    
{The $\beta$ and $\mu$ control jointly the importance of correctly classified unknown samples. We set $\mu$ as a dynamical parameter depending on $\beta$: $\mu=\frac{ n \beta}{n'+0.0001}$, where $n'$  is number of samples in training samples actually predicted as unknown. For example, if $\beta=0.05$, $n$ is $1000$, there are $10$ samples in training samples are predicted as unknown, then $\mu\approx5$.}
%

\textbf{Step 4 ($\mathbf{Softmax}_{C+1}$).} Initialize an open-set learning neural network with samples $S$ and $T$ as the input and $C+1$ Softmax \cite{qin2019rethinking} nodes as the output.

\textbf{Step 5 (Open-set Learning).} Train the $\mathbf{Softmax}_{C+1}$ neural network with the cost function defined in Eq. \eqref{optimial problem1} with both $S$ and $T$.

%% file: sec5.tex
\section{Experiments and Results}
\begin{figure*}[t]
    \centering 
    \subfigure[Double-moon dataset]{\includegraphics[scale=0.17, trim=0  270 0 300, clip]{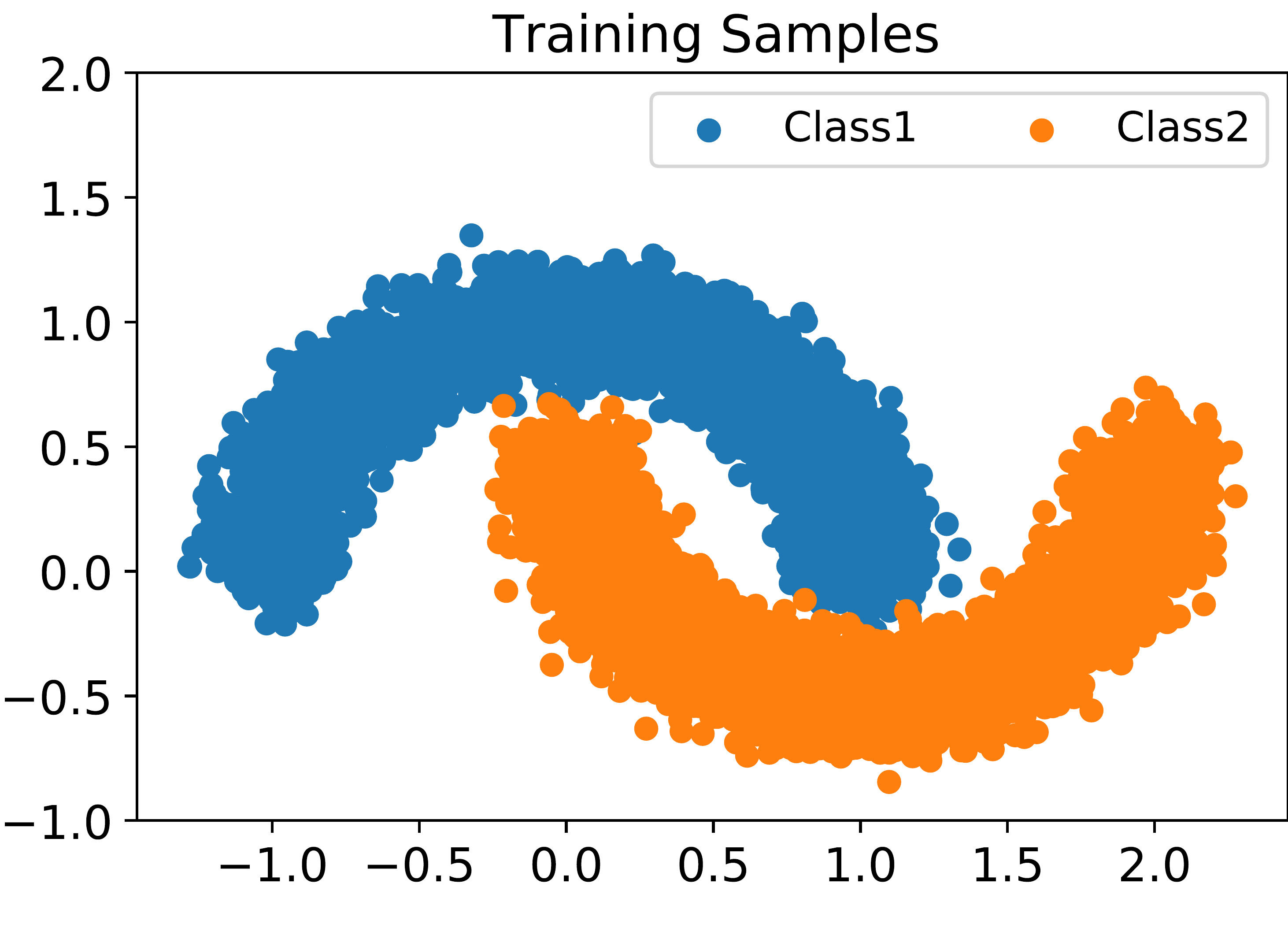}}~~~~~~~
    \subfigure[ Closed-set classification]{\includegraphics[scale=0.17, trim=0  270 0 300, clip]{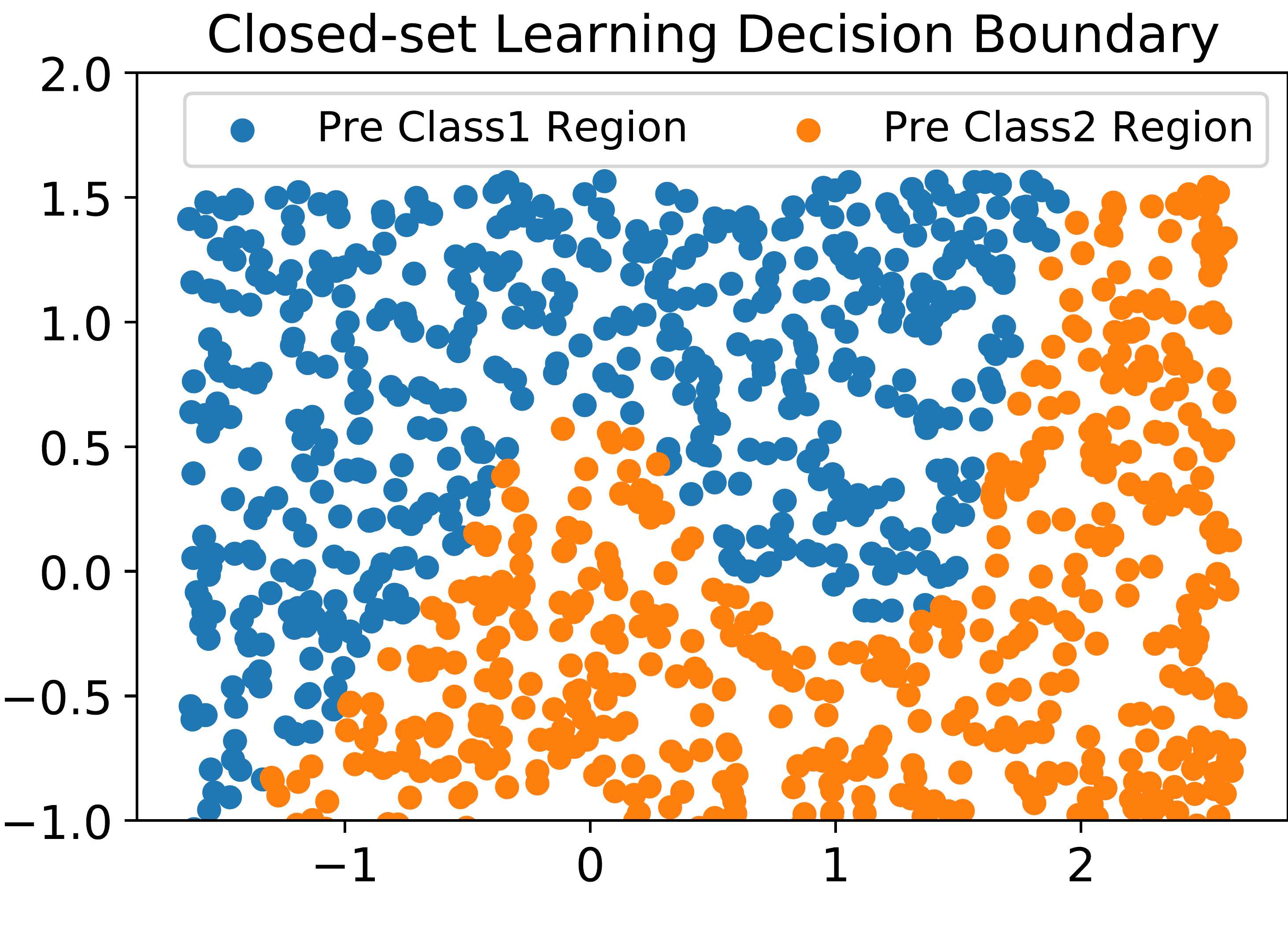}}~~~~~~~
    \subfigure[Open-set classification]{\includegraphics[scale=0.17, trim=0  270 0 300, clip]{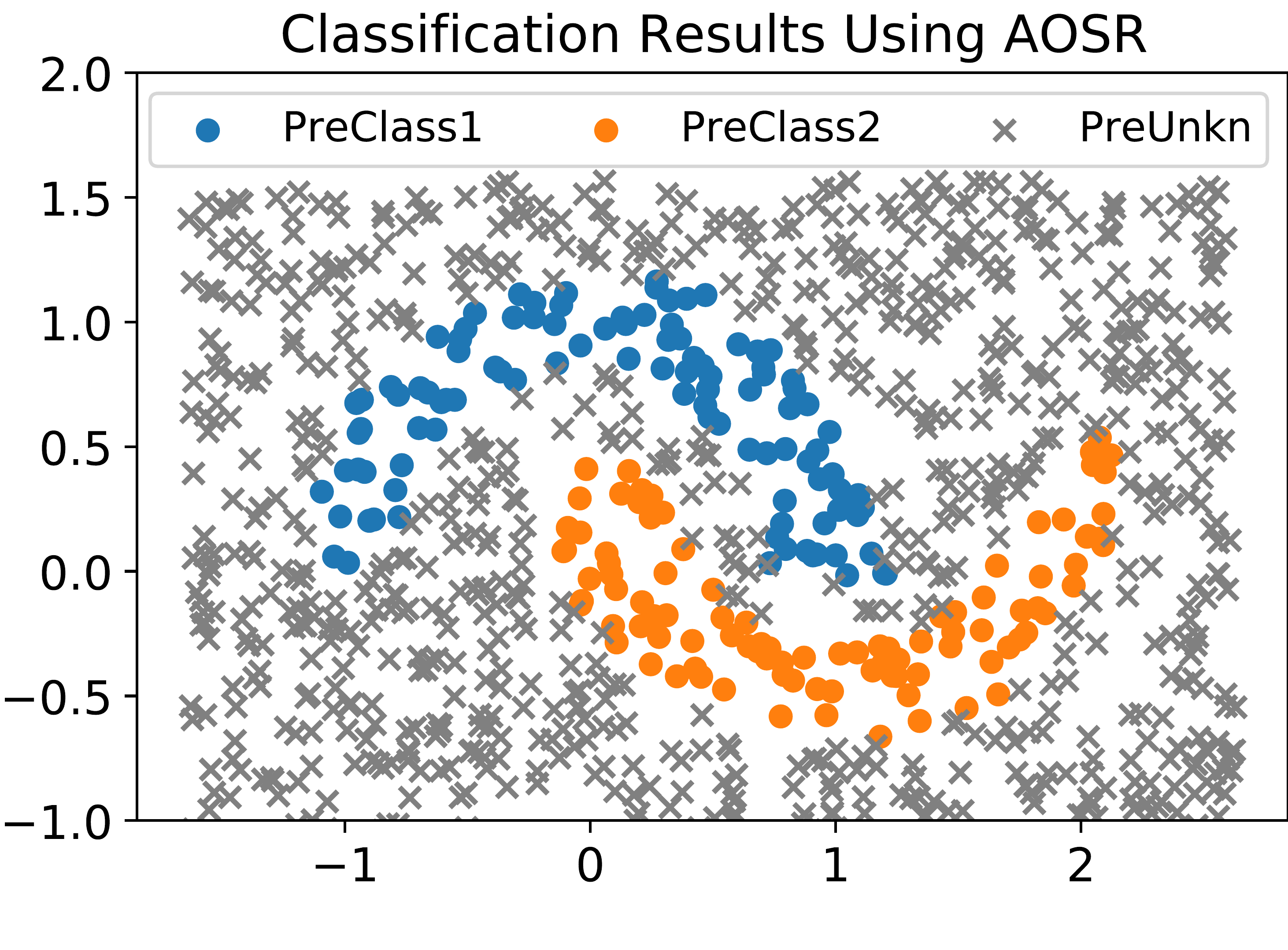}}
    \\ \subfigure[Error and Sample Size]{\includegraphics[scale=0.17, trim=0  270 0 280, clip]{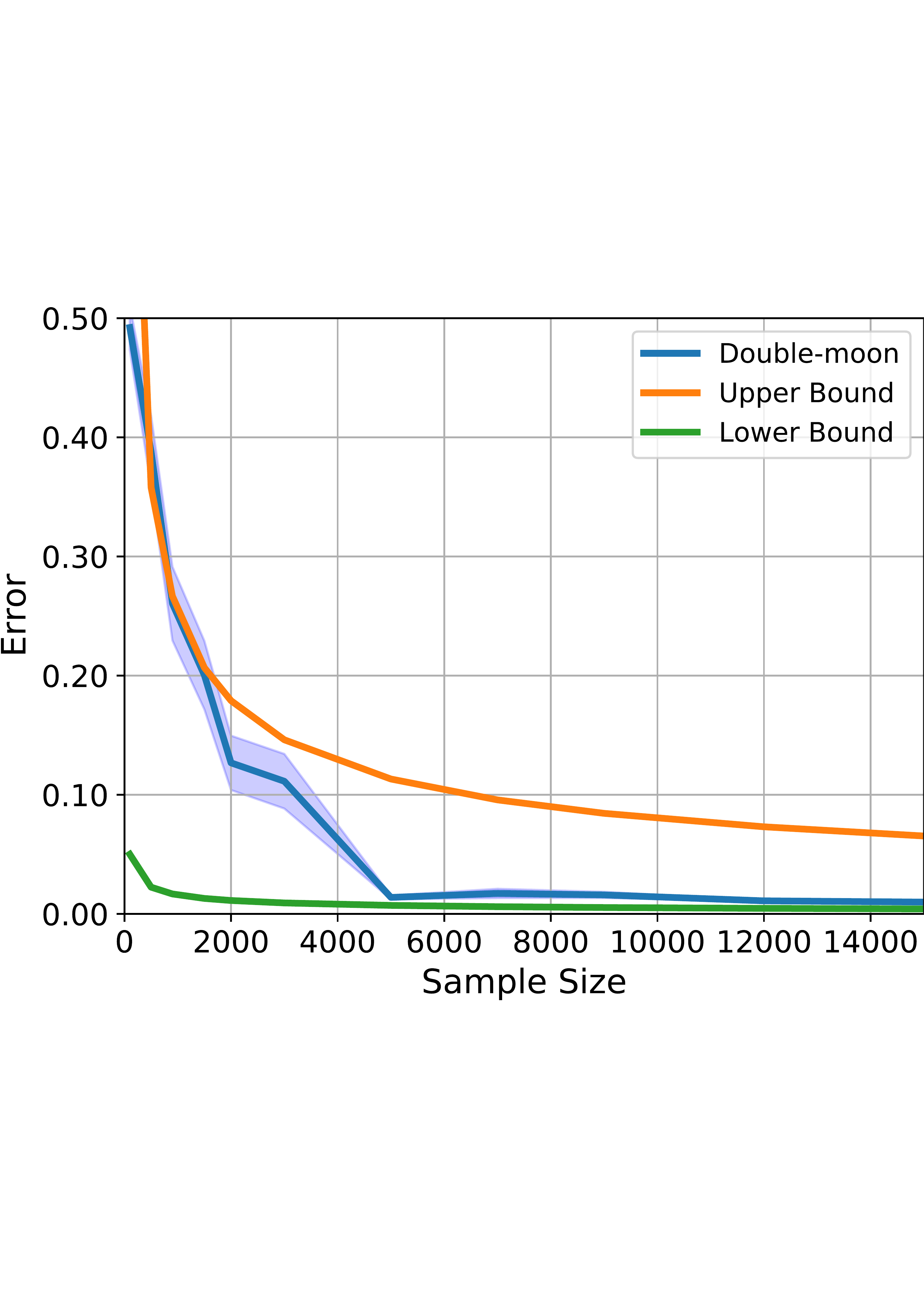}}~~~~~~~
    \subfigure[Parameter Analysis for $\beta$]{\includegraphics[scale=0.17, trim=0  270 0 280, clip]{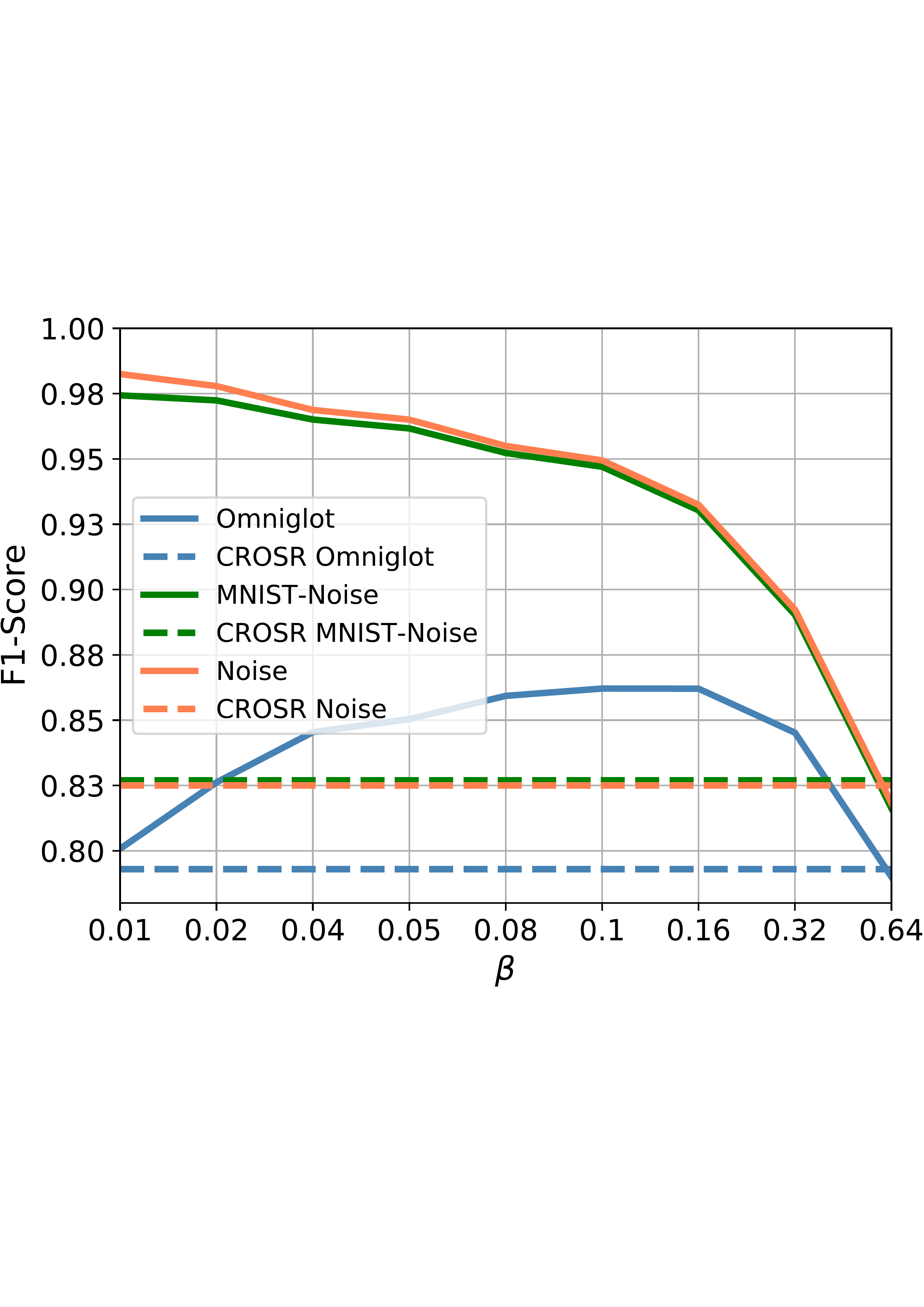}}~~~~~~~
    \subfigure[Parameter Analysis for $t$]{\includegraphics[scale=0.17, trim=0  270 0 280, clip]{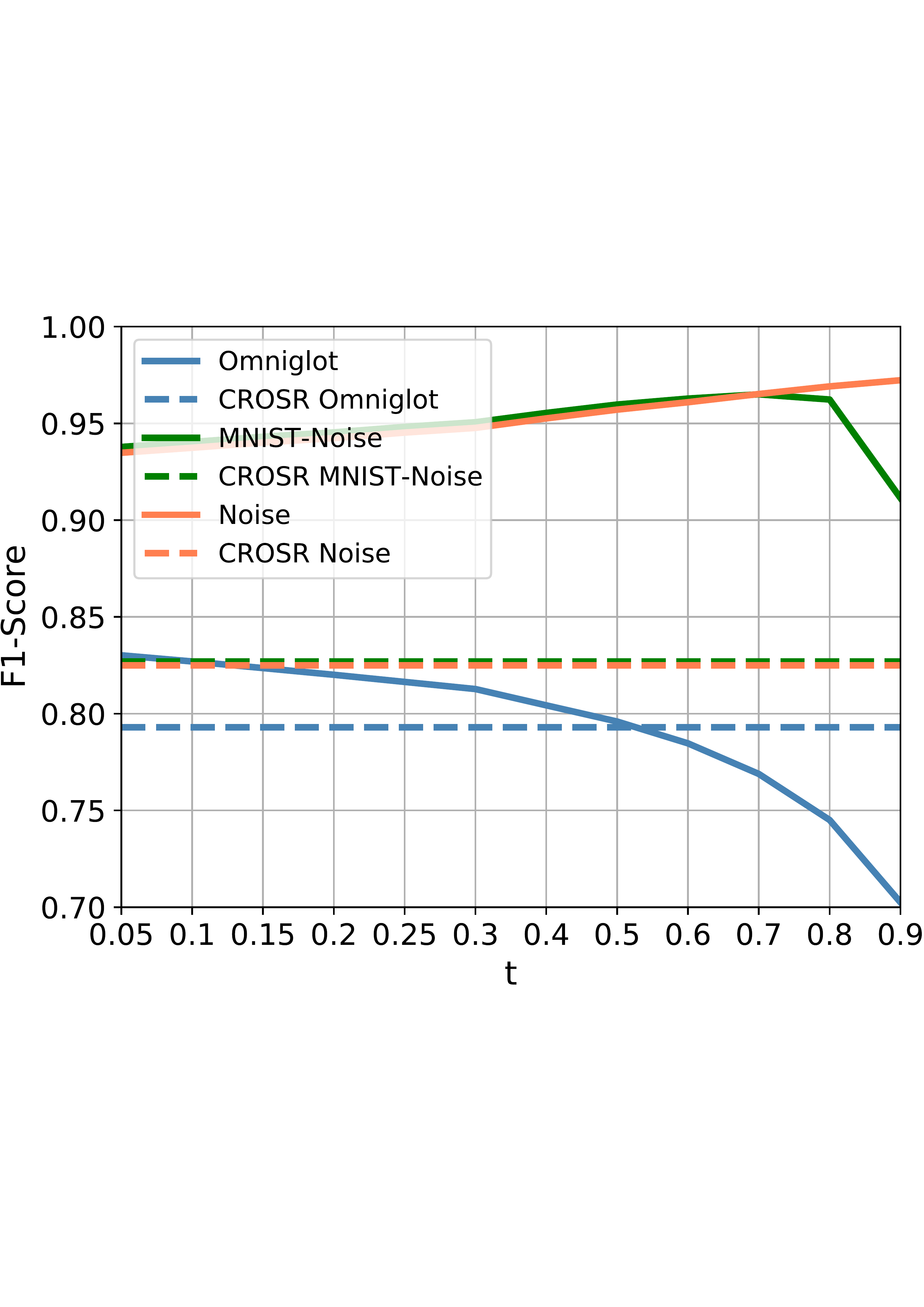}}
    \caption{(a) is the training samples for double-moon dataset. (b) is the decision regions under closed-set learning setting for double-moon dataset. (c) is the decision regions under open-set learning setting for double-moon dataset. (d) is the relationship between error and sample size. (e) is the parameter analysis for $\beta$. (f) is the parameter analysis for $t$.}
    \label{fig:train}
    \vspace{-1em}
\end{figure*}
First, we implement AOSR on toy dataset with different sample size to reveal the relationship between sample size $n$ and error \big($O(1/\sqrt{n})$\big ). Then, we evaluate the efficacy of AOSR on benchmark datasets. 


\subsection{Datasets}

In this paper, we verify the efficacy of algorithm AOSR on double-moon dataset and several real world datasets:

$\bullet$ Double-moon dataset (toy). The double-moon dataset consists of two different clusters. Samples from different clusters are regarded as known samples with different label. Samples from other region are regarded as unknown samples drawn from uniform distribution, i.i.d. The ratio between the sizes of known and unknown samples is $1$.
 
$\bullet$   Following the set up in \citeauthor{DBLP:conf/cvpr/YoshihashiSKYIN19} (\citeyear{DBLP:conf/cvpr/YoshihashiSKYIN19}), we use MNIST \cite{lecun-mnisthandwrittendigit-2010} as the training samples and use Omniglot \cite{ager2008omniglot}, MNIST-Noise, and Noise \cite{liu2021invertible} datasets as unknown classes. Omniglot contains alphabet characters. Noise is synthesized by sampling each pixel value from a uniform distribution on $[0, 1]$. MNIST-Noise is synthesized by adding noise on MNIST test samples. Each dataset has $10, 000$ test samples.
   
$\bullet$    Following \citeauthor{DBLP:conf/cvpr/YoshihashiSKYIN19} (\citeyear{DBLP:conf/cvpr/YoshihashiSKYIN19}), we use CIFAR-$10$ \cite{krizhevsky2010convolutional} as training samples and collect unknown samples from ImageNet and LSUN. We resized/cropped them so that they would be the same size as the known samples. Hence, we generate four datasets ImageNet-crop, ImageNet-resize, LSUN-crop and LSUN-resize as unknown classes. Each dataset contains $10, 000$ test samples.

$\bullet$ Following \citeauthor{DBLP:conf/cvpr/YoshihashiSKYIN19} (\citeyear{DBLP:conf/cvpr/YoshihashiSKYIN19}), \citeauthor{DBLP:conf/eccv/ChenQ0PLHP020} (\citeyear{DBLP:journals/corr/abs-2103-00953}), \citeauthor{DBLP:journals/corr/abs-2003-08823} (\citeyear{DBLP:journals/corr/abs-2003-08823}), we use  MNIST \cite{lecun-mnisthandwrittendigit-2010}, SVHN \cite{netzer2011reading} and CIFAR-$10$ \cite{krizhevsky2010convolutional} to construct different OSL tasks. For MNIST, SVHN and CIFAR-$10$, each dataset is
randomly divided into $6$ known classes and $4$ unknown
classes. In addition, we construct CIFAR+$10$ and CIFAR+$50$ by randomly selection $6$ known classes and $10$ or $50$  unknown classes from CIFAR-$100$ \cite{krizhevsky2010convolutional}. 

\input{sec5.1}
\input{sec5.2}
\input{sec5.3}

%% file: sec5.1.tex
\subsection{Open-set Learning Demonstration}
\begin{table*}[t]
\small
\centering
\caption{The performance on dataset CIFAR-$10$ is evaluated by macro-averaged F1 scores in 11 classes ($10$ known classes and $1$ unknown class). We report the experimental results reproduced by \citeauthor{DBLP:conf/cvpr/YoshihashiSKYIN19} (\citeyear{DBLP:conf/cvpr/YoshihashiSKYIN19}). A larger score is better.}
\vspace{1mm}
\begin{tabular}{lllll}
\toprule
Algorithm            & ImageNet-crop & ImageNet-resize & LSUN-crop & LSUN-resize \\
\midrule
Softmax         & 0.639         & 0.653           & 0.642     & 0.647       \\
Openmax \cite{DBLP:conf/cvpr/BendaleB16}         & 0.660         & 0.684           & 0.657     & 0.668       \\
Counterfactual \cite{DBLP:conf/eccv/NealOFWL18}  & 0.636         & 0.635           & 0.650     & 0.648  
\\
CROSR \cite{DBLP:conf/cvpr/YoshihashiSKYIN19}           & 0.721         & 0.735           & 0.720     & 0.749       \\
C2AE \cite{DBLP:conf/cvpr/OzaP19}              &{0.837}         & {0.826}           & 0.783     & 0.801   \\
CGDL  \cite{DBLP:journals/corr/abs-2003-08823}             &  \textbf{0.840}         &  \textbf{0.832}           & 0.806     & 0.812       \\
\midrule
Ours (AOSR)            & {0.798}         & {0.795}           &\textbf{0.839}     & \textbf{0.838}     \\
\bottomrule
\end{tabular}
\label{tab:cifar10}
\vspace{-1em}
\end{table*}

\begin{table}[t]
\small
\centering
\caption{The performance on dataset MNIST is evaluated by macro-averaged F1 scores in 11 classes.  }
\vspace{1mm}
\begin{tabular}{llll}
\toprule
Algorithm        & Omniglot & MNIST-Noise & Noise \\
\midrule
Softmax   & 0.595    & 0.801       & 0.829 \\
Openmax  & 0.780    & 0.816       & 0.826 \\
CROSR    & 0.793    & 0.827       & 0.826 \\
CGDL   & \textbf{0.850}    & 0.887       & 0.859 \\
\midrule
Ours (AOSR)          &{0.825}    & \textbf{0.953}       & \textbf{0.953} \\
\bottomrule
\end{tabular}
\label{tab:mnist}
\vspace{-1.5em}
\end{table}

\begin{table*}[t]
\small
\centering
\caption{The performance on MNIST, SVHN, CIFAR-$10$, CIFAR+$10$ and CIFAR+$50$ are evaluated by macro-averaged F1 scores. We report the experimental results reported by \citeauthor{DBLP:journals/corr/abs-2003-08823} (\citeyear{DBLP:journals/corr/abs-2003-08823}).}\label{tab:TinyImage}
\vspace{1mm}
\begin{tabular}{llllll}
\toprule
Algorithm            & MNIST & SVHN & CIFAR-$10$ & CIFAR+$10$& CIFAR+$50$ \\
\midrule
Softmax         & 0.768         & 0.725           & 0.600    & 0.701 &0.637       \\
Openmax \cite{DBLP:conf/cvpr/BendaleB16}         & 0.798       & 0.737           & 0.623     & 0.731 &0.676       \\

CROSR \cite{DBLP:conf/cvpr/YoshihashiSKYIN19}           & 0.803         & 0.753           & 0.668     & 0.769   &0.684
\\
GDFR \cite{DBLP:conf/cvpr/PereraMJMWOP20}& 0.821&0.716&0.700&\textbf{0.776}&0.683
\\
 CGDL \cite{DBLP:journals/corr/abs-2003-08823}             & 0.837         & 0.776           & 0.655     & 0.760 &0.695     \\
\midrule
Ours (AOSR)            & \textbf{0.850}         & \textbf{0.842}           &\textbf{0.705}     & {0.773}  & \textbf{0.706}   \\
\bottomrule
\end{tabular}
\vspace{-1em}
\end{table*}

Here we break down the entire learning process and demonstrate the inter-media process of each step on the toy dataset. This experiment is aiming to provide an visualization aid on understanding the open-set learning process.

To start with, we plot the double-moon dataset in Figure \ref{fig:train} (a). The objective of closed-set learning is to build a classifier that can split the samples with different labels. To achieve this goal, we build a simple neural network with sparse categorical cross-entropy as the loss function.

The closed-set learning result is shown in Figure \ref{fig:train} (b). In this case, the closed-set classifier splits the samples with different labels well. However, the closed-set classifier does not consider the boundary of support set for training domain, that is, any new samples that does not located in the support set, the closed-set classifier still gives a known label.

Figure \ref{fig:train} (c) is the open-set learning result. To recognize the unknown samples, the open-set classifier should delineate a boundary between the known
and unknown classes. To achieve this goal, we use $\mathbf{Softmax}_{C+1}$ as the final output and Eq. \eqref{optimial problem1} as the cost function. The AOSR will push the neural network to give label $\mathbf{y}_{C+1}$ on unknown samples. 

%% file: sec5.2.tex
\vspace{-0.5em}
\subsection{Experimental Setup} 
$\bullet$ AOSR has several hyper-parameters: $\beta$, $t$, $\mu$ and $m$. For all tasks, we set $m=3n,~ t=10\%$ as default. $\mu$ is a dynamic parameter depending on $\beta$. $\beta$ is selected from $0.01$ to $2.5$. Details on the selection of parameters are available at \httpsurl{github.com/Anjin-Liu/Openset_Learning_AOSR}. 

$\bullet$ For  datasets MNIST, Omnilot, MNIST-Noise, Noise, we use the same setting of \citeauthor{DBLP:conf/cvpr/YoshihashiSKYIN19} (\citeyear{DBLP:conf/cvpr/YoshihashiSKYIN19}) and \citeauthor{DBLP:journals/corr/abs-2003-08823} (\citeyear{DBLP:journals/corr/abs-2003-08823}) to extract the features. Same as \citeauthor{DBLP:conf/cvpr/YoshihashiSKYIN19} (\citeyear{DBLP:conf/cvpr/YoshihashiSKYIN19}), DHRNet-$92$ is used as the backbone for CIFAR-$10$, ImageNet and LSUN datasets.  For different tasks MNIST, SVHN, CIFAR-$10$, CIFAR+$10$ and CIFAR+$50$, the
backbone is the re-designed VGGNet used by \citeauthor{DBLP:conf/cvpr/YoshihashiSKYIN19} (\citeyear{DBLP:conf/cvpr/YoshihashiSKYIN19}) and \citeauthor{DBLP:journals/corr/abs-2003-08823} (\citeyear{DBLP:journals/corr/abs-2003-08823}).

$\bullet$ We select baseline algorithms as follows: SoftMax, OpenMax \cite{DBLP:conf/cvpr/BendaleB16}, Counterfactual \cite{DBLP:conf/eccv/NealOFWL18}, CROSR \cite{DBLP:conf/cvpr/YoshihashiSKYIN19}, C2AE \cite{DBLP:conf/cvpr/OzaP19},  and CGDL \cite{DBLP:journals/corr/abs-2003-08823}.

\subsection{Evaluation}
 Following \citeauthor{DBLP:conf/cvpr/YoshihashiSKYIN19} (\citeyear{DBLP:conf/cvpr/YoshihashiSKYIN19}), the macro-average F1 scores are used to evaluate OSL. The area under the receiver operating characteristic (AUROC) \cite{DBLP:conf/eccv/NealOFWL18} is also frequently used \cite{DBLP:conf/eccv/NealOFWL18,DBLP:conf/eccv/ChenQ0PLHP020}. Note that AUROC used in \cite{DBLP:conf/eccv/NealOFWL18,DBLP:conf/eccv/ChenQ0PLHP020} is suitable for global threshold-based OSL algorithms that recognize unknown samples by a fix threshold \cite{DBLP:conf/eccv/NealOFWL18}. However, AOSR recognizes unknown samples based on the score of hypothesis function, thus, AOSR uses different thresholds for different samples. This implies that AUROC used in \cite{DBLP:conf/eccv/NealOFWL18,DBLP:conf/eccv/ChenQ0PLHP020} may be not suitable for our algorithm. In this paper, we use  macro-average F1 scores to evaluate our algorithm.

%% file: sec5.3.tex
\subsection{Experimental Evaluation and Result Analysis}



Experiment results on double-moon dataset are summarized in Figure \ref{fig:train} (d). We implement double-moon dataset with varying size $n$ $\footnote{ select $n$ from
       $[1,5,9,15,20,30,50,70,90,120,150]*100$
   }$. We also generate $n$ test samples.  For a different sample size, we run $100$ times and report the mean accuracy and standard error in Figure \ref{fig:train} (d). Based on Figure \ref{fig:train} (d), the accuracy increases as the increase of training sample size $n$ increases. When $n\rightarrow 15,000$,  the accuracy approximates at $100\%$. In particular, the green curve $0.5/\sqrt{n}$ and the yellow curve $8/\sqrt{n}$ jointly control the curve of accuracy, implying the error of AOSR is controlled by $O(1/\sqrt{n})$. 

\begin{table*}[t]
\small
\centering
\caption{Ablation study  on dataset MNIST, Omnilot, MNIST-Noise and Noise.}
\vspace{1mm}
\begin{tabular}{lllllll}
\toprule
 Tasks           & Only iForest & $\beta$=0 & $\mu$=0 & w/KMM& w/KuLSIF& AOSR \\
\midrule
Avg            & {0.680}         & {0.677} & {0.677}          &{0.907}     & {0.855}&\textbf{0.910}     \\
\bottomrule
\end{tabular}
\label{tab:abs}
\vspace{-1em}
\end{table*}

Experiment results on real datasets are summarized in Tables \ref{tab:cifar10}, \ref{tab:mnist} and \ref{tab:TinyImage}. For all tasks, we run AOSR $5$ times and report the mean results by using F1 score \cite{powers2020evaluation}.  In general, AOSR shows the promising performance when compared to baseline algorithms. The effectiveness of AOSR indicates that our theory is effective and practical.

Parameter analysis for $\beta$ and $t$ is given in Figure  \ref{fig:train} (e), (f).  We run AOSR with varying values of $\beta,t$ on MNIST tasks. From Figure \ref{fig:train} (e), we observe that 1) when $\beta$ increases from $0.01$ to $0.64$, the F1 scores  for Noise and MNIST-Noise decrease;  2) as increasing $\beta$ from $0.01$ to $0.16$, the F1 score for Omniglot increases. When $\beta > 1.6$, the performance for Omniglot dramatically dropped to baseline. Additionally, according to Figure  \ref{fig:train} (f),   we find that by changing $t$ in the range of $[0.05,0.30]$, AOSR  achieve stable performance.

Ablation study on datasets MNIST, Omnilot, MNIST-Noise and Noise is shown in Table \ref{tab:abs}. By adjusting different components of AOSR, Table \ref{tab:abs} indicates that each component of AOSR is important and necessary. Note that if we replace iForest by KMM in AOSR, the performance ($0.907$) is close to AOSR ($0.910$). This implies that KMM may be a good choice, if we omit the time complexity of KMM. 